%% file: main.tex
\documentclass{arxivtmpl}

\input{commands}

\usepackage{url}
\usepackage{longtable}
\usepackage{array}
\usepackage{xltabular}
\usepackage{pdflscape}
\usetikzlibrary{shapes.geometric,arrows.meta,calc}
\usepackage{pgfplots}
\pgfplotsset{compat=1.18}
\definecolor{SkillBlue}{HTML}{011F5B}
\definecolor{SkillPaleBlue}{HTML}{E8ECF3}
\definecolor{SkillInk}{HTML}{1A1A1A}
\definecolor{navyblue}{HTML}{0071BC}
\makeatletter
\newcommand{\panellabel}[2]{\begingroup\protected@edef\@currentlabel{\thefigure#1}\label{#2}\endgroup}
\makeatother
\iftrue
\newcommand{\displaytodo}[1]{#1}
\else
\newcommand{\displaytodo}[1]{}
\fi

\usepackage{float}
\usetikzlibrary{positioning,backgrounds,fit,decorations.pathreplacing}
\input{macros}

\usepackage{fontawesome5}

\definecolor{titlered}{HTML}{A90533}
\renewcommand{\titlefont}{\normalfont\bfseries\fontsize{21}{26}\selectfont\color{titlered}}

\title{InteractionBench: A Real-Time Interaction Benchmark for Streaming Video Systems}

\author[1,*]{Enxin Song}
\author[1,*]{Suhao Yu}
\author[1,*]{Yifei Xu}
\author[2]{Barbara Su}
\author[3]{Weili Xu}
\author[4]{Wenhao Chai}
\author[1]{Yao Tang}
\author[5]{Jie Deng}
\author[6]{Haiyang Xu}
\author[1]{Jiatao Gu}
\affil[1]{University of Pennsylvania}
\affil[2]{Stanford University}
\affil[3]{University of Illinois at Urbana-Champaign}
\affil[4]{Princeton University}
\affil[5]{Hong Kong Polytechnic University}
\affil[6]{University of California, San Diego}

\authornote{\faGlobe~\textbf{Project Page:} \href{https://www.enxinsong.com/projects/interactionbench/}{https://www.enxinsong.com/projects/interactionbench/}\\
  \faGithub~\textbf{GitHub:} \href{https://github.com/Espere-1119-Song/InteractionBench}{https://github.com/Espere-1119-Song/InteractionBench}\\
  \faDatabase~\textbf{Hugging Face:} \href{https://huggingface.co/datasets/InteractionBench/InteractionBench}{https://huggingface.co/datasets/InteractionBench/InteractionBench}}

\begin{document}

\maketitle
\footnotetext[1]{Equal contribution}

\thispagestyle{firstpagestyle}

\vspace{-8pt}
\input{tex/0_abstract}

\teaserfigure{\includegraphics[width=\linewidth]{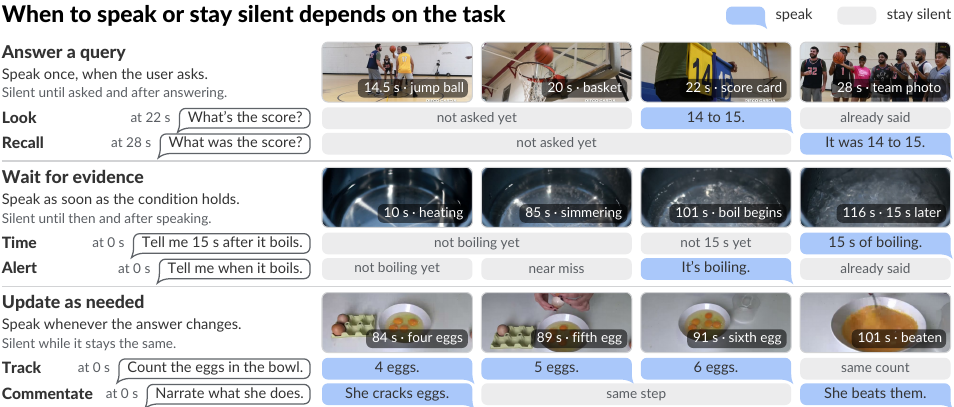}}{\textbf{A frame can require speech in one task and silence in its pair.} Each task pair shares one video, and each cell shows the reply or silence required at that frame.\label{fig:timeline}\label{fig:teaser}\label{fig:main_case}}

\clearpage
\input{tex/1_introduction}

\clearpage
\setcounter{tocdepth}{2}
\begingroup
\hypersetup{linkcolor=black}
\tableofcontents
\endgroup
\setcounter{tocdepth}{2} %
\clearpage

\input{tex/2_benchmark}
\input{tex/3_experiments}
\input{tex/4_conclusion}
\input{tex/5_statements}

\clearpage

\bibliography{references}

\input{tex/6_appendix}

\end{document}

%% file: commands.tex
\def\1{\bm{1}}

\DeclareMathAlphabet{\mathsfit}{\encodingdefault}{\sfdefault}{m}{sl}
\SetMathAlphabet{\mathsfit}{bold}{\encodingdefault}{\sfdefault}{bx}{n}

\tcbuselibrary{skins,breakable}
\newtcolorbox{findingbox}{enhanced,breakable,frame hidden,colback=gred!6,borderline west={3pt}{0pt}{gred},sharp corners,boxsep=0pt,left=12pt,right=10pt,top=8pt,bottom=8pt,before skip=12pt,after skip=12pt}
\newtcolorbox{notebox}{enhanced,breakable,frame hidden,colback=gblue!6,borderline west={3pt}{0pt}{gblue},sharp corners,boxsep=0pt,left=12pt,right=10pt,top=8pt,bottom=8pt,before skip=12pt,after skip=12pt}
\newtcolorbox{tipbox}{enhanced,breakable,frame hidden,colback=ggreen!6,borderline west={3pt}{0pt}{ggreen},sharp corners,boxsep=0pt,left=12pt,right=10pt,top=8pt,bottom=8pt,before skip=12pt,after skip=12pt}
\newtcolorbox{warnbox}{enhanced,breakable,frame hidden,colback=gyellow!6,borderline west={3pt}{0pt}{gyellow},sharp corners,boxsep=0pt,left=12pt,right=10pt,top=8pt,bottom=8pt,before skip=12pt,after skip=12pt}

\tcbuselibrary{listings}
\newcommand{\boxfont}{\fontsize{9}{11}\selectfont}
\newtcblisting{promptbox}[1][]{%
  enhanced, listing only, breakable, frame hidden, width=\linewidth,
  colback=gblue!6, borderline west={3pt}{0pt}{gblue},
  sharp corners, boxsep=0pt,
  left=12pt, right=10pt, top=8pt, bottom=8pt, before skip=12pt, after skip=12pt,
  before upper={\def\promptlabel{#1}\ifx\promptlabel\empty\else{\sffamily\bfseries\color{gblue}#1}\par\smallskip\fi}, %
  listing options={basicstyle=\ttfamily\boxfont, escapeinside={(*}{*)}},
}

\newtcolorbox{transcriptbox}{%
  enhanced, breakable, frame hidden,
  colback=ggreen!6, borderline west={3pt}{0pt}{ggreen},
  sharp corners, boxsep=0pt, fontupper=\ttfamily\boxfont, halign upper=flush left, %
  left=12pt, right=10pt, top=6pt, bottom=8pt, before skip=12pt, after skip=12pt,
}

\newcommand{\teaserfigure}[2]{%
  \begingroup\centering\captionsetup{type=figure}
  #1\par\vspace{0.4em}\caption{#2}%
  \par\endgroup\vspace{1em}
}

\colorlet{codebg}{gyellow!6}
\setminted{fontsize=\boxfont, breaklines, autogobble, tabsize=4, style=vs}
\newminted[codebox]{python}{bgcolor=codebg, frame=leftline, framerule=2.5pt, rulecolor=gyellow, framesep=2.5mm}

\hypersetup{
    colorlinks=true,
    linkcolor=linkcol,
    citecolor=linkcol,
    urlcolor=linkcol,
    pdfborderstyle={},
    pdfborder={0 0 0}
}
\renewcommand\Affilfont{\sffamily\mdseries}

\colorlet{perfhue}{gblue}
\colorlet{perf0}{white}
\colorlet{perf10}{perfhue!12}
\colorlet{perf20}{perfhue!22}
\colorlet{perf30}{perfhue!33}
\colorlet{perf40}{perfhue!44}
\colorlet{perf50}{perfhue!55}
\colorlet{perf60}{perfhue!66}
\colorlet{perf70}{perfhue!77}
\colorlet{perf80}{perfhue!88}

\newcommand{\RN}[1]{%
	\textup{\lowercase\expandafter{\it \romannumeral#1}}%
}

\newcommand{\cmark}{\ding{51}}%
\newcommand{\xmark}{\ding{55}}%

\SetKwInput{KwInput}{Input}
\SetKwInput{KwOutput}{Output}
\DontPrintSemicolon

\SetKwComment{Comment}{\color{ggreen}\# }{}

\SetKwProg{Function}{def}{:}{}

\SetKwProg{For}{for}{:}{}
\SetKwProg{If}{if}{:}{}

%% file: macros.tex
\usepackage{xspace}

\newcommand{\vigil}{\textsc{InteractionBench}\xspace}  %
\newcommand{\bench}{\textsc{InteractionBench}\xspace}

\newcommand{\ivqa}{\textsc{Look}\xspace}

\newcommand{\toa}{\textsc{Time}\xspace}

\newcommand{\lvm}{\textsc{Recall}\xspace}

\newcommand{\ptr}{\textsc{Alert}\xspace}

\newcommand{\cst}{\textsc{Track}\xspace}

\newcommand{\lcg}{\textsc{Commentate}\xspace}

\newcommand{\resp}{\ensuremath{\hat{\mathcal{E}}}}      %
\newcommand{\rstar}[1]{\ensuremath{r_{#1}^{\ast}}}      %
\newcommand{\win}[1]{\ensuremath{W_{#1}}}               %
\newcommand{\dly}[1]{\ensuremath{d_{#1}}}               %
\newcommand{\dbound}{\ensuremath{\Delta}}               %
\newcommand{\viol}{\ensuremath{V}}                      %

\newcommand{\Acc}{\ensuremath{\mathrm{Acc}}}            %
\newcommand{\TA}{\ensuremath{\mathrm{TA}}}              %
\newcommand{\SC}{\ensuremath{\mathrm{SC}}}              %
\newcommand{\Total}{\ensuremath{\mathrm{Overall}}}        %
\newcommand{\pairsel}{\ensuremath{\mathrm{PairSel}}}    %
\newcommand{\vdc}{\textsc{VDCScore}\xspace}             %

\definecolor{markyes}{HTML}{137333}
\definecolor{markno}{HTML}{9AA0A6}
\renewcommand{\cmark}{\textcolor{markyes}{\ding{52}}}  %
\renewcommand{\xmark}{\textcolor{markno}{\ding{55}}}   %
\newcommand{\best}[1]{\textbf{#1}}

\definecolor{vigilblue}{HTML}{6FA1F2}
\definecolor{vigilorange}{HTML}{D9722C}
\definecolor{vigilgray}{HTML}{6B7280}
\definecolor{vigilgreen}{HTML}{47AF7D}
\definecolor{vigilred}{HTML}{DE6F66}
\definecolor{pennblue}{HTML}{6FA1F2}
\definecolor{pennred}{HTML}{DE6F66}
\definecolor{mblue}{HTML}{4C72B0}
\definecolor{morange}{HTML}{DD8452}
\definecolor{mgreen}{HTML}{55A868}
\definecolor{mred}{HTML}{C44E52}
\definecolor{pnavy}{HTML}{003049}
\definecolor{psteel}{HTML}{669BBC}
\definecolor{pcrim}{HTML}{C1121F}
\definecolor{pmaroon}{HTML}{780000}
\definecolor{pcream}{HTML}{FDF0D5}
\definecolor{pnavy}{HTML}{003049}
\definecolor{psteel}{HTML}{669BBC}
\definecolor{pcrim}{HTML}{C1121F}
\definecolor{pmaroon}{HTML}{780000}
\definecolor{pcream}{HTML}{FDF0D5}

\newif\ifshownotes
\shownotestrue                                   %

%% file: tex/0_abstract.tex
\begin{abstract}
{\sffamily\fontseries{eb}\normalsize\selectfont Abstract}\par\vspace{0.15\baselineskip}
\small
A video assistant must speak when its instruction warrants a response and
stay silent otherwise. We introduce a benchmark that evaluates
this decision for the complete system of model, memory, and response
controller. \vigil{} covers query responses, event triggers, and ongoing
updates in 1{,}060 interactions over 812 videos, with 69 negative streams
and 53 suites that pair counted events with look-alike near misses. It
scores content accuracy, timing accuracy, and silence compliance on the
video clock. Timely speech costs
silence across systems. Polled Qwen3-VL-8B \citep{bai2025qwen3vl} reaches
77.8 timing accuracy but 10.9 silence compliance. A native real-time
interaction system reaches 29.2 silence compliance at 66.8 timing accuracy,
yet emits on 89.9\% of negative streams. No open-weight system clears a
third of the near-miss suites. Fewer replies help only when chosen, as
random deletion merely trades timing for silence. Offline scores miss these
failures and mispredict online behavior. Adding restraint is costly, as the
native system's controller adds little by itself and agentic systems add it
only at about 30\,s per poll.

\end{abstract}
\vspace{1.0em}

%% file: tex/1_introduction.tex
\section{Introduction}
\label{sec:intro}

``Tell me when the water boils'' asks an assistant to make two decisions.
It must respond when boiling becomes visible and remain silent while the
pot merely simmers. After reporting the event, it should keep watching
without repeating the same message. Answering a question about a sampled
frame tests only part of this behavior. The complete interaction also
depends on when the system speaks, when it waits, and when it resumes monitoring after it has responded.

Streaming video models process frames as they arrive and respond during the stream
\citep{chen2024videollmonline,zhang2024flashvstream,wang2024mmduet}.
Interaction systems add memory and turn management around a visual
backbone \citep{joyaivl2026}. Their observable behavior therefore belongs
to the complete system. A capable model can still speak at every poll,
repeat an answer, or stop attending after its first response.
Conversely, a system can avoid false alarms by missing every event.
Useful interaction requires both timely speech and disciplined silence,
and an evaluation must score both.

Existing benchmarks study proactive responses and streaming
interaction \citep{proactivevideoqa2025,river2026,omnipro2026,omniinteract2026}.
SPOT-Bench \citep{spotbench2026} and LiveProBench
\citep{liveprobench} penalize premature and repeated replies but include no
negative streams or near misses.
\vigil{} focuses on the paired obligation created by a standing
instruction, to respond to sufficient evidence and withhold speech when that
evidence is absent. We annotate the earliest answerable moments, score
repeated emissions, include streams where the requested event never
occurs, and pair real events with visually similar near misses.
These tests separate detecting a scene from judging whether its instruction is met.

\vigil{} contains 812 videos and 1{,}060 interactions across six tasks.
\ivqa{} and \lvm{} answer a user query. \toa{} and \ptr{} wait for
temporal evidence or an event. \cst{} and \lcg{} update repeatedly as the
stream changes. Tasks that wait or update hold 69.6\% of interactions. The same visual task can require speech at one
moment and silence at another. Our protocol scores content accuracy, timing
accuracy, and silence compliance, and combines them into one Overall score.
It reports wall-clock latency separately, as a property of the serving
stack.

The evaluated systems show different imbalances between these obligations.
Qwen3-VL-8B \citep{bai2025qwen3vl} under a polling adapter reaches 34.8 Overall, with timing
accuracy of 77.8 but silence compliance of 10.9. Only 5 of 18 timed configurations finish 90\% of
one-second polls in time, and agents take about half a minute per poll. Negative streams and matched near misses reveal
additional false alarms, even among systems that speak less often.
Controlled adapter changes and offline comparisons on identical items show that
offline grounding and online decisions rank systems differently, and
that denser frame sampling buys little silence.
We make three contributions:
\begin{itemize}
  \item We introduce \vigil{}, which tests when a video assistant should speak
        or stay silent across query responses, event triggers, and ongoing updates
        in 1{,}060 interactions.
  \item Our stream-time protocol separates answer content, response timing, and
        unnecessary speech, and tests silence directly with negative streams and
        matched near-miss suites.
  \item We evaluate native streaming systems, polled turn-based models, and agents
        on \vigil{}, separate repeated speech, premature responses, false alarms, and
        missed follow-up events, and find that timely speech costs silence across systems.
\end{itemize}

%% file: tex/2_benchmark.tex
\section{Evaluating Speech and Silence}
\label{sec:bench}

\subsection{The System Decision}
\label{sec:task}
\label{sec:protocol}

A system receives a video stream $v$ and a question or standing instruction
$q$. At stream time $t$, it can access only the video prefix $v_{\leq t}$,
its instruction, and its own interaction history. It chooses to remain
silent or emit text $y$. The output is a time-ordered sequence
$\resp=((t_1,y_1),\ldots,(t_M,y_M))$ of $M$ non-empty emissions, where
emission $i$ carries text $y_i$ at stream time $t_i$ on the video clock.
The evaluated system includes its visual model, memory, sampling policy,
and response controller. A polling adapter is one such controller for
turn-based models. It asks the model at every stream second whether to
speak at that moment, and silence
is a valid answer to any poll.

Each positive item has $N$ reference events. Event $n$ has reference
content $c_n$ and an earliest answerable moment \rstar{n}, defined by the
video prefix that first supplies sufficient evidence. A negative item has
no event, and silence is correct throughout. As shown in
Figure~\ref{fig:teaser}, the required response pattern depends on the task.
Accurate content alone cannot establish a correct interaction, because the
same answer can be premature, timely, or redundant.
The case studies in Appendix~\ref{app:cases} add late replies, missed responses,
wrong content, and false alarms on a negative stream, all on one railway video.

\subsection{Three Complementary Scores}
\label{sec:timing}
\label{sec:families}

For event-based items, event $n$ owns the response window
$\win{n}=[\rstar{n}-\epsilon,\rstar{n+1}-\epsilon)$.
The last window extends to the end of the stream. We use
$\epsilon=1$\,s, so a reply in the same second as the evidence is on time. The first emission in a
window is its matched response, regardless of content. A matched response
at time $t_i$ has delay $\dly{n}=\max(0,t_i-\rstar{n})$. Later emissions in
the same window are redundant, and emissions before the first window are
premature. Redundant and premature emissions form the violation set \viol. With $H(a,b)=2ab/(a+b)$ for $a+b>0$ and
$H(0,0)=0$, the item scores are
\begin{equation}
\label{eq:score}
\begin{aligned}
\TA &= \frac{100}{N}\sum_{n=1}^{N}
       \left(1-\frac{\dly{n}}{\dbound}\right)_{+},
&\SC &= 100\left(1-\frac{|\viol|}{N}\right)_{+},\\
\Total &= \frac{\Acc+H(\TA,\SC)}{2},
&\dbound &= 5\,\mathrm{s}.
\end{aligned}
\end{equation}
Here $(x)_{+}=\max(x,0)$, an unmatched event adds nothing to \TA, and
timing credit decays linearly to zero once lateness reaches the delay
scale \dbound. Content accuracy \Acc{} grades what the matched response
says, timing accuracy \TA{} measures when it arrives, and silence
compliance \SC{} penalizes unnecessary emissions. The harmonic mean
balances the two behavioral components, and averaging it with \Acc{} keeps
credit for accurate content when timing or silence fails. Dataset Overall
averages item scores rather than combining the averaged \Acc, \TA, and
\SC.

Scoring follows the response format. Query-answer items use the first
post-query response, with the query as the timing anchor. Trigger items
average content scores over matched responses, while counting items use
exact final-count accuracy. Free-form updates concatenate speech within
each reference segment, and \vdc{} \citep{chai2024auroracap} checks
whether that text answers questions derived from the reference.
Their Overall equals segment accuracy, including zeros for empty segments.
Commentary may need several utterances per segment, and their timing and
silence scores therefore serve as diagnostics. A negative item
with $M$ emissions scores $\Total=\SC=100\max(0,1-M)$, and a single
emission forfeits all of its credit. See Appendix~\ref{app:scoring} for edge cases,
a uniform variant, and fallback matching.

Constant speech and blanket silence both score poorly. A fully
silent positive item receives zero Overall. Its silence component is
undefined and drops out of the silence aggregate, and blanket silence
therefore earns no positive-item credit. Silence compliance averages
over negative streams and the positive items a system speaks on. Among six scripted policies, one
that always speaks reaches 98.8 timing accuracy but 0.0 Overall. One that
never speaks earns only the 6.5 Overall of the negative streams, and no
policy exceeds 15.6 Overall, see Appendix~\ref{sec:gamed}.

\subsection{Three Response Obligations}
\label{sec:taxonomy}

As shown in Figure~\ref{fig:timeline}, the six tasks differ in the
evidence they need and in what each new response must contain. Look
answers a query from the current view, whereas Recall draws on earlier
visual evidence for a later query. Time waits for a temporal relation such
as an order, duration, or scheduled condition, whereas Alert watches for a
specified visual event. Track maintains an accumulated count or state
across events, whereas Commentate describes the scene as it unfolds, as in
sports play-by-play. Each pair shares one response obligation. Silence is required before evidence arrives, between
unchanged states, and after a reported event, and negative streams and
near misses test it directly. The display names map to the released
labels and sub-labels, see Appendix~\ref{app:bench}.
In Figure~\ref{fig:main_case}, each task pair shares one video
\citep{navy2025basketball,sertion2011boiling,eleneb2021tortilla} and faces
identical frames. When the water begins to boil, Alert speaks while Time
waits 15 more seconds. Track reports each egg added to the bowl, whereas
Commentate stays silent until the next step. See Appendix~\ref{app:cases} for the provenance and sources of all three videos.

\subsection{Data and Quality Checks}
\label{sec:curation}
\input{fig_tex/fig5_stats}

We collect procedural and monitoring videos where a visible change,
completed action, or scheduled step anchors a response. Automatic
filtering and human screening exclude unsuitable clips, broken timelines,
and answer overlays. Annotators define an observable response condition
before writing the instruction, reference answer, and earliest answerable
timestamp. Positive items must be solvable from the prefix, and negative
items must keep their condition false throughout the video. Screening also
rejects ambiguous intent, and a text-only audit revises or removes keyed
questions that a language model answers without frames.
\label{sec:stats}As shown in Figure~\ref{fig:stats}, the 812 videos
span ten domains, led by sports, and average 159\,s. Their 1{,}060
interactions comprise 626 questions and 434 instructions. The 991 positive
interactions hold 3{,}693 reference events, and 69 are negative streams.

The 53 matched suites test whether a system answers counted events but
stays silent at near misses. Each suite pairs a positive counting item with
near-miss moments on the same video, 106 in total, and a near miss
resembles the counted event without completing it. For suite $j$, let
$h_j=1$ if its positive item receives nonzero timing credit, and let
$s_j=1$ if the system emits nothing within five seconds of every near
miss. Both are 0 otherwise. The suite pass rate \pairsel{} is the
percentage of the 53 suites with $h_js_j=1$. The positive criterion is timing credit, and
content accuracy is reported alongside this diagnostic. Suites
cover counting streams only, see Appendix~\ref{app:eval} for how overlapping near-miss and response windows are scored.

\paragraph{Unique silence checks.}
\input{fig_tex/tab_benchmarks}
As shown in Table~\ref{tab:benchmarks}, \vigil{} is the only compared
benchmark that combines whole negative streams with near misses from the
same video. The comparison covers ProactiveVideoQA
\citep{proactivevideoqa2025}, RIVER \citep{river2026}, OmniPro
\citep{omnipro2026}, OmniInteract \citep{omniinteract2026}, ESTP-Bench
\citep{estpbench}, ProAssist \citep{proassist}, IPIBench
\citep{ipibench2026}, SPOT-Bench \citep{spotbench2026}, PaSBench-Video
\citep{pasbench2026}, and LiveProBench \citep{liveprobench}. The dataset
columns mark whether one instruction can require several responses and
whether near misses are annotated, with item counts as each benchmark
defines them. The evaluation columns mark whether the system itself decides
when to respond and whether premature or redundant replies are penalized. \vigil{} penalizes
premature replies as silence violations and late replies by a linear
decay. Streaming question-answering benchmarks such as StreamingBench
\citep{lin2024streamingbench}, OVO-Bench \citep{li2025ovobench}, SVBench
\citep{yang2025svbench}, and OmniMMI \citep{omnimmi2025} include no negative
streams and are compared with these columns in Table~\ref{tab:benchmarks_qa}.

%% file: fig_tex/fig5_stats.tex
\begin{figure}[tp]
\centering
\includegraphics[width=\linewidth]{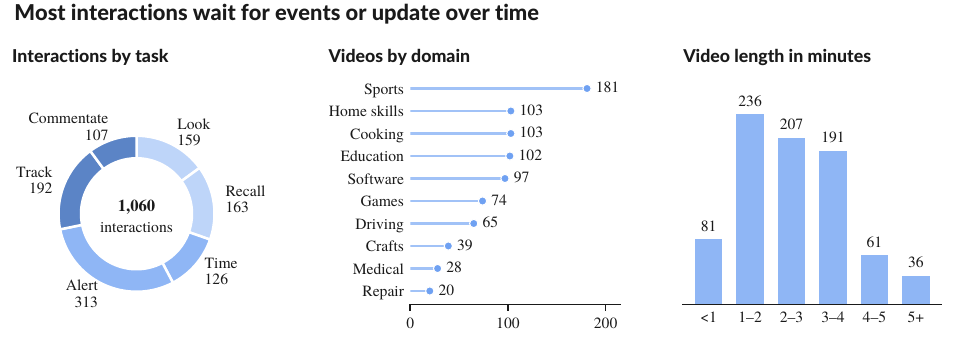}
\caption{\textbf{Most interactions wait for events or update over time.} Left counts interactions by task, Middle videos by domain, and Right videos by length, where the last bin holds five minutes or more.}
\label{fig:stats}
\end{figure}

%% file: fig_tex/tab_benchmarks.tex
\begin{table}[tp]
\centering\fontsize{7.9}{9.8}\selectfont %
\setlength{\tabcolsep}{2pt}
\caption{\textbf{Only \vigil{} pairs negative streams with near misses.} Negatives are streams that warrant no response, and graded delay lowers credit continuously with lateness.}
\label{tab:benchmarks}
\begin{tabular*}{\textwidth}{@{\extracolsep{\fill}}lrrccccccc@{}}
\toprule
& \multicolumn{5}{c}{\textbf{Dataset}} & \multicolumn{4}{c}{\textbf{Evaluation}} \\
\cmidrule(lr){2-6}\cmidrule(lr){7-10}
\textbf{Benchmark} & \textbf{Videos} & \textbf{Items} & \textbf{Multi-response} & \textbf{Negatives} & \textbf{Near misses} & \textbf{Proactive} & \textbf{Premature} & \textbf{Graded delay} & \textbf{Redundant} \\
\midrule
ProactiveVideoQA~\citep{proactivevideoqa2025} & 1,377 & 1,427 & \cmark & \xmark & \xmark & \cmark & \xmark & \cmark & \xmark \\
RIVER~\citep{river2026} & 1,067 & 4,278 & \cmark & \xmark & \xmark & \cmark & \cmark & \cmark & \xmark \\
OmniPro~\citep{omnipro2026} & 1,262 & 2,700 & \cmark & \xmark & \xmark & \cmark & \cmark & \xmark & \cmark \\
OmniInteract~\citep{omniinteract2026} & 250 & 1,430 & \cmark & \xmark & \xmark & \cmark & \cmark & \cmark & \cmark \\
ESTP-Bench~\citep{estpbench} & 890 & 2,264 & \cmark & \xmark & \xmark & \cmark & \cmark & \cmark & \cmark \\
ProAssist~\citep{proassist} & 340 & 1,020 & \cmark & \xmark & \xmark & \cmark & \cmark & \cmark & \cmark \\
IPIBench~\citep{ipibench2026} & 1,831 & 3,738 & \xmark & \xmark & \xmark & \cmark & \cmark & \xmark & \xmark \\
SPOT-Bench~\citep{spotbench2026} & 940 & 5,106 & \cmark & \xmark & \xmark & \cmark & \cmark & \cmark & \cmark \\
PaSBench-Video~\citep{pasbench2026} & 740 & 740 & \xmark & \cmark & \xmark & \cmark & \cmark & \xmark & \xmark \\
LiveProBench~\citep{liveprobench} & 911 & 18,146 & \cmark & \xmark & \xmark & \cmark & \cmark & \xmark & \cmark \\
\midrule
\textbf{\vigil{}} & 812 & 1,060 & \cmark & \cmark & \cmark & \cmark & \cmark & \cmark & \cmark \\
\bottomrule
\end{tabular*}
\end{table}

%% file: tex/3_experiments.tex
\section{Speech and Silence in Current Systems}
\label{sec:experiments}
\subsection{Evaluation Settings}
\label{sec:setup}
\input{fig_tex/tab11_fullset}
We group systems by the component that decides when to speak. JoyAI-VL
\citep{joyaivl2026} is a native real-time interaction system whose own
controller and memory make that choice, and kernel swaps keep them but
replace its model. Native streaming video LLMs use their released
mechanisms and input rates, as in MMDuet2 \citep{wang2025mmduet2},
VideoLLM-online \citep{chen2024videollmonline}, VideoChat3
\citep{videochat3}, MOSS-Video \citep{moss2026}, LiveCC
\citep{chen2025livecc}, and the streaming controller AsynKV
\citep{spotbench2026}. OneStreamer-4B \citep{zeng2026onestreamer}, evaluated and
contributed by its authors, uses its native three-action protocol. Turn-based models answer one prompt at a time, and
a polling adapter asks them at every stream second whether to speak. It
gives each model a sliding window of 16 frames at 2\,fps and a 512-pixel
long side, with a fresh prompt at every poll. The turn-based models are
Qwen3-VL-8B and Qwen3-VL-4B \citep{bai2025qwen3vl}, LLaVA-OV2-8B
\citep{llavaov2}, LLaVA-OV \citep{li2025llavaonevision}, Gemma-3n-E4B
\citep{google2025gemma3n}, MiniCPM-o-4.5 \citep{minicpmo2026},
Qwen2.5-Omni-7B \citep{xu2025qwen25omni}, Claude Sonnet 5
\citep{anthropic2026claude}, and Gemini 2.5 Flash \citep{google2026gemini}.
Mage-VL-4B \citep{yang2026mage} runs under the same adapter, evaluated and
contributed by its authors.
Flash-VStream \citep{zhang2024flashvstream} is a streaming video LLM without
a response decision of its own, and the same adapter polls its streaming
memory. The agentic interaction system wraps a turn-based model in a
tool-using loop with Qwen-MM-Plugins \citep{qwenmmplugins2026} at each poll.
One Claude Sonnet 5 run uses Claude Code tools \citep{anthropic2026claudecode} instead, see Appendix~\ref{app:agents}.

Offline temporal grounding gives a system the whole video and asks when and what it would respond. The six open backbones, Claude Sonnet 5~\citep{anthropic2026claude}, and the Claude and GPT-5.6 Sol \citep{openai2026gpt} agents run in this mode. Grounding thus has future access,
whereas online systems see only the prefix. Dispider
\citep{qian2025dispider} releases only whole-video inference and runs in
this mode as well. Two references anchor the comparison. A blind Qwen3-VL-8B~\citep{bai2025qwen3vl} run sees only the instruction, and volunteers answer during live playback. An option-aware judge
scores their free-text answers.

Current runs open each response window one second before its reference
time. Keyed answers are matched to options, counting items require the exact
final count, and a Qwen3-14B judge \citep{yang2025qwen3} grades
free-text responses. We calibrated seven candidate judges on 954
question--answer triples. Gemini 3.5 Flash \citep{google2026gemini} labeled
each triple first, and annotators screened these labels and corrected
wrong ones. Greedy Qwen3-14B agrees with 89.7\% of the final labels, on par with
Qwen3-32B~\citep{yang2025qwen3} and gpt-oss-20b \citep{gptoss2025}. We chose it to balance
reliability and efficiency. One
decoding pass scores free-text answers and segments, see Appendix~\ref{app:judge} for all seven
candidates. Scores use the video clock, and every poll also records its
wall-clock time. On B200 hardware, JoyAI-VL~\citep{joyaivl2026}, Qwen3-VL-8B~\citep{bai2025qwen3vl}, Qwen3-VL-4B, and
LLaVA-OV2~\citep{llavaov2} keep pace with a live stream and finish at least 94.8\% of polls
within the one-second interval. Gemma-3n~\citep{google2025gemma3n}, MiniCPM-o~\citep{minicpmo2026}, Qwen2.5-Omni~\citep{xu2025qwen25omni},
Flash-VStream~\citep{zhang2024flashvstream}, and the polled API models finish at most 6.0\%, see Appendix~\ref{app:latency}.

\subsection{Timely Speech Costs Silence}
\label{sec:main}\label{sec:e4}
\input{fig_tex/fig_delta_tasc}

As shown in Table~\ref{tab:fullset}, systems that time their replies well
rarely stay silent. Each family is one evaluation setting, and each row names
its model. Rows ascend by Overall within a family. The video clock waits
for every poll, and scores describe decisions rather than live deployment.
Replaying the polled runs at their measured latency lowers timing accuracy by
up to 41.5 points for the API models. Models that keep pace lose at most 7.6,
see Appendix~\ref{app:latency}.

\paragraph{Tasks separate answering from restraint.}
On \ivqa, Qwen3-VL-8B \citep{bai2025qwen3vl} reaches 75.3 content accuracy
and 96.6 timing accuracy, but only 3.0 silence compliance. LLaVA-OV2
\citep{llavaov2} shows the same imbalance, with 64.5 content, 98.9 timing,
and 3.0 silence. Both answer promptly but keep speaking. \lvm{} produces
a different profile. JoyAI-VL \citep{joyaivl2026} leads Qwen3-VL-8B in
content accuracy, 71.8 to 24.4, and in silence compliance, 73.0 to 66.2.
One aggregate score hides this gap between current and remembered evidence. On \toa{},
Qwen3-VL-8B reaches 79.6 timing accuracy with 0.0 silence
compliance, and on \ptr{}, 79.1 with 9.8. Reaching an event
window does not imply that the system waited for it or stopped afterward.
\cst{} also demands a correct count. Qwen3-VL-8B
pairs 75.3 timing accuracy with 8.6 content accuracy, and JoyAI-VL
58.7 with 3.2. On \lcg, polled Claude Sonnet 5 \citep{anthropic2026claude} reaches 84.1 content
accuracy but only 4.0 silence compliance. Free-form Overall uses segment
content, and only the silence diagnostic reveals that its commentary rarely pauses.

\paragraph{Systems fail in opposite directions.}
Frequent responses hit event windows, but each extra utterance lowers
silence compliance. It reaches its floor of zero on 72\% to 97\% of positive
items for polled models. Their repeated replies per event still range from
2.8 to 19.2, see Appendix~\ref{app:pertask}. The ranking
and the trade-off both hold when timing credit requires correct content and
when silence compliance takes its smooth form, see Appendix~\ref{app:scoring}. Among the six open turn-based models, the leading
one, Qwen3-VL-8B~\citep{bai2025qwen3vl}, reaches 77.8 timing accuracy but only 10.9
silence compliance. LLaVA-OV2~\citep{llavaov2} has a similar Overall of 34.5, with timing
accuracy 81.0 and silence compliance 16.9, and its predecessor LLaVA-OV
\citep{li2025llavaonevision} scores 25.6. Flash-VStream
\citep{zhang2024flashvstream} shows the same imbalance at a lower level,
with 50.2 timing accuracy and 8.2 silence compliance. Speaking less changes
the failure profile. JoyAI-VL~\citep{joyaivl2026} gives up some timing accuracy for more silence compliance,
and VideoLLM-online \citep{chen2024videollmonline} raises silence
compliance further at the cost of timing and content. Among the native streaming video LLMs we ran, VideoChat3
\citep{videochat3} leads, ahead of
MOSS-Video \citep{moss2026} and MMDuet2 \citep{wang2025mmduet2}, but rarely
answers Recall queries. LiveCC \citep{chen2025livecc} comments on every
clip and has almost no silence compliance on Look, Time, Alert, and Track.
The contributed OneStreamer-4B \citep{zeng2026onestreamer} reaches
36.7 Overall, the highest native streaming score, with 57.2 timing accuracy
and 37.9 silence compliance. The human reference reaches 53.6 Overall
with 74.7 silence compliance. Every configuration above it sees the
whole video or is an agent whose logged polls take about half a minute. \textbf{Neither frequent speech nor broad
restraint satisfies both obligations.}

\input{fig_tex/fig14_silence}
\paragraph{The trade-off survives other settings.}
Rescoring every current configuration with delay scales from 2 to 10
seconds keeps the Overall ranking at a Kendall $\tau$ \citep{kendall1938}
of at least 0.95 against the five-second default. Overall moves by at
most 3.6 points for any configuration. As shown in Figure~\ref{fig:delta}, timing accuracy moves by up
to 18.0 points, for the human reference between the 2 and 10 second scales, and the timing order holds. One-second shifts or one-second Gaussian jitter of every event timestamp after the query keep the
Overall ranking at a Kendall $\tau$ of at least 0.91, with free-text content held fixed. Deleting replies tests whether fewer utterances alone
improve selectivity. As shown in Figure~\ref{fig:tasc}, lowering the keep
probability from 1.0 to 0.1 raises silence compliance at the cost of missed
events for all four traced systems. Random deletion removes correct and
redundant replies alike and cannot raise both scores. The response threshold of
VideoLLM-online \citep{chen2024videollmonline} moves along the same
trade-off. Raising it from 0.5 to 0.9 lifts timing accuracy from 30.2 to
51.7 while silence compliance falls from 56.4 to 18.4. Deleting only the
premature, repeated, and negative-stream replies instead lifts silence
compliance to 100.0 at unchanged timing accuracy for the same four systems.
The agent items form one fixed subset that every agent run shares. On them, this oracle filter raises Overall by 32.3 points for Qwen3-VL-8B~\citep{bai2025qwen3vl} and 25.5 for JoyAI-VL~\citep{joyaivl2026},
without rewriting any retained reply, see Appendix~\ref{app:planned}. \textbf{Fewer replies help only when the system chooses which
ones to withhold, and none of the tested controls, random reply deletion or a
native response threshold, makes that selective choice.}

\subsection{Four Failures of Selectivity}
\label{sec:when}\label{sec:e9}\label{sec:e10}

As shown in Figure~\ref{fig:silence}, repetition and premature speech
dominate different systems. Polled with a 64-frame window, Qwen3-VL-8B
\citep{bai2025qwen3vl} produces 21.4 redundant emissions per reference
event. Its fresh prompt has no record of its previous reply,
and while the evidence stays visible, each poll can repeat the same answer.
MMDuet2 \citep{wang2025mmduet2} instead produces 7.5 premature emissions
per reference event, exceeding its redundant count. A familiar scene alone
does not establish the requested condition, and the system must keep
waiting until it holds. The same answer content can conceal
opposite failures of response control. In a saved Alert item with four
green-light events, both systems always choose the correct option.
Qwen3-VL-8B produces 104 emissions, including five premature replies, four
matched responses, and 95 repetitions. JoyAI-VL \citep{joyaivl2026} emits
three replies, one premature, one matched, and one repeated, and misses
three of the four events, see Appendix~\ref{app:cases} for the full traces of both systems.

The 69 negative streams contain no valid trigger, yet JoyAI-VL~\citep{joyaivl2026} emits on
89.9\% of them. Qwen3-VL-8B~\citep{bai2025qwen3vl} in offline grounding, with future access, emits on 5.8\%.
The blind reference raises false alarms on only 1.4\% of streams but
fails every suite. On these suites, LLaVA-OV2 \citep{llavaov2} passes
26.4\% and JoyAI-VL 22.6\%, while the polled Qwen3-VL models pass 5.7\% to
13.2\%. No open-weight system clears a third of them. The two checks capture
different failures, as Qwen3-VL-8B raises the fewest false alarms of the four
systems yet passes fewer suites than a scripted burst policy. Each suite
takes its event and near-miss moments from one video, so the scene
alone does not tell a system when to answer and when to stay silent. \textbf{Selective interaction requires waiting
for evidence, responding once, and rejecting both absent events and near
misses. None of the systems in Figure~\ref{fig:silence} manages all four at once on
Alert and Track streams.}

\subsection{Online Control Needs Direct Measurement}
\label{sec:why}\label{sec:e4blind}\label{sec:e4full}\label{sec:e11}\label{sec:e14}
\input{fig_tex/fig12_backbone}
\input{fig_tex/fig16_grounding}

\paragraph{Offline scores mispredict online behavior.}
As shown in Figure~\ref{fig:backbone}, offline answer accuracy, online
scores, and poll speed rank the six open turn-based models differently under the
same adapter. Qwen3-VL-4B \citep{bai2025qwen3vl} leads offline accuracy but
ranks fifth in online Overall. Gemma-3n \citep{google2025gemma3n}
ranks last offline, leaving 43.8\% of keyed items unanswered in grounding, yet
first in online timing accuracy. See Appendix~\ref{app:backbone} for all answer rates. \textbf{An offline content score alone
misses the evaluated system's timing and silence.} The blind Qwen3-VL-8B~\citep{bai2025qwen3vl} baseline
scores 15.7 Overall without frames and serves as a diagnostic reference,
not as a lower bound for the configurations that see them. Grounding also differs from online use. As shown in
Figure~\ref{fig:grounding}, grounding and online scores of eight
configurations on identical items correlate at 0.18 for content, 0.06
for timing, and $-0.08$ for silence. Per backbone, offline and online
keyed answers agree at a Cohen's $\kappa$ \citep{cohen1960kappa} of at most 0.40. Online decisions need direct
measurement on the stream clock.

\vspace{-10pt}

\subsection{The Response Mechanism Moves Silence}
\label{sec:agents}\label{sec:pillar3}\label{sec:e12}\label{sec:e13}\label{sec:e15}
\input{fig_tex/fig_mechanisms}
\input{fig_tex/fig_frontier}

\paragraph{Inputs and controllers barely help.}
As shown in Figure~\ref{fig:mechanisms}, the way a system decides when to
speak moves silence compliance more than any input setting. Under polling
with a fresh prompt, the group median reaches 64.0 timing accuracy
but only 17.2 silence compliance, and more frames or new hints leave this
profile in place. Across 0.5--4\,fps, the Overall and timing accuracy of Qwen3-VL-8B
\citep{bai2025qwen3vl} change by about two points. Widening its window
from 16 to 256 frames raises Overall and silence compliance by four to six
points, and the median poll then slows from 0.48 to 4.1 seconds. Two hint
rewrites, Hint 2 and Hint 3, move Overall by less than one point. The native interaction system supplies little restraint
by itself. On the agent items, JoyAI-VL \citep{joyaivl2026} reaches 25.9 silence compliance,
and with Qwen3-VL-8B inside its controller, silence compliance falls to
11.0, close to the 5.9 of the same model under polling. With Claude Opus 5
\citep{anthropic2026opus} inside, it rises to 33.2 only by leaving 41\%
of positive items unanswered. Inside the same controller, Qwen3-VL-4B~\citep{bai2025qwen3vl} reaches
9.6 and Claude Sonnet 5 \citep{anthropic2026claude} 23.0. \textbf{Within this controller, restraint therefore
tracks the swapped-in model.}

\vspace{-15pt}

\paragraph{Memory and foresight cost timing.}
Reply history, agent loops, and grounding all raise silence, and each pays
in timing, speed, or stability. Reply history keeps earlier frames and
replies in one conversation, whereas a fresh prompt starts anew each poll.
It raises silence compliance for all six open turn-based models, which
suggests that much repetition comes from forgetting the previous reply. Its
effect on Overall depends on the backbone, as three models gain and three
lose, and a decoding change confounds the MiniCPM-o
\citep{minicpmo2026} gain. Agent loops make restraint deliberate. On the agent
items, the Claude Sonnet 5~\citep{anthropic2026claude} agents reply about a quarter as often as the polled
model and roughly triple its silence compliance at some cost in timing. Yet a
logged agent poll takes about half a minute, far beyond the one-second
interval, see Appendix~\ref{app:latency}. Swapping the plugins for Claude
Code tools alone moves the Claude agent's Overall by 5.2 points, and under the
same plugins GPT-5.6 Sol \citep{openai2026gpt} scores twice as high as Gemini 3.5
Flash \citep{google2026gemini}. Grounding shows that foresight alone does not
buy timing. With the whole video, Qwen3-VL-8B~\citep{bai2025qwen3vl} raises its silence compliance
more than fivefold but halves its timing accuracy, and Claude Sonnet 5 shows
the same shift. Query responses stay nearly immediate, but later events are
hard to locate in a finished video without tools that inspect chosen frames.
Dispider \citep{qian2025dispider} earns almost no timing credit on them. With
such tools, the Claude Sonnet 5 agent in grounding is the only configuration
above 70 in both timing and silence. \textbf{An agentic system is evaluated
with its loop and tools, not by its model alone.}

\paragraph{Selective silence at live pace comes first.}
Figure~\ref{fig:frontier} places each configuration by its timing accuracy
and silence compliance. Among configurations that keep pace with the stream,
none exceeds 40 on both, so a system should first learn which responses to
withhold and then keep task state across responses, see Appendix~\ref{app:planned}.

%% file: fig_tex/tab11_fullset.tex
\begin{table}[!t]
\centering\fontsize{8}{9.7}\selectfont %
\renewcommand{\arraystretch}{1.0}
\setlength{\tabcolsep}{0.65pt}
\caption{\textbf{Systems trade timing for silence.} Acc is content accuracy, TA timing accuracy, and SC silence compliance, all higher-is-better. See Appendix~\ref{app:models}. $^\dagger$Evaluated with the released code and contributed by the model authors: Senqiao Yang (Mage-VL) and Xiangyu Zeng (OneStreamer).}
\label{tab:fullset}
\begin{tabular*}{\textwidth}{@{\extracolsep{\fill}}l@{\hspace{3pt}}c@{\hspace{3pt}}*{5}{r@{\hspace{1.2pt}}r@{\hspace{1.2pt}}r@{\hspace{3pt}}}r@{\hspace{1.2pt}}r@{\hspace{1.2pt}}r@{}}
\toprule
\multirow{2}{*}{\textbf{System}} & \multirow{2}{*}{\textbf{Overall}} & \multicolumn{3}{c}{\textbf{Look}} & \multicolumn{3}{c}{\textbf{Recall}} & \multicolumn{3}{c}{\textbf{Time}} & \multicolumn{3}{c}{\textbf{Alert}} & \multicolumn{3}{c}{\textbf{Track}} & \multicolumn{3}{c}{\textbf{Commentate}} \\
\cmidrule(lr){3-5}\cmidrule(lr){6-8}\cmidrule(lr){9-11}\cmidrule(lr){12-14}\cmidrule(lr){15-17}\cmidrule(lr){18-20}
 & & Acc & TA & SC & Acc & TA & SC & Acc & TA & SC & Acc & TA & SC & Acc & TA & SC & Acc & TA & SC \\
\midrule
\multicolumn{20}{l}{\textbf{References}}\\
Qwen3-VL-8B~\citep{bai2025qwen3vl}, blind & 15.7 & 21.7 & 77.5 & 3.6 & 22.4 & 95.5 & 69.1 & 0.0 & 0.7 & 0.0 & 6.2 & 6.3 & 66.7 & 0.0 & 0.0 & 100.0 & 1.0 & 25.4 & 15.5 \\
Human reference & 53.6 & 78.7 & 74.0 & 80.0 & 44.4 & 57.4 & 61.1 & 63.0 & 51.0 & 77.8 & 19.5 & 53.3 & 77.2 & 36.8 & 69.0 & 73.0 & 56.9 & 70.1 & 69.5 \\
\midrule
\multicolumn{20}{l}{\textbf{Native streaming}}\\
VideoLLM-online~\citep{chen2024videollmonline} & 14.5 & 2.4 & 94.2 & 88.5 & 6.4 & 81.9 & 87.8 & 1.6 & 8.7 & 3.8 & 3.7 & 7.7 & 18.1 & 6.8 & 5.6 & 52.0 & 0.0 & 7.1 & 64.8 \\
LiveCC~\citep{chen2025livecc} & 14.5 & 68.8 & 71.1 & 0.0 & 0.0 & 5.6 & 100.0 & 34.6 & 24.5 & 3.2 & 35.8 & 14.5 & 2.6 & 1.7 & 2.2 & 3.0 & 17.0 & 42.0 & 23.9 \\
MMDuet2~\citep{wang2025mmduet2} & 17.0 & 55.4 & 62.9 & 54.9 & 7.7 & 37.8 & 7.3 & 15.3 & 37.2 & 0.9 & 13.6 & 35.2 & 15.1 & 1.2 & 26.2 & 34.1 & 24.1 & 47.4 & 21.7 \\
MOSS-Video~\citep{moss2026} & 19.9 & 56.2 & 71.5 & 6.7 & 9.7 & 15.7 & 36.4 & 26.9 & 18.1 & 0.0 & 63.3 & 25.3 & 10.2 & 3.3 & 4.6 & 3.4 & 14.2 & 27.7 & 50.1 \\
AsynKV~\citep{spotbench2026} & 27.6 & 66.9 & 59.7 & 89.2 & 12.8 & 11.9 & 85.7 & 31.0 & 20.6 & 71.9 & 29.9 & 17.2 & 68.6 & 0.0 & 0.0 & 100.0 & 1.9 & 3.1 & 74.5 \\
VideoChat3~\citep{videochat3} & 28.5 & 62.5 & 56.1 & 87.1 & 6.5 & 3.0 & 100.0 & 65.4 & 11.2 & 10.4 & 72.5 & 9.2 & 21.2 & 25.0 & 11.3 & 28.8 & 18.5 & 22.6 & 61.0 \\
OneStreamer-4B~\citep{zeng2026onestreamer}$^\dagger$ & 36.7 & 77.7 & 78.8 & 49.4 & 47.4 & 76.9 & 87.9 & 50.0 & 39.5 & 15.3 & 79.0 & 54.8 & 19.1 & 10.5 & 46.3 & 27.6 & 18.5 & 38.1 & 60.4 \\
\midrule
\multicolumn{20}{l}{\textbf{Turn-based polling}}\\
Flash-VStream~\citep{zhang2024flashvstream} & 14.3 & 30.7 & 66.2 & 3.1 & 2.6 & 18.0 & 13.9 & 21.4 & 53.1 & 0.0 & 42.7 & 61.3 & 10.7 & 7.4 & 49.4 & 15.1 & 2.2 & 41.6 & 2.7 \\
Qwen2.5-Omni~\citep{xu2025qwen25omni} & 18.7 & 23.5 & 25.3 & 41.5 & 3.2 & 9.6 & 66.7 & 42.9 & 40.0 & 9.0 & 56.0 & 51.9 & 9.4 & 4.9 & 20.3 & 42.6 & 19.4 & 54.2 & 9.8 \\
LLaVA-OV~\citep{li2025llavaonevision} & 25.6 & 37.5 & 55.0 & 18.8 & 22.6 & 48.4 & 86.7 & 92.3 & 57.5 & 7.7 & 39.3 & 16.7 & 23.7 & 8.3 & 6.3 & 8.3 & 28.5 & 86.5 & 0.0 \\
Qwen3-VL-4B~\citep{bai2025qwen3vl} & 26.2 & 71.1 & 74.6 & 2.7 & 21.2 & 23.2 & 63.4 & 15.1 & 7.4 & 27.3 & 51.5 & 41.7 & 29.4 & 1.9 & 2.7 & 87.1 & 31.0 & 46.2 & 22.5 \\
Gemma-3n~\citep{google2025gemma3n} & 27.8 & 57.2 & 98.1 & 3.0 & 28.8 & 79.9 & 67.2 & 49.1 & 57.6 & 5.6 & 66.6 & 80.3 & 4.6 & 7.4 & 78.1 & 4.5 & 30.7 & 97.8 & 0.0 \\
MiniCPM-o~\citep{minicpmo2026} & 29.0 & 75.3 & 87.2 & 3.8 & 44.2 & 86.9 & 68.6 & 15.9 & 5.6 & 16.0 & 15.9 & 12.9 & 51.7 & 8.0 & 20.5 & 36.5 & 53.8 & 81.3 & 3.4 \\
Mage-VL-4B~\citep{yang2026mage}$^\dagger$ & 33.7 & 68.1 & 98.9 & 3.0 & 35.3 & 72.6 & 65.2 & 51.6 & 53.6 & 4.9 & 65.5 & 66.8 & 16.0 & 7.4 & 20.2 & 43.9 & 37.3 & 82.3 & 4.0 \\
LLaVA-OV2~\citep{llavaov2} & 34.5 & 64.5 & 98.9 & 3.0 & 34.6 & 90.5 & 68.3 & 57.1 & 76.3 & 0.8 & 72.3 & 75.3 & 8.7 & 10.2 & 55.6 & 23.2 & 43.0 & 98.2 & 0.0 \\
Qwen3-VL-8B & 34.8 & 75.3 & 96.6 & 3.0 & 24.4 & 44.1 & 66.2 & 66.7 & 79.6 & 0.0 & 78.7 & 79.1 & 9.8 & 8.6 & 75.3 & 11.7 & 57.1 & 96.3 & 0.0 \\
Gemini 2.5 Flash~\citep{google2026gemini} & 35.5 & 81.2 & 100.0 & 0.0 & 40.0 & 93.3 & 71.4 & 46.2 & 97.4 & 0.0 & 71.3 & 89.4 & 0.0 & 17.6 & 85.0 & 4.5 & 68.1 & 98.6 & 0.0 \\
Claude Sonnet 5~\citep{anthropic2026claude} & 44.6 & 87.5 & 98.8 & 0.0 & 46.7 & 100.0 & 73.3 & 84.6 & 84.4 & 0.0 & 76.0 & 78.7 & 11.1 & 18.5 & 52.1 & 15.9 & 84.1 & 89.0 & 4.0 \\
\midrule
\multicolumn{20}{l}{\textbf{Offline temporal grounding}}\\
Gemma-3n & 16.1 & 14.5 & 24.0 & 96.4 & 14.1 & 23.7 & 100.0 & 7.1 & 6.2 & 15.9 & 15.0 & 7.7 & 39.2 & 6.8 & 11.0 & 49.4 & 4.2 & 17.9 & 57.9 \\
Dispider~\citep{qian2025dispider} & 24.8 & 75.0 & 93.8 & 93.8 & 61.3 & 100.0 & 100.0 & 38.5 & 0.0 & 52.6 & 18.5 & 0.8 & 62.7 & 5.0 & 1.2 & 74.6 & 1.6 & 0.9 & 95.5 \\
Qwen2.5-Omni & 27.2 & 44.0 & 69.9 & 100.0 & 47.4 & 78.8 & 100.0 & 5.6 & 2.8 & 39.8 & 11.7 & 2.0 & 59.3 & 0.6 & 1.2 & 88.4 & 0.0 & 0.0 & 8.3 \\
MiniCPM-o & 30.6 & 59.6 & 95.7 & 96.6 & 66.7 & 100.0 & 100.0 & 8.7 & 8.4 & 6.5 & 11.4 & 3.4 & 36.4 & 8.0 & 11.9 & 32.6 & 2.6 & 8.8 & 46.3 \\
Qwen3-VL-4B & 33.8 & 63.9 & 96.3 & 96.5 & 65.4 & 99.4 & 100.0 & 7.1 & 4.5 & 9.3 & 15.3 & 3.5 & 35.8 & 3.1 & 9.5 & 46.6 & 0.8 & 11.9 & 35.3 \\
LLaVA-OV2 & 33.9 & 64.5 & 96.3 & 97.6 & 60.9 & 99.4 & 100.0 & 7.1 & 5.8 & 19.4 & 11.0 & 4.3 & 38.6 & 7.4 & 8.3 & 63.1 & 3.8 & 5.7 & 79.7 \\
Qwen3-VL-8B & 35.7 & 63.9 & 96.9 & 97.9 & 67.9 & 100.0 & 100.0 & 6.3 & 2.1 & 11.7 & 13.1 & 4.0 & 40.9 & 10.5 & 13.4 & 48.2 & 4.7 & 19.1 & 50.9 \\
Claude Sonnet 5 & 38.6 & 69.3 & 89.7 & 97.9 & 65.4 & 87.8 & 100.0 & 29.4 & 10.5 & 85.5 & 35.7 & 5.6 & 81.0 & 0.0 & 0.8 & 100.0 & 5.1 & 6.5 & 79.8 \\
GPT-5.6 Sol~\citep{openai2026gpt} agent & 58.8 & 89.2 & 95.8 & 96.4 & 91.7 & 100.0 & 100.0 & 57.9 & 48.4 & 65.9 & 52.3 & 39.2 & 79.3 & 21.7 & 33.2 & 72.8 & 18.7 & 38.9 & 55.5 \\
Claude Sonnet 5 agent & 66.1 & 90.4 & 97.4 & 97.6 & 92.3 & 100.0 & 100.0 & 65.9 & 59.7 & 76.6 & 57.2 & 50.0 & 81.5 & 35.1 & 50.9 & 83.9 & 29.2 & 80.7 & 45.1 \\
\midrule
\multicolumn{20}{l}{\textbf{Native Real-time Interaction System (JoyAI-VL~\citep{joyaivl2026})}}\\
Claude Opus 5~\citep{anthropic2026opus} & 30.5 & 81.2 & 85.0 & 13.4 & 20.0 & 26.7 & 100.0 & 23.1 & 20.6 & 42.9 & 45.6 & 34.9 & 43.8 & 17.3 & 28.2 & 24.1 & 48.6 & 46.3 & 15.0 \\
Qwen3-VL-8B & 32.4 & 75.0 & 98.8 & 0.0 & 73.3 & 80.0 & 66.7 & 7.7 & 90.5 & 0.0 & 66.2 & 97.9 & 3.7 & 0.0 & 76.2 & 8.4 & 70.0 & 97.8 & 0.0 \\
JoyAI-VL & 33.8 & 59.0 & 96.5 & 15.7 & 71.8 & 97.4 & 73.0 & 41.4 & 55.8 & 10.3 & 52.7 & 38.5 & 29.1 & 3.2 & 58.7 & 22.4 & 38.4 & 73.7 & 21.5 \\
Qwen3-VL-4B & 35.5 & 68.8 & 68.2 & 0.0 & 38.7 & 60.0 & 94.7 & 69.2 & 53.8 & 0.0 & 85.1 & 65.0 & 1.2 & 3.3 & 7.9 & 0.0 & 37.6 & 77.4 & 0.8 \\
Claude Sonnet 5 & 43.3 & 79.7 & 94.4 & 7.3 & 84.1 & 91.5 & 66.7 & 70.0 & 70.0 & 7.1 & 63.4 & 64.1 & 37.0 & 19.6 & 60.4 & 10.4 & 55.8 & 84.0 & 6.3 \\
\midrule
\multicolumn{20}{l}{\textbf{Agentic Interaction System (Qwen-MM-Plugins~\citep{qwenmmplugins2026})}}\\
Gemini 3.5 Flash~\citep{google2026gemini} & 29.4 & 62.5 & 48.8 & 80.0 & 46.7 & 44.0 & 87.5 & 30.8 & 23.1 & 40.0 & 20.8 & 12.0 & 62.7 & 0.0 & 4.8 & 75.7 & 12.0 & 10.7 & 32.1 \\
Claude Sonnet 5 & 56.0 & 87.5 & 93.8 & 68.8 & 86.7 & 93.3 & 100.0 & 50.0 & 31.7 & 54.2 & 49.4 & 55.8 & 47.5 & 8.0 & 46.0 & 47.4 & 69.2 & 89.6 & 15.0 \\
GPT-5.6 Sol & 59.1 & 93.8 & 92.5 & 93.8 & 86.7 & 93.3 & 100.0 & 30.8 & 28.5 & 38.5 & 59.9 & 50.7 & 69.6 & 0.3 & 70.4 & 30.7 & 69.4 & 87.9 & 16.5 \\
\midrule
\multicolumn{20}{l}{\textbf{Agentic Interaction System (Claude Code tools~\citep{anthropic2026claudecode})}}\\
Claude Sonnet 5 & 61.2 & 93.8 & 90.0 & 62.5 & 93.3 & 100.0 & 86.7 & 53.8 & 54.9 & 58.3 & 69.4 & 56.3 & 50.9 & 14.3 & 53.1 & 54.5 & 77.9 & 83.7 & 8.6 \\
\bottomrule
\end{tabular*}
\end{table}

%% file: fig_tex/fig_delta_tasc.tex
\begin{figure}[tp]
\centering
\begin{minipage}[t]{0.491\linewidth}
\centering
\includegraphics[width=\linewidth]{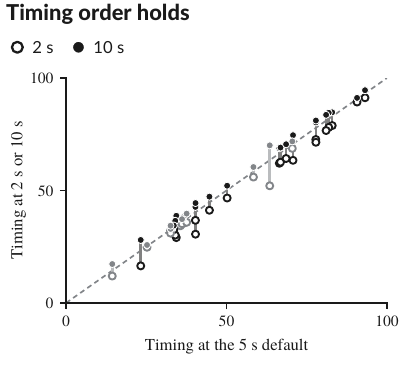}
\caption{\textbf{Timing order holds.} Each vertical line is one configuration at its timing accuracy under the default 5\,s delay scale, running from its 2\,s score (open) to its 10\,s score (filled). References are gray.}
\label{fig:delta}
\end{minipage}\hfill
\begin{minipage}[t]{0.491\linewidth}
\centering
\includegraphics[width=\linewidth]{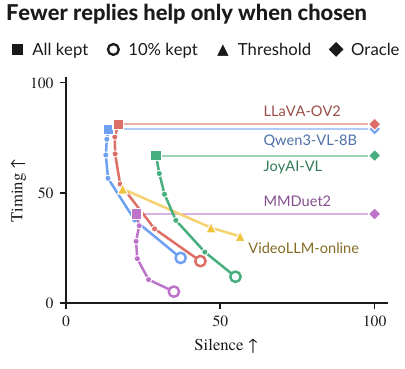}
\caption{\textbf{Fewer replies help only when chosen.} Deleting replies at random, from all (squares) down to 10\% (circles), trades timing for silence, as do VideoLLM-online~\citep{chen2024videollmonline} thresholds on the agent items. The oracle deletes only violations and keeps timing.}
\label{fig:tasc}
\end{minipage}
\end{figure}

%% file: fig_tex/fig14_silence.tex
\begin{figure}[tp]
\centering
\includegraphics[width=\linewidth]{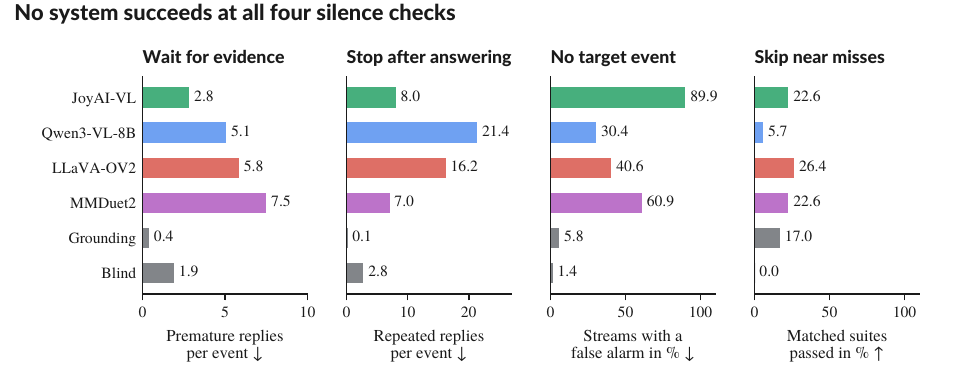}
\caption{\textbf{No system passes all four silence checks.} Qwen3-VL-8B \citep{bai2025qwen3vl} is polled with a 64-frame window, and the gray rows are its grounding run and the blind reference.}
\label{fig:silence}
\end{figure}

%% file: fig_tex/fig12_backbone.tex
\begin{figure}[tp]
\centering
\includegraphics[width=\linewidth]{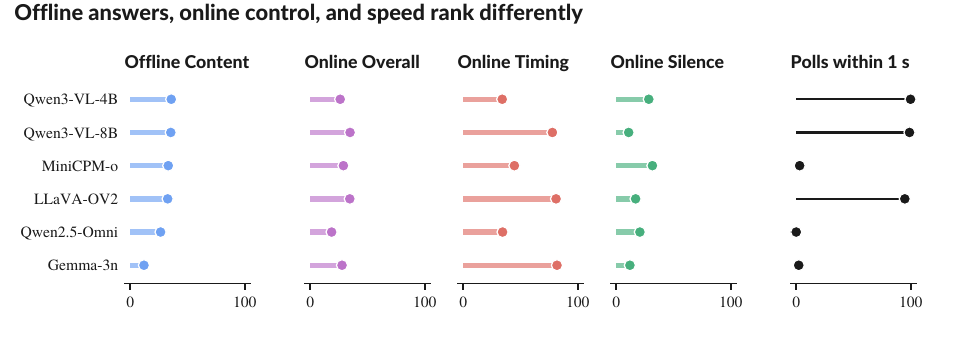}
\caption{\textbf{Offline content, online scores, and speed disagree.} Six backbones share one polling adapter, and offline content counts unanswered keyed items as wrong.}
\label{fig:backbone}
\end{figure}

%% file: fig_tex/fig16_grounding.tex
\begin{figure}[tp]
\centering
\includegraphics[width=\linewidth]{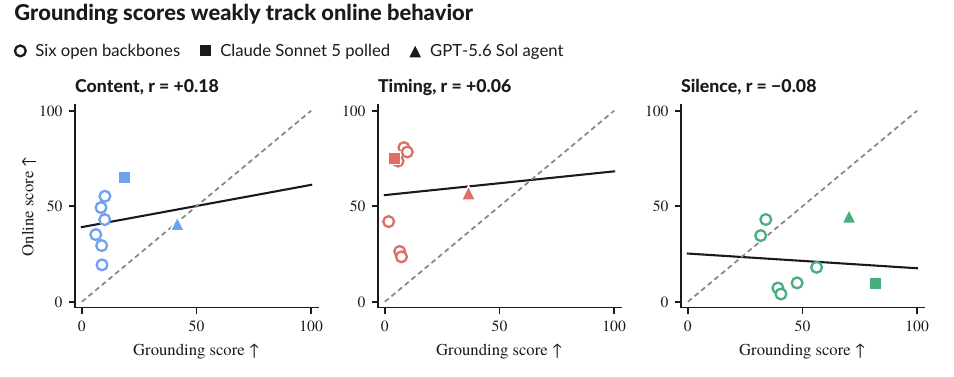}
\caption{\textbf{Grounding and online scores diverge.} Both scores use identical Time, Alert, Track, and Commentate items. Solid lines are least-squares fits, and r is the Pearson correlation \citep{pearson1895}.}
\label{fig:grounding}
\end{figure}

%% file: fig_tex/fig_mechanisms.tex
\begin{figure}[tp]
\centering
\includegraphics[width=\linewidth]{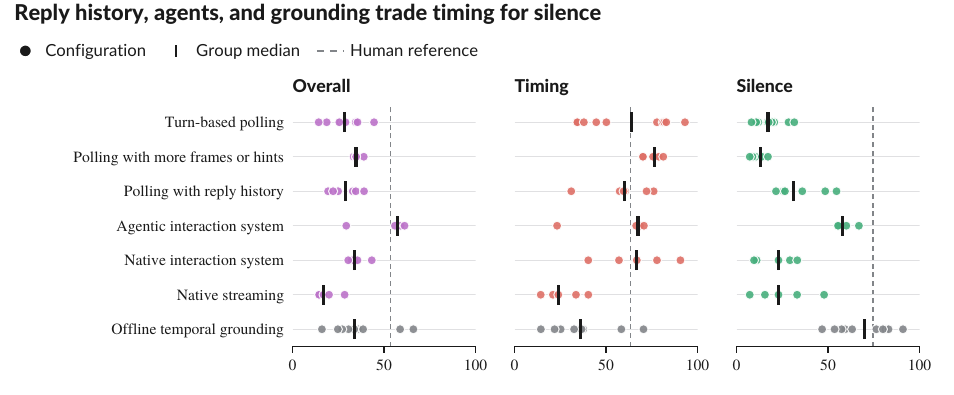}
\caption{\textbf{Reply history, agents, and grounding trade timing for silence.} Each dot is one configuration, grouped by how the system decides when to speak, and ticks mark group medians.}
\label{fig:mechanisms}
\end{figure}

%% file: fig_tex/fig_frontier.tex
\begin{figure}[tp]
\centering
\includegraphics[width=\linewidth]{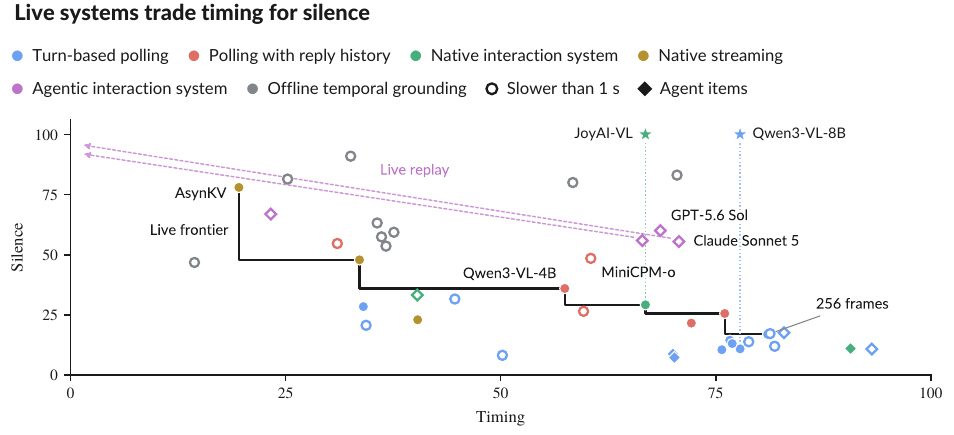}
\caption{\textbf{Live systems trade timing for silence.} Filled markers keep pace with the stream, and the line is their frontier. Stars apply oracle restraint, and arrows replay two agents at logged latency.}
\label{fig:frontier}
\end{figure}

%% file: tex/4_conclusion.tex
\section{Conclusion and Discussion}
\label{sec:conclusion}

\vigil{} evaluates whether a video assistant speaks when its instruction
warrants a response and stays silent otherwise. High timing scores coexist
with frequent unnecessary speech, restrained output misses events, and both
behaviors belong to the complete system. The tested native controller adds little restraint by itself. Reply history and agentic
systems add it, and logged agent polls take about half a minute, far slower
than the one-second poll interval.
Future systems should retain an instruction, recognize new evidence,
respond once, and resume monitoring. Progress reports should cover both
obligations on matched items, speech when due and silence otherwise, with
runtime measured separately. On the Alert and Track streams that carry
them, negative streams and near-miss suites separate systems whose aggregate
scores look similar.

%% file: tex/5_statements.tex
\subsection*{AI use statement}
Generative models served as the open judge for free-text items and as
the first labeler of the judge-calibration triples before human screening
and correction. They labeled content domains for the statistics and
served as the blind auditor of the keyed pool. They also assisted with evaluation, analysis, and figure
code, with prose editing, and with checking bibliographic metadata against
arXiv and publisher records. Dataset annotation remains a human responsibility. This draft uses AI-assisted
editing and figure generation, and the authors verify the final text,
analyses, and submission.

\subsection*{Ethics statement}
All videos are public and referenced by id, faces are not annotated, and
any video is removed from the release at the request of its uploader. Two
domains carry risk. A trigger such as ``tell me when a person appears''
can serve surveillance. The first-aid and driving items can invite
over-reliance on a system that this paper shows to be far from reliable.
The human reference was collected from volunteers who consented to the
task and could withdraw at any time.

\subsection*{Reproducibility statement}
Appendices~\ref{app:release} and~\ref{app:scoring} document the release format
and scoring rules. Appendices~\ref{app:models}, \ref{app:latency}, and~\ref{app:agents}
record configurations, run dates, and the measured latency of every run that
logged it.

%% file: tex/6_appendix.tex
\clearpage
\beginsupplement
\crefalias{section}{appendix} %

\phantomsection
\addcontentsline{toc}{section}{Appendix}
\addtocontents{toc}{\protect\setcounter{tocdepth}{-1}}
\section*{Appendix}
\noindent This appendix holds the related work, the data and evaluation details,
the full tables behind the analysis figures, and the limitations.
\begin{itemize}
  \item Appendix~\ref{sec:related} covers related streaming video systems and benchmarks.
  \item Appendix~\ref{app:bench} groups the six tasks by response obligation and gives their released labels.
  \item Appendix~\ref{app:stats} gives counts by task, event spans, video durations, upload years, and domains.
  \item Appendix~\ref{app:curation} describes collection, screening, annotation with its six stages and the count kept after each, and negative streams.
  \item Appendix~\ref{app:tasks} gives an example of each task.
  \item Appendix~\ref{app:release} lists the released files and what a reproduction must retain.
  \item Appendix~\ref{app:eval} defines the near-miss suites and their pass rate.
  \item Appendix~\ref{app:scoring} specifies matching and scoring, content-gated timing and smooth silence, alternative aggregations, and timing tolerance.
  \item Appendix~\ref{app:judge} reports judge calibration, rescoring with other judges, and free-form segment scoring.
  \item Appendix~\ref{app:prompts} reproduces the evaluation prompts.
  \item Appendix~\ref{app:models} records the configuration and provenance of every system, including AsynKV~\citep{spotbench2026}, and the human reference.
  \item Appendix~\ref{app:latency} reports wall-clock latency per poll, the cost of one evaluation, and live replays.
  \item Appendix~\ref{app:backbone} gives the offline accuracy and answer rates of the polled backbones.
  \item Appendix~\ref{sec:gamed} audits the protocol with scripted policies.
  \item Appendix~\ref{app:pertask} gives aggregate and per-task scores, violations per event, and ablations.
  \item Appendix~\ref{app:sweep} rescores streams after random emission thinning.
  \item Appendix~\ref{app:cases} shows the source videos and full traces of the case studies.
  \item Appendix~\ref{app:agents} details the tools, isolation, and runs of the agentic systems.
  \item Appendix~\ref{app:planned} covers oracle restraint, the blind reference, reply history, what helps at live pace, and the native response threshold.
  \item Appendix~\ref{app:errors} relates errors to domain and video duration and shows verified failure cases.
  \item Appendix~\ref{app:scene} relates errors to scene clutter and target difficulty.
  \item Appendix~\ref{app:limitations} states limitations and future directions.
\end{itemize}

\input{tex/appendix/related}
\input{tex/appendix/details}

%% file: tex/appendix/related.tex
\section{Related Work}
\label{sec:related}

\paragraph{Streaming video systems.}
VideoLLM-online, Flash-VStream, and MMDuet model responses to incoming
video rather than a completed clip
\citep{chen2024videollmonline,zhang2024flashvstream,wang2024mmduet}.
Dispider separates perception, decision, and reaction
\citep{qian2025dispider}, and StreamChat adds a hierarchical memory for
multi-round interaction \citep{streamingmemory}. StreamBridge attaches a
light activation model that decides when an offline video model should
reply \citep{streambridge2025}, and LiveStar chooses between a response and
silence in one verification pass \citep{livestar2025}. JoyAI-VL wraps its
model in an interaction system with memory and turn management \citep{joyaivl2026}.
These designs place response control in different components. \vigil{}
evaluates their emitted behavior under a common instruction and video
clock, treating the controller and memory as part of the system.

\paragraph{Offline and streaming understanding.}
\input{fig_tex/tab_benchmarks_qa}
As shown in Table~\ref{tab:benchmarks_qa}, no streaming question-answering
benchmark includes negative streams, and only
OVO-Bench \citep{li2025ovobench} tests near misses.
Video-MME and MLVU emphasize answers about completed videos
\citep{fu2024videomme,zhou2024mlvu}, and E.T. Bench scores event-level
timestamps offline \citep{liu2024etbench}. StreamingBench and OVO-Bench
introduce streaming settings and time-sensitive questions
\citep{lin2024streamingbench}, and SVBench chains temporally
linked questions over multi-turn dialogues \citep{yang2025svbench}.

\paragraph{Proactive and interactive evaluation.}
OmniMMI adds proactive alerting and a turn-taking task that credits
silence on irrelevant turns within a stream rather than on whole streams
\citep{omnimmi2025}. RIVER covers memory, live
perception, and anticipation \citep{river2026}. ProactiveVideoQA scores
free-running responses with a temporal metric \citep{proactivevideoqa2025},
and OmniPro and OmniInteract extend proactive evaluation to audiovisual
streams, response slots, and interruption handling
\citep{omnipro2026,omniinteract2026}. ESTP-Bench and ProAssist evaluate
egocentric proactive assistance and penalize replies outside answer
windows \citep{estpbench,proassist}. IPIBench scores per-second triggering
within a local window around each event and proposes IPI-Agent, an
agent with temporal gating \citep{ipibench2026}, and
SPOT-Bench penalizes premature, late, and redundant outputs
\citep{spotbench2026}. LiveProBench scores per-second decisions under
standing requests \citep{liveprobench}. Safety benchmarks also test restraint.
PaSBench-Video includes safe videos whose correct output is silence
\citep{pasbench2026}, and EgoSafetyBench contrasts scenes that differ in
one critical cue \citep{egosafetybench2026}.

\paragraph{Positioning.}
\vigil{} brings these elements together for general assistant
instructions. It scores content, timing, and silence together on the video
clock. It also includes whole streams in which the requested event never
occurs and pairs counted events with near misses from the same video. It also
compares native systems, polling adapters, and agents with measured
wall-clock latency.

%% file: fig_tex/tab_benchmarks_qa.tex
\begin{table}[tp]
\centering\fontsize{7}{8.7}\selectfont
\setlength{\tabcolsep}{2pt}
\caption{\textbf{Streaming question-answering benchmarks.} Columns follow Table~\ref{tab:benchmarks}, and every mark is a binary yes or no.}
\label{tab:benchmarks_qa}
\begin{tabular*}{\textwidth}{@{\extracolsep{\fill}}lrrccccccc@{}}
\toprule
& \multicolumn{5}{c}{\textbf{Dataset}} & \multicolumn{4}{c}{\textbf{Evaluation}} \\
\cmidrule(lr){2-6}\cmidrule(lr){7-10}
\textbf{Benchmark} & \textbf{Videos} & \textbf{Items} & \textbf{Multi-response} & \textbf{Negatives} & \textbf{Near misses} & \textbf{Proactive} & \textbf{Premature} & \textbf{Graded delay} & \textbf{Redundant} \\
\midrule
StreamingBench~\citep{lin2024streamingbench} & 900 & 4,500 & \xmark & \xmark & \xmark & \cmark & \cmark & \xmark & \xmark \\
OVO-Bench~\citep{li2025ovobench} & 644 & 2,814 & \xmark & \xmark & \cmark & \cmark & \cmark & \cmark & \xmark \\
SVBench~\citep{yang2025svbench} & 1,353 & 49,979 & \xmark & \xmark & \xmark & \xmark & \xmark & \xmark & \xmark \\
OmniMMI~\citep{omnimmi2025} & 1,121 & 2,290 & \xmark & \xmark & \xmark & \cmark & \cmark & \xmark & \xmark \\
\midrule
\textbf{\vigil{}} & 812 & 1,060 & \cmark & \cmark & \cmark & \cmark & \cmark & \cmark & \cmark \\
\bottomrule
\end{tabular*}
\end{table}

%% file: tex/appendix/details.tex
\section{Task Taxonomy}
\label{app:bench}
\input{fig_tex/tab2_taxonomy}
Table~\ref{tab:taxonomy} groups the six tasks by the response obligation
their names describe. The released labels
remain IVQA for Look, LVM for Recall, TOA for Time, PTR for Alert, CST for
Track, and LCG for Commentate. Causal reasoning and belief revision are
sub-labels, not additional headline tasks. Silence is a valid system
action across all six categories. It is the only valid action on negative
items whose requested condition never occurs.
Annotation counts and evaluator groupings serve different purposes.
The annotation taxonomy places seven causal items under Recall, giving
159 Look and 163 Recall items. The current score export merges these
seven items into IVQA, giving 166 Look and 156 Recall items in per-task
result tables. Both conventions include the same 1{,}060 interactions.
One revision item is included in Track, giving 192 items. Renaming a
display category does not relabel or rescore any item.

\section{Dataset Coverage}
\label{app:stats}\label{app:counts}\label{app:domains}
\input{fig_tex/fig9_extstats}
\input{fig_tex/tab6_domains}
The corpus contains 812 videos and 1{,}060 interactions. There are 159
Look, 163 Recall, 126 Time, 313 Alert, 192 Track, and 107 Commentate items
under the annotation taxonomy. Questions account for 626 interactions
and instructions for 434. The 991 positive items contain 3{,}693 reference
events. Their event spans are under one minute for 690 items, one to two
minutes for 154, two to five minutes for 138, and over five minutes for 9.
Video durations range from 15 to 941 seconds, with a mean of 159 seconds.
As shown in Figure~\ref{fig:extstats}, 309 of the 597 videos with upload dates
were uploaded in 2025 or later. Domain labels use title and description
metadata, followed by author review. As shown in Table~\ref{tab:domains}, sports and exercise
coaching is the largest domain, with 22.3\% of the videos.

\section{Curation and Annotation}
\label{app:curation}\label{app:rstar}\label{app:contam}
Annotators collect procedural and monitoring videos and screen them for
observable response conditions. Automatic filters examine duration,
resolution, views, and upload dates. Human screening excludes unsuitable
clips, compilations that reset the timeline, and overlays that reveal the
answer.
Eight graduate students annotated the benchmark in six stages, and
Figure~\ref{fig:annotation_pipeline} gives the count kept after each stage,
and Figure~\ref{fig:annotation_interface} shows the annotation interface.
Retrieval runs search queries over a grid of domains and tasks, takes at
most eight videos per query, and removes duplicates. The automatic filters
target clips of 20\,s to five minutes, a range that holds 94.7\% of the
final videos. Retrieval and the filters yield 4{,}851 candidate videos, and
human screening keeps 1{,}089 of them. In
annotation, the annotators cross-annotate the same batch of videos, write
the instruction, the reference answers, and the earliest answerable time,
and keep the harder questions. 276 videos yield no usable interaction, which
leaves 813 videos with 1{,}061 interactions. A language model then drafts the
distractors of the keyed questions, up to five per question. Each distractor
is a semantic neighbor of the answer, another object or action in the same
scene, a state that holds before the event, a neighboring number, or a wrong
order. Three language models then answer the questions without frames. The annotators review the 230 questions that these models
answer correctly and revise them where needed. In counting videos, we mine
moments that look like the counted event but do not satisfy the condition,
check the candidates from 86 videos frame by frame, and keep 53 suites with
106 near misses. Finally, the annotators cross-check the final items, in
particular the items that every system answers wrong, and remove one
interaction, which leaves 812 videos and 1{,}060 interactions.
For each positive interaction, the annotation specifies its instruction
or question, query time, task label, response format, and reference events.
Each event has content and an earliest answerable timestamp. An online
solvability check asks whether the prefix contains enough evidence at
that timestamp and whether earlier prefixes do not. A held-out annotator
checks the decisive visual evidence without the target timestamp.
Appendix~\ref{app:scoring} tests timestamp noise under Timing tolerance.

\input{fig_tex/fig_annotation_pipeline}
\input{fig_tex/fig_annotation_interface}
\label{app:negatives}
The 69 negative interactions include 39 Alert and 30 Track items.
Their instruction predicates remain false throughout the video.
A text-only audit screens keyed questions for language-only solutions.
Premature-state distractors distinguish the requested event from an
earlier scene.

\section{Task Examples}
\label{app:tasks}
Look and Recall answer a query. Look asks which lane a swimmer occupies
when the starting list is visible, and Recall asks about a distance
demonstrated earlier in a driving lesson. Both require an answer at the
query, drawing on current or historical visual evidence, and neither
licenses repeated copies of it. Time and Alert wait for
evidence. Time includes deciding which of two windows opens first, and the
response becomes due when the required order can be established. Alert
asks the system to report when a traffic light turns green, and a standing
instruction can require several such reports, separated by intervals of
silence. Track and Commentate update as needed. Track maintains a running
count of appearances or revises a state when new evidence arrives, and
Commentate reports successive on-screen headings or procedural steps. The
requested update frequency follows the reference events or segments rather
than an instruction to speak at every frame.

\section{Release Format}
\label{app:release}
The full collection is the evaluation set, and there is no training split.
Annotations reside in
\texttt{results/<channel>/<video\_id>/annotation.json}.
Each file records the video duration and interactions, including the query,
query time, task and time types, negative flag, and timestamped references.
Video files, metadata, and multiple-choice keys accompany the annotations.
Metadata records source identifiers, URLs, upload dates when available,
and decoding properties.
Predictions contain one JSON line per interaction, identified by video
and item index. Each prediction lists emissions as a stream timestamp and
text. Optional fields record poll counts and latencies. A missing prediction
line is treated as an empty emission list. The evaluator produces per-item
scores and aggregates. Reproductions must retain the item list, scoring
version, judge configuration, and input-access policy with each result.

\section{Matched Events and Near Misses}
\label{app:eval}\label{sec:accuracy}
There are 53 suites from Track counting streams. Each contains a positive
item and near-miss moments from the same video, 106 timestamps in total.
For suite $j$, let $h_j=1$ if its positive item receives nonzero timing
credit and zero otherwise. Let $s_j=1$ if the system emits nothing within
five seconds of every near miss and zero otherwise. The suite pass rate is
\begin{equation}
\pairsel=\frac{100}{53}\sum_{j=1}^{53}h_js_j.
\end{equation}
The positive criterion is timing credit, and content accuracy is reported
alongside this diagnostic. Ten near-miss windows
intersect a positive response window. Excluding emissions in the overlap
changes individual scores by at most eight points in the reported check.
Scripted policies set a floor. The burst policy passes 22.6\% of the suites,
as many as JoyAI-VL \citep{joyaivl2026} and MMDuet2 \citep{wang2025mmduet2},
while always-fire and never-fire pass none. Offline grounding by the Claude
Sonnet 5 \citep{anthropic2026claude} and GPT-5.6 Sol \citep{openai2026gpt} agents
passes 60.4\% and 54.7\%.

\section{Scoring Specification}
\label{app:scoring}
\input{fig_tex/tab8_families}
Table~\ref{tab:families} lists how each response format contributes to
Overall. For each item, the evaluator removes empty emissions and sorts the rest
by timestamp. Negative items receive 100 only if the list is empty.
For event-based positive items, each window starts one second before its
reference timestamp and ends where the next window starts. The last
window extends to the end of the video. The first emission within a
window is matched regardless of content, with nonnegative delay measured
from the reference timestamp. Earlier speech is premature and subsequent
speech in an occupied window is redundant. Matching ignores content.
Matching by position keeps
timing and silence independent of the judge, and content accuracy then
grades the matched reply. A content-aware match would reward
emitting candidate replies until one fits.

Missing event responses contribute zero timing credit. Content accuracy
uses the first post-query response for query-answer items and the
mean over matched responses for triggers. Keyed answers first match an
option letter in the first 40 characters. Without a usable letter, the
scorer compares token F1 with every option, requiring a unique best match
of at least 0.5. Unique full-option containment is the final fallback. The
keyed pool has 688 items, including 637 six-option questions. Free-text
answers use the open judge. Lexical fallback takes the maximum of exact
match, token F1, and a length-guarded numeric match. Human free-text
responses are judged against the keyed answer text. Counting accuracy
compares the last number in the emitted sequence with the final reference
count. This check uses the final count. A per-event variant also
requires each matched reply to state the running count of its event. It
stays at or below 38.5 for every system, against 40.0 for the human
reference, and exceeds final-count accuracy for 12 of the 20 configurations
scored on all 160 counting items, for example 26.2 against 8.8 for polled
Qwen3-VL-8B \citep{bai2025qwen3vl}. The final count is therefore the stricter
criterion for most configurations. The violation budget equals the number of reference
events. A fully silent positive item has zero accuracy and timing. Its
silence component is undefined and its Overall is zero.

Free-form items use reference timestamps to partition the video. The
scorer concatenates emissions within each segment and compares them
with its reference. Empty segments score zero. The item averages these
segment scores, and this average is its Overall. Timing and silence are
reported only as diagnostics for this format. A uniform variant also
scores free-form items as the mean of content and the harmonic mean of timing
and silence. It keeps the ranking of the 30 current configurations at a
Kendall rank correlation \citep{kendall1938} of 0.890. Chatty polled systems
lose the most, up to 4.5 points for polled Claude Sonnet 5
\citep{anthropic2026claude}.
Dataset aggregation averages each field over its defined item values.
A silent system earns Overall only on negative items, giving the observed
6.5-point dataset score for the never-fire policy.

\paragraph{Alternative aggregations.}
A softer negative-stream penalty changes little. Scoring a negative stream
with $M$ false alarms as $100/(1+M)$ instead of $100\max(0,1-M)$ keeps the
Overall ranking of the 30 current configurations at a rank correlation of
0.959. Charging each false alarm five seconds of a stream of length $T$, as
$100\max(0,1-5M/T)$, keeps it at 0.926. Overall moves by at most 2.7 and 5.9
points, in both cases for VideoLLM-online \citep{chen2024videollmonline}.
Weighting timing against silence also changes little. Replacing
$H(\TA,\SC)$ by $H_\beta=(1+\beta^2)\,\TA\,\SC/(\beta^2\TA+\SC)$ with
$\beta=2$, which favors silence, or $\beta=0.5$, which favors timing, keeps
the ranking at 0.972 or higher. Shifting weight between content and behavior
matters more. Weighting content by 0.25 or 0.75 instead of one half gives
rank correlations of 0.733 and 0.825, so the results tables report the three
components separately. Silence compliance also floors at zero once violations
reach the number of reference events. A smooth variant, $100/(1+|\viol|/N)$,
never floors and keeps the silence ranking at 0.954, and the violations per
event in Table~\ref{tab:violations} separate systems below the floor. Fully
silent positive items drop out of the silence aggregate. Most configurations
speak on most positive items, with a median of 92.8\%, but the blind
reference speaks on 34.6\% and Claude Sonnet 5~\citep{anthropic2026claude} in grounding on 54.1\%.
Counting their silent items as zero lowers their silence compliance from 44.2
to 17.2 and from 91.0 to 51.9. Qwen3-VL-8B~\citep{bai2025qwen3vl} still raises its silence
compliance more than fivefold in grounding under this count, from 9.7 to
53.9. Overall already scores fully silent items zero, so this selection
affects only the silence diagnostic.

\paragraph{Content-gated timing.}
Timing accuracy credits a matched first response whether or not its content
is correct. Content-gated timing keeps the same event windows and
first-response matching and counts the credit only when that response is
correct. With $c_n$ equal to 1 when the first matched response to event $n$
is correct and 0 otherwise, content-gated timing and the smooth silence
variant above are
\begin{equation}
\label{eq:variants}
\TA_{\mathrm{correct}} = \frac{100}{N}\sum_{n=1}^{N}
       c_n\left(1-\frac{\dly{n}}{\dbound}\right)_{+},
\qquad
\SC_{\mathrm{smooth}} = \frac{100}{1+|\viol|/N}.
\end{equation}
For counting instructions, a response is correct when it
states the running count after its event.
Table~\ref{tab:metric_variants} reports both for the 20 full-set
configurations, with content-gated timing averaged over all reference events
like the timing column beside it, and recomputes Overall per item
under each change. The Overall ranking agrees with the default at rank
correlations of 0.86 with content-gated timing, 0.83 with smooth silence,
and 0.81 with both, and the first two places never change. Content-gated
timing reorders the timing column itself, at 0.52 against timing accuracy,
because VideoLLM-online \citep{chen2024videollmonline} and polled Gemma-3n
\citep{google2025gemma3n} respond promptly with wrong content. Smooth
silence moves JoyAI-VL \citep{joyaivl2026} from eighth to third, since it
violates less per event than the polled backbones that share its floor.
Among the ten online configurations, timing and silence remain negatively
rank-correlated at $-0.29$ under both changes.
\input{fig_tex/tab_metric_variants}

\input{fig_tex/fig_aggregation}
\paragraph{Aggregations without the harmonic mean.}
Figure~\ref{fig:aggregation} rescores the 20 full-set configurations and two
scripted policies of Appendix~\ref{sec:gamed}, one that always responds and one
that never responds, under three alternatives to Overall: an arithmetic
mean of timing and silence in place of the harmonic mean, the plain mean of
the three scores per item, and the mean of the three table columns. Under
Overall the two policies score 0.0 and 6.5, below every real configuration.
Under the alternatives the always-responding policy scores 20.6, 27.5, and
32.9 and overtakes 3, 6, and 7 real configurations, and under the column
mean the never-responding policy scores 33.3. The harmonic mean is the term
that keeps both policies from scoring. The ranking of the real
configurations agrees with Overall at rank correlations of 0.78, 0.75, and
0.77.

\paragraph{Timing tolerance.}\label{app:sensitivity}
\input{fig_tex/tab_delta}
The default delay scale is five seconds. As shown in
Table~\ref{tab:delta}, rescoring the 30 configurations that use the current
export at delay scales of 2, 3, 8, and 10 seconds preserves their order.
Kendall rank correlations \citep{kendall1938} of the Overall ranking
against the default stay between 0.954 and 0.995, and those of the timing
ranking between 0.913 and 0.982. No configuration moves more than three
ranks, the top three never change, and mean Overall shifts from 34.4 to
35.5. Timing accuracy itself moves more, for example from 52.1 to 70.1 for
the human reference. Overall moves by at
most 3.6 points for any configuration. The scale therefore changes
magnitudes rather than the ordering.
A second check perturbs the reference timestamps. Moving every event timestamp after
the query by one or two seconds in either direction changes which replies
are matched. Independent Gaussian jitter with a standard deviation of one or
two seconds does the same. Keyed items need no judge and are rescored in
full. Free-text content accuracy keeps its original value. One-second
shifts and one-second jitter keep the Overall ranking of the 30
configurations at a rank correlation of at least 0.917. The timing ranking
stays at or above 0.945 and the silence ranking at or above 0.954. Mean absolute Overall changes
stay below 2.0 points. Two-second changes keep the Overall ranking at 0.876
or higher. Offline grounding by the two agents is the sensitive case. When
every reference moves one second later, Overall drops by 14.2 points for the
Claude Sonnet 5~\citep{anthropic2026claude} agent and 11.7 for the GPT-5.6 Sol agent \citep{openai2026gpt}. The premature replies
of the Claude Sonnet 5 agent rise from 0.30 to 0.56 per item. Earlier
references change their Overall by under 3 points, since each window opens
one second early. The early tolerance matters mainly for the same agents.
With free-text content again held fixed, setting $\epsilon$ to 0.5 or 2
seconds instead of 1 keeps the Overall ranking at a rank correlation of
0.963, and no configuration moves by more than 3.9 points. Without a
tolerance, at $\epsilon=0$, the rank correlation falls to 0.913 and the
Claude Sonnet 5 agent in grounding loses 14.9 points, because its premature
replies rise from 0.30 to 0.57 per item. Those of polled Qwen3-VL-8B
\citep{bai2025qwen3vl} rise only from 6.33 to 6.65.

\section{Judge and Segment Scoring}
\label{app:judge}
\input{fig_tex/tab14_judge}
\input{fig_tex/tab_judge_rescore}
Seven candidate judges are calibrated against 954 question--answer
triples. Gemini 3.5 Flash \citep{google2026gemini} first labeled each triple as correct or incorrect,
and annotators screened these labels and corrected wrong ones.
As shown in Table~\ref{tab:judge}, greedy Qwen3-14B \citep{yang2025qwen3}
agrees with 89.7\% of the final labels, with Cohen's kappa
\citep{cohen1960kappa} of 0.79, on par with Qwen3-32B and gpt-oss-20b
\citep{gptoss2025} at 90.0\% and 89.9\%. Its five sampled votes are unanimous
on 98.3\% of the triples, against 92.1\% and 92.3\% for those two, and a
five-vote majority raises its agreement only to 89.8\%. The final scorer
uses greedy Qwen3-14B to balance reliability and efficiency.
Rescoring all 30 current configurations with greedy Qwen3-32B keeps the
Overall ranking at a Kendall rank correlation \citep{kendall1938} of 0.972
against Qwen3-14B, and no Overall moves by more than 1.3 points.
Table~\ref{tab:judge_rescore} extends this rescoring to gpt-oss-20b,
Gemma-3-27B \citep{gemma3}, and Claude Opus 5.5
\citep{anthropic2026opus55}, each relabeling all
16{,}355 triples that the 30 configurations send to the judge, with speed
measured one triple at a time. Every judge keeps the
Overall ranking at a rank correlation of at least 0.959 against Qwen3-14B.
Among the judges whose calibration agreement is close to 90\%, Qwen3-14B is
the fastest and needs the least memory, and its weights are public and
fixed, so anyone can rescore offline and obtain the same verdicts.
The 108 free-form items comprise 107 Commentate items and one revision item.
The segment scorer follows \vdc{} \citep{chai2024auroracap}. Each
reference is decomposed into at most five question--answer pairs, averaging
2.0 across 303 cached reference texts. Answers are extracted from the
segment's emitted text and judged against those reference answers.
Caching fixes the decomposition across systems.

\section{Evaluation Prompts}
\label{app:prompts}
Two prompt shapes drive the adapted models. Single-turn modes, cumulative or
with a fresh prompt, restate the full task at every poll. The reply-history
mode runs one interleaved conversation, states the task once in the system
message, and sends a compact nudge per poll. The
system prompt reads verbatim as follows.
\begin{quote}\small\ttfamily\raggedright
You are a real-time visual assistant watching a LIVE video stream. You
receive frames as they arrive; the LAST image is the current moment. You
must decide, right now, whether to speak or to keep watching silently. Only
speak when this exact moment calls for it --- do not speak early, and do not
repeat what you already reported. If nothing needs saying, wait silently.
\end{quote}
Each single-turn poll then renders the stream time, the standing request, one
hint chosen by task type, and a fixed output format. The hint depends only on
the task type and never on whether an item contains an event, and positive
and negative items of the same task therefore receive identical prompts. The hints read
verbatim as follows.
\begin{center}\small
\begin{tabularx}{\linewidth}{@{}lX@{}}
\toprule
Task & Hint \\
\midrule
Look & Answer the question as soon as you are asked, from what you can see. \\
Recall & Answer from what you saw earlier in the stream. Answer immediately. \\
Recall, causal & Answer the question as soon as it becomes answerable from what you saw. \\
Time & Answer at the earliest moment the question becomes decidable from what you have seen; before that, wait. \\
Alert & Speak only at the exact moment the requested event happens. If it never happens, stay silent for the whole stream. \\
Track & Track the requested state across frames. Each time it changes (e.g. the count increases), speak the update (for counts: just the new number). Otherwise wait. \\
Track, revision & Track the requested state; speak whenever the correct answer changes, correcting your earlier answer if new evidence contradicts it. \\
Commentate & Follow the stream and speak one short line for each new step or scene as it happens, like a live commentator or guide. \\
\bottomrule
\end{tabularx}
\end{center}
The output format reads verbatim as follows.
\begin{quote}\small\ttfamily\raggedright
Respond in EXACTLY this format, nothing else:\\
DECISION: SPEAK or WAIT\\
RESPONSE: <if SPEAK, one short line --- the exact thing you would say out
loud now; if WAIT, leave blank>
\end{quote}
The reply-history mode adds to its system message ``You can see your own previous
turns. If you already reported this and nothing new happened, WAIT instead
of repeating.'', and each poll sends only
``\texttt{[stream time t\,s] New frames above. Decide now (DECISION /
RESPONSE).}''
{
For keyed items the rendered item replaces the standing request. It shows
the rewritten stem where the key provides one and the annotation question
otherwise, followed by one option per line and a fixed instruction. The
polling adapter and the agentic systems share the following verbatim
rendering.
\begin{quote}\small\ttfamily\raggedright
<stem>\\
A. <option 1>\\
B. <option 2>\\
\ldots\\
F. <option 6>\\
(When you speak, answer with just the option letter.)
\end{quote}
}

\section{System Configuration and Provenance}
\label{app:models}
The polling default is two frames per second, a 16-frame
sliding window, a 512-pixel long side, greedy decoding, and at most 96
generated tokens. The MiniCPM-o \citep{minicpmo2026} fresh-prompt run uses beam search instead. Polls occur once per video second. Query-answer items
permit a ten-second post-query response period. Reply-history variants
retain earlier frames and replies in one conversation. Fresh-prompt variants
start anew at each poll.

Native streaming systems keep their released interfaces and recommended
input rates. JoyAI-VL \citep{joyaivl2026} runs its own server-side memory
and silence decision at 4\,fps, and each tick pushes only new frames.
MMDuet2 \citep{wang2025mmduet2} replies or emits no reply after every
turn of two frames at 4\,fps. VideoLLM-online \citep{chen2024videollmonline}
decides after every frame at 8\,fps through its frame-interval gate with
the released threshold of 0.725. VideoChat3 \citep{videochat3} chooses
silence, standby, or a response in each one-second round. MOSS-Video
\citep{moss2026} consumes frames at real-time pace and emits silence tokens
while it observes. LiveCC \citep{chen2025livecc} comments on each
one-second clip and has no silence mechanism. Flash-VStream
\citep{zhang2024flashvstream} has no response decision of its own. Its
memory ingests 8\,fps, the polling adapter queries that memory once per
second, and the row keeps a lenient output parser. Dispider
\citep{qian2025dispider} exposes only whole-video question answering in
its released code, and it answers the grounding prompt with the times at
which it would respond.

Polled Claude calls the Anthropic API with the model name
\texttt{claude-sonnet-5} \citep{anthropic2026claude}. The Claude kernel
swaps call \texttt{claude-sonnet-5} or \texttt{claude-opus-5} as the
JoyAI-VL~\citep{joyaivl2026} model, and JoyAI-VL's memory summarizer runs on
\texttt{claude-haiku-4-5} in both. The Qwen3-VL-8B kernel swap serves the
model \citep{bai2025qwen3vl} on a B200 GPU inside the same controller. The Claude agents run
the Claude Code command-line agent \citep{anthropic2026claudecode} with the model alias \texttt{sonnet}.
The shell-tool run started on September 13, 2026, and the media-tool run on
September 18. Polled Gemini \citep{google2026gemini} calls \texttt{gemini-2.5-flash}
through the OpenAI-compatible endpoint with reasoning disabled, and 32
blocked or empty replies count as silence. The GPT agent calls the \texttt{gpt-5.6} API alias
\citep{openai2026gpt}, and the Gemini agent runs the Gemini command-line
agent with Gemini 3.5 Flash \citep{google2026gemini}. Providers can update
the model behind a name, and each result therefore keeps its run dates.

\paragraph{Human reference.} Volunteers answer in a browser that plays each
video at normal speed, and a volunteer can drag the progress bar and
re-watch the video within the part that has already played. The
question appears on screen at its query time. A volunteer presses a key as
soon as a response is possible, and that moment on the video clock becomes
the response time, as for a system emission. The video then pauses while
the volunteer types and resumes after submission. Alert items take a
response at every occurrence, and counting items take every change of the
count. A volunteer can update an answer by pressing the key again, and each
update counts as a new response. Volunteers answer independently. Some alerts are a bare ``now'', which
earns timing credit but no content credit, and 13.9\% of Alert replies take
this form. The reference shows live behavior rather than a ceiling.
Grounding sees the whole video, and logged agent polls take about half a
minute on a clock that waits for them. A live viewer has neither
advantage.
Figure~\ref{fig:human_interface} shows the interface in which volunteers answer keyed
questions by option letter.
\input{fig_tex/fig_human_interface}

AsynKV \citep{spotbench2026}
is the streaming controller of SPOT-Bench, built on Qwen2.5-VL-7B \citep{bai2025qwen25vl}. We run the released model on all
1{,}060 interactions. It scores 27.6 Overall, with content 25.6, timing 19.6, and
silence 78.0, the highest silence compliance of any online configuration
in Table~\ref{tab:system_scores}. It never responds on the 69 negative
streams and answers 66.9\% of Look questions correctly. Its gate also
suppresses the instructions that need continued responses: it stays silent
on 159 of the 160 counting instructions and responds on 10.3\% of Commentate
items.

\section{Measured Latency}
\label{app:latency}
\input{fig_tex/tab_latency}
Every prediction file stores the wall-clock duration of each poll, and
Table~\ref{tab:latency} summarizes these durations. A poll spans one
second of stream time, and a system keeps pace with a live stream only
when most polls finish within that second. Scores ignore these durations
because the runners advance the video clock only after a poll returns.
MOSS-Video \citep{moss2026} is the exception, since it consumes frames at
real-time pace. A live replay estimates what these durations cost. Eight polled runs are
replayed, the six open backbones and polled Claude Sonnet 5
\citep{anthropic2026claude} and Gemini 2.5 Flash \citep{google2026gemini}. Flash-VStream \citep{zhang2024flashvstream} logs irregular polls and is left out. In the replay, a
poll starts only after the previous one returns, uses the latest recorded
decision it could see, and replies after its measured latency. Keyed items
are rescored in full, and free-text content keeps its original value. Timing
accuracy then falls from 82.9 to 45.5 for polled Claude Sonnet 5 and from
93.1 to 51.6 for polled Gemini 2.5 Flash. Qwen3-VL-8B and Qwen3-VL-4B
\citep{bai2025qwen3vl} and LLaVA-OV2 \citep{llavaov2} keep pace and lose at most 7.6 points of timing accuracy. Slower
polls also repeat less, and low silence scores dominate the harmonic mean.
Overall therefore moves by at most 3.3 points across the eight replayed runs.
Only 0.8\% of the polls of the Claude Sonnet 5 agent with media tools and
1.4\% with shell tools return within the five-second delay scale. These two
agent runs logged every poll, and the same replay confirms the cost. Timing
accuracy falls from 66.5 and 70.7 to 0.3, since only 65 and 55 of their 711
and 695 recorded replies are ever sent, 22 and 23 of them after the video
ends. Silence compliance rises to 91.8 and 95.6, and Overall falls from 56.0
and 61.2 to 29.2 and 28.5, below the 41.3 of polled Claude Sonnet 5 under
the same replay. The GPT-5.6 Sol \citep{openai2026gpt} and Gemini agent runs
logged no poll times and are not replayed.
The measurements come from the hardware that produced each score, a B200
GPU, a 90\,GB MIG slice of a B200, an L40S GPU, or an API endpoint.
VideoLLM-online \citep{chen2024videollmonline} logs no per-frame time, and
MMDuet2 \citep{wang2025mmduet2} logs frame ingestion and generation as
separate steps, and both are omitted. The Claude Opus 5 kernel swap
\citep{anthropic2026opus} logged poll times on 64 of its items. Offline grounding issues one
whole-video call per item and has no per-poll latency.

\paragraph{Cost of one evaluation.}
Table~\ref{tab:cost} sums the logged poll time of one complete evaluation
on the hardware of Table~\ref{tab:latency}. Flash-VStream~\citep{zhang2024flashvstream} logs irregular
polls and has no per-minute cells.
The 812 videos hold 35.9 hours of stream, and a polled system is queried
about 119{,}000 times. With a fixed window, model time per minute of video
stays flat as videos grow. Reply history and larger frame windows cost more
on longer videos because their context grows. The evaluation metric itself
does not depend on duration. One judge pass over all configurations takes
2.2 B200 GPU-hours.
\input{fig_tex/tab_cost}

\section{Offline Accuracy of the Polled Backbones}
\label{app:backbone}
\input{fig_tex/tab19_backbone}
Table~\ref{tab:backbone} lists the numbers behind Figure~\ref{fig:backbone}.
Offline accuracy rescores the grounding run of each backbone on the 688
keyed items. Grounding lets a model decline to answer, and every declined item
counts as wrong. Gemma-3n \citep{google2025gemma3n} leaves 301 keyed items
unanswered and reaches 21.3 on the rest, close to the 19.0 expected from
random guessing on this option mix. Qwen2.5-Omni \citep{xu2025qwen25omni}
answers only 53.6\% of the items but is the most accurate when it answers.
\input{fig_tex/tab_keyed_agreement}
As shown in Table~\ref{tab:keyed_agreement}, the same backbone answers
different keyed items correctly offline and online. Agreement ranges from
44.2\% to 73.0\%, and Cohen's kappa \citep{cohen1960kappa} stays at or below
0.13 for five of the six backbones and reaches 0.40 for MiniCPM-o~\citep{minicpmo2026}.

\section{Protocol Audit}
\label{sec:gamed}
\input{fig_tex/fig11_audit}
Six scripted policies probe simple failure modes, and as shown in
Figure~\ref{fig:audit}, none competes with the main polling configurations
in the current score export. Always-fire scores 98.8 in timing but
approximately zero Overall. Never-fire scores 6.5 Overall, reflecting
credit on the negative items. Periodic emissions, bursts, background
chatter, and the parrot baseline also stay far below the polled systems.
These checks reject the tested shortcuts as competitive solutions.

\section{Detailed System Results}
\input{fig_tex/tab_system_scores}
\label{app:pertask}\label{app:analysis_tables}
\input{fig_tex/tab_task_acc}
\input{fig_tex/tab_task_ta}
\input{fig_tex/tab_task_sc}
Table~\ref{tab:system_scores} gives the aggregate content, timing, and
silence of every configuration, and Tables~\ref{tab:task_acc},
\ref{tab:task_ta}, and~\ref{tab:task_sc} separate them by task. Their
values come directly from the same export as Table~\ref{tab:fullset}. Overall averages item scores under the
format-specific scoring rule, and runtime is excluded.
Offline grounding receives the whole video and reports when it would have responded.
It serves only as a diagnostic reference.
\input{fig_tex/tab_violations}
Silence compliance reaches zero once an item's violations match its
reference events. As shown in Table~\ref{tab:violations}, this floor covers
72\% to 97\% of positive items for the polled models, whose repeated replies
per event still range from 2.8 to 19.2.
\input{fig_tex/fig15_harness}
\input{fig_tex/tab16_ablation}
\input{fig_tex/fig13_ablation}

As shown in Table~\ref{tab:ablation}, sampling rate and prompt hints move
Overall by at most 1.4 points across all polling-adapter ablations, whereas
a 256-frame window adds 4.2. Response history
lowers Overall by up to 9.6 points and raises it by up to 10.1. As shown in
Figure~\ref{fig:ablation}, reply history moves silence compliance more than
any frame or prompt setting.
As shown in Figure~\ref{fig:harness}, both
Claude Sonnet 5 \citep{anthropic2026claude} agent runs lower timing
accuracy and raise silence compliance relative to polling. Overall of the same
model moves from 44.6 under polling to 56.0 with media tools and to 61.2
with shell tools.

\section{Random Emission Thinning}
\label{app:sweep}
Each recorded emission is independently retained with a prescribed
probability, and the resulting stream is rescored. The four exported
curves use probabilities from 1.0 to 0.1. At low retention, silence
compliance generally improves while timing accuracy declines. These curves describe random post-processing
of fixed output traces and do not model a confidence-calibrated threshold.
The native threshold sweep in Appendix~\ref{app:planned} moves a system's
own trigger instead.

\section{Case Studies}
\label{app:cases}
\paragraph{Task overview.}
Figure~\ref{fig:teaser} takes twelve full frames
from three Wikimedia Commons videos. They are a basketball game in the public
domain \citep{navy2025basketball}, water heated to a boil under CC BY-SA 1.0
\citep{sertion2011boiling}, and a Spanish omelette under CC BY-SA 4.0
\citep{eleneb2021tortilla}. Each row keeps its source's clock. Four rows at
the top and bottom of the boiling-water source are trimmed to reach 16:9,
and no other image is cropped or retouched. The six instructions and their
replies are new illustrative annotations outside the scored benchmark.
Their times follow visible evidence and are not calibrated
earliest-answerable times. Source links, checksums, observations, and a
frame-extraction script accompany the assets.

\paragraph{Response errors on a railway crossing.}
\input{fig_tex/fig2_response_errors}
Figure~\ref{fig:response_errors} places constructed response errors on
a railway-crossing video \citep{weenybeany2024crossing},
licensed under CC BY 4.0. A reply before the train enters is premature, a second reply
to the same event is redundant, and a reply after the window opens loses
timing credit with its delay. Its false-alarm row uses a different
instruction, to report a vehicle going around a lowered barrier. No such
event occurs in the reviewed clip. The other rows ask for the train
entering the crossing. These are constructed examples, not measured
system traces.

\paragraph{Benchmark cases and system traces.}
\input{fig_tex/fig18_cases}
Figures~\ref{fig:case_IVQA}, \ref{fig:case_LVM}, \ref{fig:case_TOA},
\ref{fig:case_PTR}, \ref{fig:case_CST}, and~\ref{fig:case_LCG} show the
instruction, reference times, sampled frames, and system emissions on one
item per task. In each trace, ticks mark utterances, gray dashed lines mark
reference events, the black dashed line marks the query, n counts replies, and the right column
quotes each first reply. Grounding sees the whole video. They illustrate how an instruction
can remain active after the first answer, how repeated speech differs
from repeated valid events, and how grounding benefits from future access.
Individual cases are illustrative rather than estimates of aggregate error
rates. The plotted traces and answer text come from the saved case records.
\input{fig_tex/fig19_main_case}
The Alert item in Figures~\ref{fig:case_PTR} and~\ref{fig:measured_case} also illustrates the difference
between repeated speech and missed events in the saved traces.
Under chronological first-response matching with the one-second early
tolerance, the polled Qwen3-VL-8B \citep{bai2025qwen3vl} emits 104 times, five before the first
window, four first matches, and 95 repetitions within answered windows.
JoyAI-VL \citep{joyaivl2026} emits at 1, 58, and 102 seconds, one premature reply, one match,
and one repetition, leaving three events unmatched.

\section{Agentic System Details}
\label{app:agents}
\input{fig_tex/tab_agents_summary}
Agents use the Qwen-MM-Plugins media tools \citep{qwenmmplugins2026} to
extract and inspect frames, crop images, and read timestamps. The Claude
\citep{anthropic2026claude} shell-tool run instead allows the built-in shell and file
tools of Claude Code \citep{anthropic2026claudecode} in a per-item working directory that holds the trimmed prefix and no
answer keys. Before each
poll, ffmpeg \citep{ffmpeg} trims the source video to the current stream time. The agent
receives only this prefix, its instruction, and earlier utterances.
The GPT-5.6 Sol loop \citep{openai2026gpt} allows at most twelve tool turns and
1{,}500 completion tokens per session. Claude and Gemini
\citep{google2026gemini} use their command-line agent loops.
Each polling session has a 240-second timeout. Timeout is not a latency measurement.
Every reported polling run works in an isolated directory that holds prefix videos
and tool configuration but no answer keys. Apart from the shell-tool run, the
tools are restricted to visual inspection.
In grounding, the Claude and GPT-5.6 Sol agents receive the whole video with the
same plugins. They report the video second of the key evidence for each
response, no later than the question time for query items. Their sessions time out after 1{,}800 and 900 seconds. Claude Sonnet 5 \citep{anthropic2026claude} without tools receives the grounding prompt of the open backbones.

As shown in Table~\ref{tab:agents_summary}, both Claude Sonnet 5~\citep{anthropic2026claude} agent runs
raise silence compliance above polling. The media-tool run replies 706
times and leaves 11 items silent. The media-tool and shell-tool runs each
retain 11{,}696 poll latencies, with medians of 28.72 and 37.17 seconds
and 90th percentiles of 87.4 and 91.6 seconds.

\section{Additional Diagnostics}
\label{app:planned}
\input{fig_tex/tab_oracle_restraint}
\paragraph{Oracle restraint.}
Table~\ref{tab:oracle_restraint} applies a reference-based filter to existing
Qwen3-VL-8B \citep{bai2025qwen3vl} and JoyAI-VL \citep{joyaivl2026} traces on the items shared with the agent runs. It removes all
negative-stream replies and retains only the first response attributed to
each reference window. The filter deletes 3,418 Qwen replies and 1,977
JoyAI replies without rewriting retained content or timestamps. Timing
is unchanged, while silence compliance reaches 100.0 by construction.
Overall rises from 33.9 to 66.2 for Qwen and from 38.4 to 63.9 for JoyAI.
Applied to all 1{,}060 items of the four randomly thinned traces, the same
filter leaves timing unchanged and lifts silence compliance to 100.0 for each.
Content falls slightly because deleting replies changes segment-level text
and counting answers. This privileged post-processing diagnostic measures
the effect of suppressing violations with privileged reference annotations.

\input{fig_tex/tab_threshold}
\paragraph{Native response threshold.}
VideoLLM-online \citep{chen2024videollmonline} speaks when the probability
of its frame-interval token falls below a threshold, released at 0.725.
Table~\ref{tab:threshold} reruns it at 0.5 and 0.9 on the items shared with the agent runs. Raising the threshold from 0.5 to 0.9 raises timing
accuracy from 30.2 to 51.7 and lowers silence compliance from 56.4 to
18.4, while Overall falls from 17.2 to 7.4. The native trigger therefore
moves along the same timing--silence trade-off as random thinning and
does not escape it.

\paragraph{Decomposition of the blind reference.}
Figure~\ref{fig:blind} splits Overall into the points contributed by each
group of items. The blind reference earns 6.4 of its 15.7 points on the 69
negative streams, where it speaks on 1.4\%, and 0.6 points on the 303 items
without options. With frames, the same backbone gains 19.1 points, mostly
from content, and loses 2.5 points on negative streams because it now
triggers on 40.6\% of them. VideoLLM-online
\citep{chen2024videollmonline}, LiveCC \citep{chen2025livecc}, and
Flash-VStream \citep{zhang2024flashvstream} score below the blind reference
through silence and output format. The first speaks on every negative
stream and names no option in 75.4\% of its multiple-choice replies, the
second comments on each one-second clip with no silence mechanism, and the
third repeats its answer on 98\% of items. For LiveCC, panel \textbf{a} follows Table~\ref{tab:fullset}, where the judge
grades its free-text answers, and panel \textbf{b} counts an answer only
when it names the right option, which 56.1\% of its multiple-choice replies
do not.
\input{fig_tex/fig_blind}

\paragraph{Reply history and task state.}
Figure~\ref{fig:reply_history} compares polling with a fresh prompt and
with reply history for six backbones. History raises silence compliance for
all six. On counting instructions, which require a report for every event,
it lowers the share of missed later events from 20.9\% to 9.4\% for
Qwen3-VL-8B~\citep{bai2025qwen3vl} and from 23.9\% to 7.5\% for LLaVA-OV2 \citep{llavaov2}.
Gemma-3n \citep{google2025gemma3n} and Qwen2.5-Omni
\citep{xu2025qwen25omni} instead stop after their first answer on 44.1\% and 24.7\% of
items, against 0.8\% and 6.8\% with a fresh prompt.
\input{fig_tex/fig_reply_history}

\paragraph{What helps at live pace.}
Among configurations that keep pace with the stream, a gain in timing
accuracy or in silence compliance costs the other, and none exceeds 40 on
both. Deleting only premature, repeated, and negative-stream responses lifts
silence compliance to 100 at unchanged timing, whereas fewer responses
overall only trade timing for silence. Reply history must also hold task
state: it reduces missed later events for two backbones, while two others
stop after their first answer (Figure~\ref{fig:reply_history}). A 256-frame
window only reaches the live frontier at 4.1\,s per poll, and agent loops
lie beyond it, but replaying the logged agent latency leaves them almost no
timing credit (Appendix~\ref{app:latency}).

\section{Errors by Domain and Duration}
\label{app:errors}
For each of the 19 systems run on all 1{,}060 interactions, we rerun the
scorer's event matching item by item and stratify the errors. The monitored
window is where the instruction is active and responses are scored: the
whole stream for standing instructions, and the 10\,s after the question for
answer-when-asked items, whose earlier stream is history.
Table~\ref{tab:errors_duration} controls for event density and clusters
standard errors by video. Longer monitoring raises missed events, while wrong content does not change. Negative streams draw a response more often as they grow longer,
but false alarms per no-event minute fall from 3.7 under one minute to 0.7
beyond five minutes, so the rise reflects longer exposure. Accuracy on
answer-when-asked items falls from 66.2 with under 30\,s of history to 34.5
with 180\,s or more, where most items are Recall. With task, duration, and
event density held fixed, domains differ mainly in content
(Table~\ref{tab:errors_domain}). Figure~\ref{fig:failure_cases} shows five
verified failures of polled Qwen3-VL-8B \citep{bai2025qwen3vl}, one for each error type, with the
instruction, the reference, the output, frames around the reference event,
and every reply on the video timeline.
\input{fig_tex/tab_errors_duration}
\input{fig_tex/tab_errors_domain}
\input{fig_tex/fig_failure_cases}

\section{Scene and Target Difficulty}
\label{app:scene}
Claude Opus 5.5 \citep{anthropic2026opus55} rated all 1{,}060 items on a fixed three-level rubric with
four scene dimensions, clutter, camera motion, shot changes, and concurrent
activity, and three target dimensions, occlusion of the target, look-alike
objects, and near-miss moments. The ratings use the instruction, the
reference events, and sampled frames, without any system output. We regress
item-level errors on these ratings with system, task, and domain fixed
effects, duration, and event density, with standard errors clustered by
video, separately for the 10 online full-set configurations, streaming
systems and per-second polling, and the 9 offline ones, which see the whole
video before answering. Figure~\ref{fig:scene_target} shows that scene
clutter raises neither premature responses nor triggered negative streams
in either group, and the near-miss rating itself does not vary with clutter,
with a Spearman correlation \citep{spearman1904} of $-0.01$. Each level of near-miss content raises the share of items
with a premature response by 8.4 points online and 7.3 offline, and the
share of triggered negative streams by 11.7 and 11.9. Each level of
occlusion raises missed events by 5.2 and 6.1 points. The groups differ in
how often an error repeats: per level of near-miss content, false alarms
per minute on negative streams rise by 2.35 online and 0.10 offline, and
shot changes add repeated responses only online.
\input{fig_tex/fig_scene_target}

\section{Limitations and Future Directions}
\label{app:limitations}
\paragraph{Coverage.}
Videos mainly last a few minutes and do not test hour-long monitoring.
All items are visual, and audio channels such as speech and sound events
remain outside the benchmark. Negative streams and counting-derived near
misses cover only part of the space of unnecessary responses.
\paragraph{Scope of the scores.}
Stream-time evaluation isolates response behavior from hardware speed.
It does not certify live deployment, and the latency measurements come
from heterogeneous hardware.
\paragraph{Scope of the comparisons.}
Aggregate correlations across a few systems cannot identify causal
bottlenecks. Existing comparisons of agentic systems change multiple components and
lack a complete factorial design. Future work should pair identical items and model
versions, test long negative intervals, and time every
system on shared hardware.

%% file: fig_tex/tab2_taxonomy.tex
\begin{table}[tp]
\centering\small
\caption{\textbf{Six tasks, three response obligations.} Counts follow annotation labels. A system must stay silent whenever a new response is unwarranted.}
\label{tab:taxonomy}
\begin{tabularx}{\textwidth}{@{}lrXX@{}}
\toprule
\textbf{Task} & \textbf{Items} & \textbf{Speak when\ldots} & \textbf{Stay silent when\ldots}\\
\midrule
Look & 159 & The user asks about the current view. & The query has already been answered.\\
Recall & 163 & The user asks about earlier evidence. & No new query or update calls for speech.\\
\addlinespace[3pt]
Time & 126 & An order or temporal condition resolves. & The required temporal evidence is incomplete.\\
Alert & 313 & The requested event occurs. & The scene is only a near miss.\\
\addlinespace[3pt]
Track & 192 & A count or state needs updating. & The tracked state has not changed.\\
Commentate & 107 & The unfolding scene calls for live commentary. & There is no relevant update to report.\\
\bottomrule
\end{tabularx}
\end{table}

%% file: fig_tex/fig9_extstats.tex
\begin{figure}[tp]
\centering
\includegraphics[width=\linewidth]{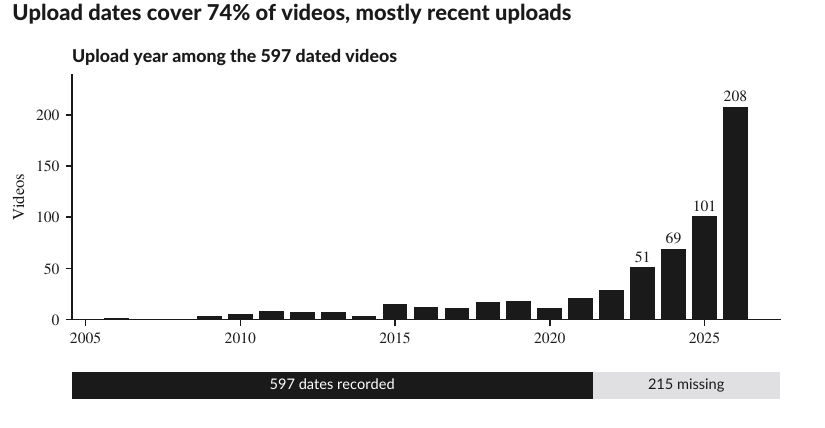}
\caption{\textbf{Half of the dated videos appeared in 2025 or later.} Upload years and date availability for the 812 videos.}
\label{fig:extstats}
\end{figure}

%% file: fig_tex/tab6_domains.tex
\begin{table}[tp]
\centering
\caption{\textbf{The ten content domains of \bench.} Videos and interactions per
domain, with the share of videos.}
\label{tab:domains}
\vspace{0pt}
\small
\setlength{\tabcolsep}{6pt}
\begin{tabular*}{\textwidth}{@{\extracolsep{\fill}}l rrr@{}}
\toprule
\textbf{Domain} & \textbf{Videos} & \textbf{Interactions} & \textbf{Share \%} \\
\midrule
Sports and exercise coaching & 181 & 254 & 22.3 \\
Cooking and kitchen & 103 & 127 & 12.7 \\
Home and life skills & 103 & 115 & 12.7 \\
Education and children & 102 & 185 & 12.6 \\
Professional, office and accessibility & 97 & 104 & 11.9 \\
Games, AR and entertainment & 74 & 85 & 9.1 \\
Driving, traffic and mobility & 65 & 76 & 8.0 \\
Crafts, creation and skills & 39 & 46 & 4.8 \\
Medical, first-aid and care & 28 & 44 & 3.4 \\
Industrial, repair and operations & 20 & 24 & 2.5 \\
\midrule
Total & 812 & 1{,}060 & 100.0 \\
\bottomrule
\end{tabular*}
\end{table}

%% file: fig_tex/fig_annotation_pipeline.tex
\begin{figure}[!htb]
\centering
\includegraphics[width=\linewidth]{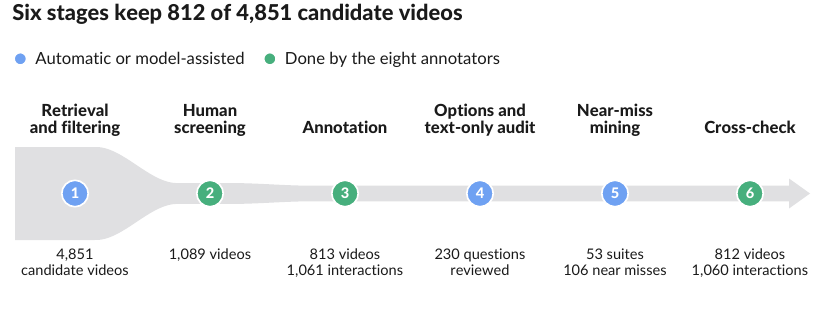}
\caption{\textbf{Six stages keep 812 of 4,851 candidate videos.} The band height is proportional to the number of videos kept after each stage, and the counts kept are listed below the band.}
\label{fig:annotation_pipeline}
\end{figure}

%% file: fig_tex/fig_annotation_interface.tex
\begin{figure}[!htb]
\centering
\includegraphics[width=\linewidth]{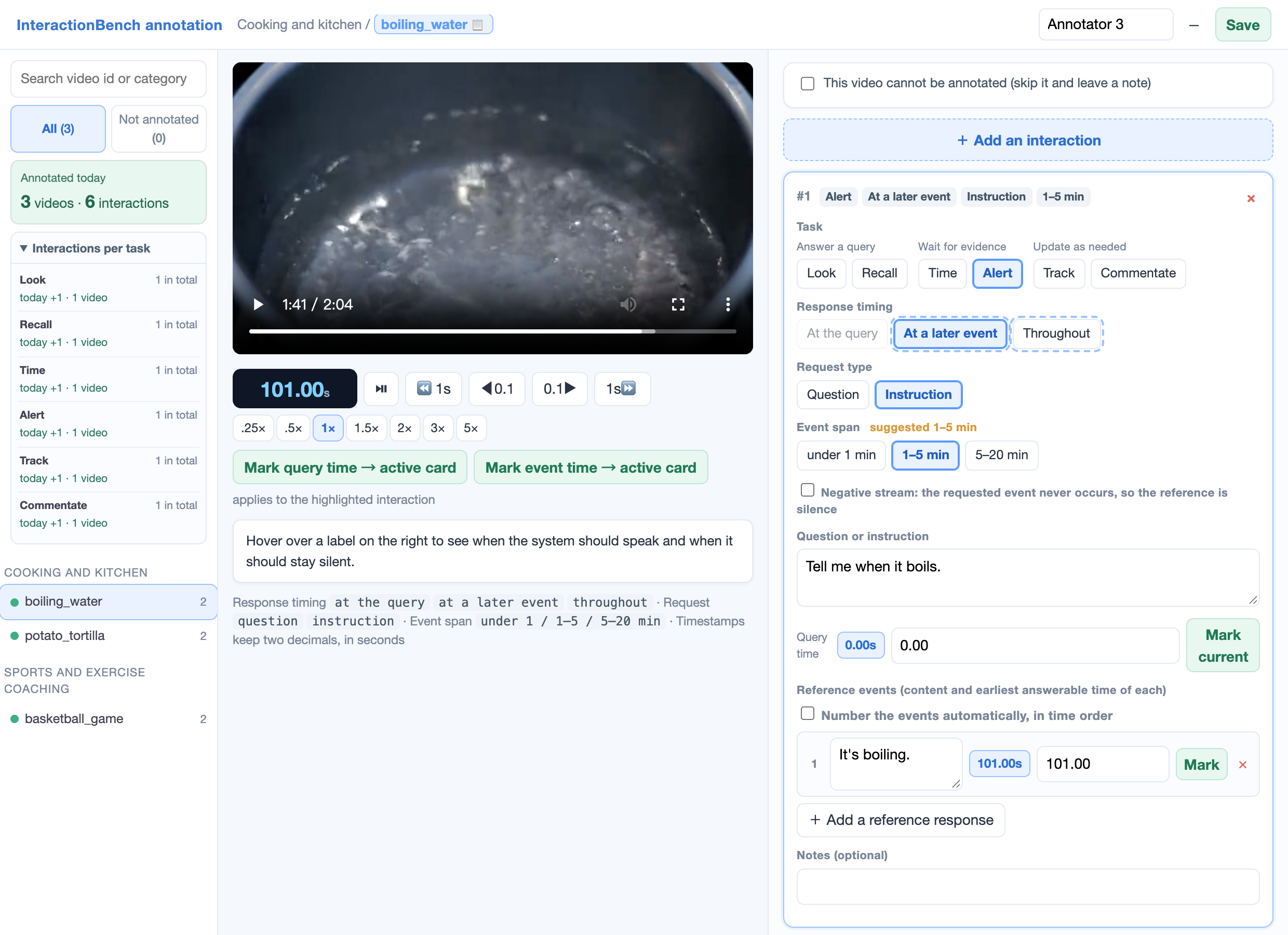}
\caption{\textbf{Annotation interface.} The annotator assigns the task and marks the query time and each reference event on the video clock, shown with English labels on a public video \citep{sertion2011boiling}.}
\label{fig:annotation_interface}
\end{figure}

%% file: fig_tex/tab8_families.tex
\begin{table}[tp]
\centering\small
\caption{\textbf{What contributes to the overall score.} Scores average
within each item before dataset aggregation. Higher is better.}
\label{tab:families}
\begin{tabularx}{\textwidth}{@{}p{.22\textwidth}XX@{}}
\toprule
\textbf{Response format} & \textbf{Content requirement} & \textbf{Overall score}\\
\midrule
Query answer & Correct first post-query answer & Mean of content and harmonic mean of timing and silence.\\
Event trigger & Correct matched event responses & Same combination as query answers.\\
Count updates & Exact final count & Same combination; content is final-count accuracy.\\
Free-form updates & Correct content in each segment & Segment content, with empty segments scoring zero.\\
No-event stream & No utterance & 100 if silent throughout; 0 otherwise.\\
\bottomrule
\end{tabularx}
\end{table}

%% file: fig_tex/tab_metric_variants.tex
\begin{table}[tp]
\centering\small
\caption{\textbf{The ranking holds under content-gated timing and smooth silence.} Rewritten is the share of correct first replies later overwritten, and ranks are among the 20 full-set configurations.}
\label{tab:metric_variants}
\setlength{\tabcolsep}{5pt}
\begin{tabular*}{\textwidth}{@{\extracolsep{\fill}}lrrrrrrr@{}}
\toprule
 & & & & \multicolumn{4}{c}{\textbf{Overall and rank}} \\
\cmidrule(lr){5-8}
\textbf{System} & $\TA$ & $\TA_{\mathrm{correct}}$ & \makecell{Rewritten\\\%} & Default & \makecell{with\\$\TA_{\mathrm{correct}}$} & \makecell{with\\$\SC_{\mathrm{smooth}}$} & \makecell{with\\both} \\
\midrule
\multicolumn{8}{l}{\textit{References}}\\
Qwen3-VL-8B~\citep{bai2025qwen3vl}, blind & 33.5 & 9.7 & 3.1 & 15.7 (18) & 11.7 (19) & 17.4 (18) & 12.3 (19) \\
\midrule
\multicolumn{8}{l}{\textit{Native streaming}}\\
VideoLLM-online~\citep{chen2024videollmonline} & 36.8 & 1.7 & 3.0 & 14.5 (19) & 2.3 (20) & 18.1 (17) & 4.9 (20) \\
MMDuet2~\citep{wang2025mmduet2} & 39.6 & 12.7 & 9.5 & 17.0 (16) & 15.8 (16) & 20.4 (16) & 17.2 (16) \\
\midrule
\multicolumn{8}{l}{\textit{Turn-based polling}}\\
Flash-VStream~\citep{zhang2024flashvstream} & 51.2 & 21.8 & 20.8 & 14.3 (20) & 14.1 (17) & 17.2 (19) & 15.1 (17) \\
Qwen2.5-Omni~\citep{xu2025qwen25omni} & 31.9 & 18.3 & 23.0 & 18.7 (15) & 17.3 (15) & 22.4 (15) & 19.8 (15) \\
Qwen3-VL-4B~\citep{bai2025qwen3vl} & 32.6 & 27.5 & 5.6 & 26.2 (14) & 25.8 (11) & 29.5 (13) & 28.6 (10) \\
Gemma-3n~\citep{google2025gemma3n} & 80.0 & 41.2 & 24.6 & 27.8 (12) & 25.0 (12) & 32.8 (10) & 28.0 (12) \\
MiniCPM-o~\citep{minicpmo2026} & 40.3 & 26.3 & 11.1 & 29.0 (11) & 26.6 (10) & 32.3 (12) & 28.8 (9) \\
LLaVA-OV2~\citep{llavaov2} & 78.9 & 45.9 & 22.2 & 34.5 (6) & 31.3 (5) & 40.2 (5) & 35.0 (6) \\
Qwen3-VL-8B & 75.6 & 50.6 & 20.6 & 34.8 (5) & 33.7 (4) & 40.6 (4) & 37.8 (3) \\
\midrule
\multicolumn{8}{l}{\textit{Offline temporal grounding}}\\
Gemma-3n & 13.9 & 6.8 & 10.6 & 16.1 (17) & 13.5 (18) & 17.0 (20) & 14.0 (18) \\
Qwen2.5-Omni & 28.3 & 17.2 & 1.0 & 27.2 (13) & 22.5 (14) & 27.2 (14) & 22.6 (14) \\
MiniCPM-o & 40.1 & 24.8 & 4.7 & 30.6 (10) & 24.9 (13) & 32.6 (11) & 26.4 (13) \\
Qwen3-VL-4B & 39.1 & 25.3 & 3.6 & 33.8 (9) & 28.4 (8) & 34.8 (9) & 29.0 (8) \\
LLaVA-OV2 & 39.3 & 24.1 & 4.5 & 33.9 (7) & 27.6 (9) & 34.8 (8) & 28.3 (11) \\
Qwen3-VL-8B & 39.9 & 25.7 & 8.4 & 35.7 (4) & 30.0 (7) & 36.5 (7) & 30.5 (7) \\
Claude Sonnet 5~\citep{anthropic2026claude} & 35.8 & 27.4 & 0.3 & 38.6 (3) & 35.1 (3) & 38.7 (6) & 35.2 (5) \\
GPT-5.6 Sol~\citep{openai2026gpt} agent & 60.8 & 53.7 & 4.2 & 58.8 (2) & 55.9 (2) & 59.6 (2) & 56.4 (2) \\
Claude Sonnet 5 agent & 69.2 & 60.5 & 4.2 & 66.1 (1) & 62.4 (1) & 66.6 (1) & 62.9 (1) \\
\midrule
\multicolumn{8}{l}{\textit{Native Real-time Interaction System (JoyAI-VL)}}\\
JoyAI-VL~\citep{joyaivl2026} & 66.0 & 39.1 & 8.0 & 33.8 (8) & 31.1 (6) & 41.1 (3) & 36.5 (4) \\
\bottomrule
\end{tabular*}
\end{table}

%% file: fig_tex/fig_aggregation.tex
\begin{figure}[tp]
\centering
\includegraphics[width=\linewidth]{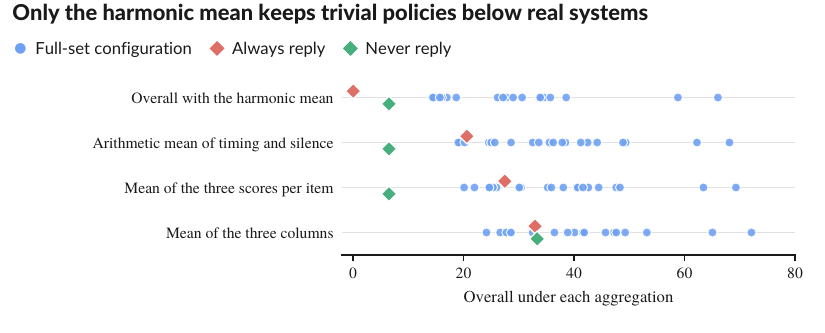}
\caption{\textbf{Only the harmonic mean keeps trivial policies below real systems.} Each row rescores the 20 full-set configurations and two scripted policies under one aggregation.}
\label{fig:aggregation}
\end{figure}

%% file: fig_tex/tab_delta.tex
\begin{table}[tp]
\centering\small
\caption{\textbf{Rescoring 30 configurations with other delay scales.} Kendall $\tau$ \citep{kendall1938} compares each ranking with the default five-second scale, and the top three configurations never change.}
\label{tab:delta}
\setlength{\tabcolsep}{5pt}
\begin{tabular*}{\textwidth}{@{\extracolsep{\fill}}lrrrr@{}}
\toprule
\textbf{Delay scale $\dbound$} & \textbf{$\tau$, Overall}$\uparrow$ & \textbf{$\tau$, timing}$\uparrow$ & \textbf{Largest rank move}$\downarrow$ & \textbf{Mean Overall} \\
\midrule
2\,s & 0.954 & 0.913 & 3 & 34.4 \\
3\,s & 0.968 & 0.949 & 2 & 34.7 \\
5\,s, default & 1.000 & 1.000 & 0 & 35.1 \\
8\,s & 0.995 & 0.982 & 1 & 35.4 \\
10\,s & 0.991 & 0.963 & 1 & 35.5 \\
\bottomrule
\end{tabular*}
\end{table}

%% file: fig_tex/tab14_judge.tex
\begin{table}[tp]
\centering\small
\caption{\textbf{Judge calibration on 954 reference-labeled answers.} Higher agreement in percent and higher kappa are better. Bold marks the selected judge.}
\label{tab:judge}
\setlength{\tabcolsep}{4pt}
\begin{tabular*}{\linewidth}{@{\extracolsep{\fill}}lrrrrr@{}}
\toprule
& \multicolumn{2}{c}{\textbf{One greedy verdict}} & \multicolumn{2}{c}{\textbf{Five-vote majority}} & \\
\cmidrule(lr){2-3}\cmidrule(lr){4-5}
\textbf{Judge} & \makecell{Agree\,\%$\uparrow$} & $\kappa\uparrow$ & \makecell{Agree\,\%$\uparrow$} & $\kappa\uparrow$ & \makecell{Unanimous\\votes\,\%}\\
\midrule
Llama-3-8B \citep{llama3} & 71.3 & .458 & 71.0 & .453 & 93.6\\
Qwen3-VL-8B \citep{bai2025qwen3vl} & 84.9 & .699 & 85.3 & .707 & 96.5\\
Gemma-3-27B \citep{gemma3} & 83.8 & .682 & 84.0 & .686 & 99.0\\
Qwen3-30B-A3B \citep{yang2025qwen3} & 87.0 & .739 & 86.9 & .737 & 97.0\\
\textbf{Qwen3-14B~\citep{yang2025qwen3}, selected} & 89.7 & .790 & 89.8 & .792 & 98.3\\
Qwen3-32B~\citep{yang2025qwen3} & 90.0 & .799 & 89.5 & .789 & 92.1\\
gpt-oss-20b \citep{gptoss2025} & 89.9 & .798 & 90.6 & .810 & 92.3\\
\midrule
\multicolumn{6}{l}{\textit{Ensembles, where dashes mark inapplicable fields}}\\
Seven-judge majority & -- & -- & 88.9 & .778 & --\\
All 35 sampled votes & -- & -- & 89.1 & .783 & --\\
Top-three judge majority & -- & -- & 91.7 & .832 & --\\
Top-three, 15 sampled votes & -- & -- & 91.7 & .833 & --\\
Top-three greedy verdicts & -- & -- & 91.6 & .831 & --\\
\bottomrule
\end{tabular*}
\end{table}

%% file: fig_tex/tab_judge_rescore.tex
\begin{table}[tp]
\centering\small
\caption{\textbf{Other judges keep the Overall ranking.} $\tau$ is the Kendall rank correlation of Overall, the largest change is over system configurations, and speed is measured on one B200 GPU.}
\label{tab:judge_rescore}
\setlength{\tabcolsep}{5pt}
\begin{tabular*}{\textwidth}{@{\extracolsep{\fill}}lrrrrrrrr@{}}
\toprule
 & \multicolumn{2}{c}{\textbf{Calibration}} & \multicolumn{4}{c}{\textbf{Against Qwen3-14B}} & & \\
\cmidrule(lr){2-3}\cmidrule(lr){4-7}
\textbf{Judge} & Agree\,\%$\uparrow$ & $\kappa\uparrow$ & Agree\,\% & $\kappa$ & $\tau$ & \makecell{Largest\\change} & \makecell{Triples\\per s$\uparrow$} & \makecell{GPU\\GB$\downarrow$} \\
\midrule
\textbf{Qwen3-14B~\citep{yang2025qwen3}, selected} & 89.7 & .790 & -- & -- & -- & -- & 3.6 & 28 \\
Qwen3-32B~\citep{yang2025qwen3} & 90.0 & .799 & 94.2 & .865 & .972 & 1.3 & 3.0 & 62 \\
gpt-oss-20b~\citep{gptoss2025} & 89.9 & .798 & 92.7 & .830 & .977 & 1.5 & 1.1 & 42 \\
Gemma-3-27B~\citep{gemma3} & 83.8 & .682 & 89.8 & .774 & .959 & 2.0 & 2.2 & 53 \\
Claude Opus 5.5~\citep{anthropic2026opus55} & 91.1 & .822 & 92.3 & .822 & .968 & 1.7 & -- & -- \\
\bottomrule
\end{tabular*}
\end{table}

%% file: fig_tex/fig_human_interface.tex
\begin{figure}[!htb]
\centering
\includegraphics[width=\linewidth]{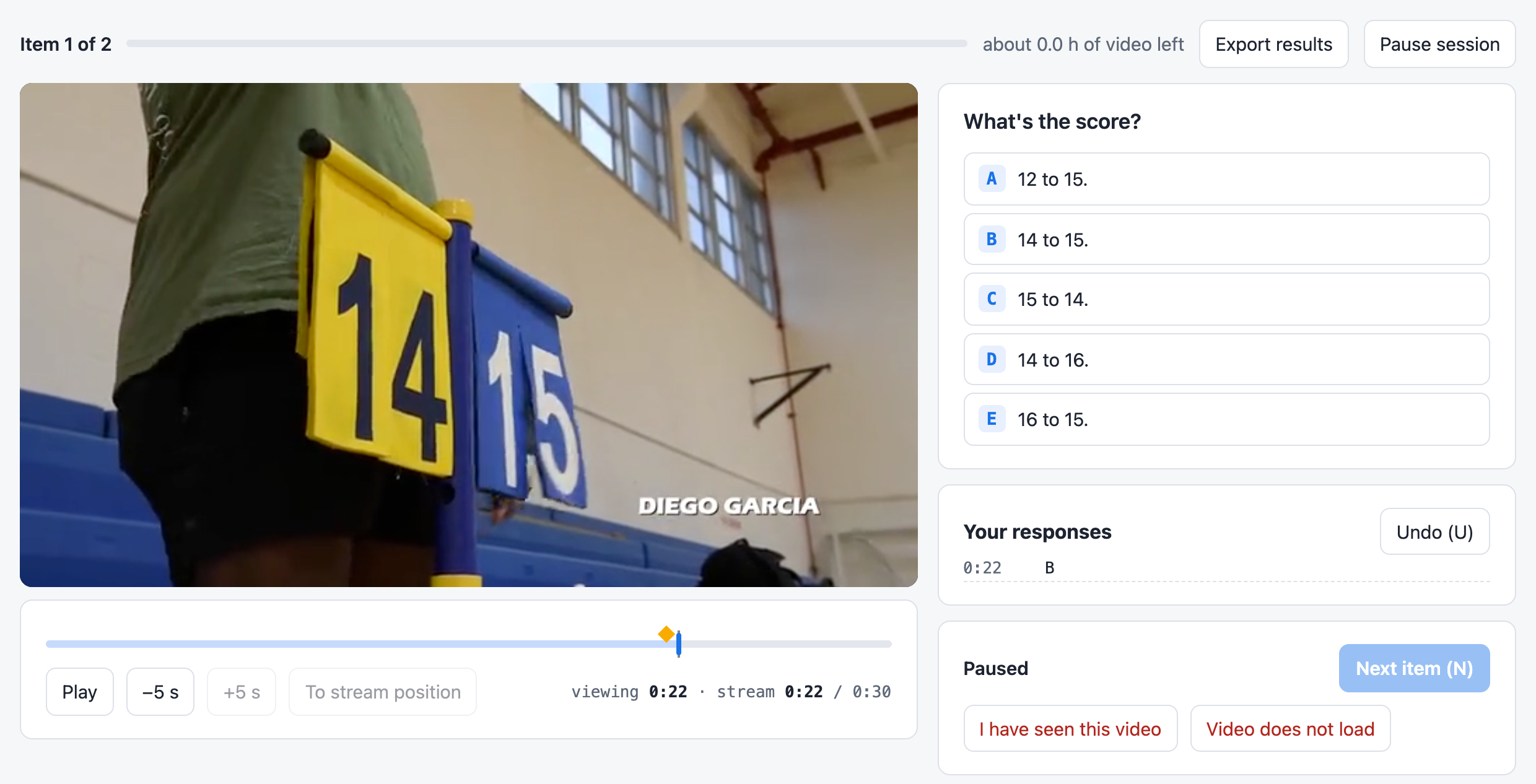}
\caption{\textbf{Human test interface.} A keyed request appears at its query time and is answered by option letter at the stream position, shown on a public video \citep{navy2025basketball} with example options.}
\label{fig:human_interface}
\end{figure}

%% file: fig_tex/tab_latency.tex
\begin{table}[tp]
\centering\small
\caption{\textbf{Measured wall-clock latency per poll.} Each system is timed on the hardware that produced its scores, and polls arrive once per stream second.}
\label{tab:latency}
\setlength{\tabcolsep}{5pt}
\begin{tabular*}{\textwidth}{@{\extracolsep{\fill}}llrrr@{}}
\toprule
\textbf{System} & \textbf{Hardware} & \textbf{Median, s}$\downarrow$ & \textbf{90th pct., s}$\downarrow$ & \textbf{Within 1\,s, \%}$\uparrow$ \\
\midrule
JoyAI-VL~\citep{joyaivl2026}, native & B200 & 0.38 & 0.78 & 98.9 \\
JoyAI / Qwen3-VL-8B~\citep{bai2025qwen3vl} & B200 & 0.44 & 0.91 & 93.9 \\
Qwen3-VL-4B~\citep{bai2025qwen3vl}, polled & B200, 90\,GB MIG slice & 0.39 & 0.43 & 99.8 \\
Qwen3-VL-8B, polled & B200, 90\,GB MIG slice & 0.48 & 0.63 & 98.9 \\
LLaVA-OV2~\citep{llavaov2}, polled & B200, 90\,GB MIG slice & 0.53 & 0.80 & 94.8 \\
Qwen3-VL-8B, reply history & B200, 90\,GB MIG slice & 0.83 & 1.93 & 64.7 \\
MiniCPM-o~\citep{minicpmo2026}, polled & B200, 90\,GB MIG slice & 1.26 & 1.70 & 3.0 \\
Qwen3-VL-8B, 64 frames & B200, 90\,GB MIG slice & 1.35 & 1.59 & 9.8 \\
Gemma-3n~\citep{google2025gemma3n}, polled & B200, 90\,GB MIG slice & 1.67 & 2.95 & 2.2 \\
Qwen2.5-Omni~\citep{xu2025qwen25omni}, polled & B200, 90\,GB MIG slice & 1.93 & 2.00 & 0.0 \\
Gemini 2.5 Flash~\citep{google2026gemini}, polled & Google API & 1.91 & 5.29 & 1.5 \\
Claude Sonnet 5~\citep{anthropic2026claude}, polled & Anthropic API & 2.42 & 3.59 & 0.0 \\
JoyAI / Claude Sonnet 5 & Anthropic API & 3.33 & 5.19 & 0.0 \\
JoyAI / Claude Opus 5~\citep{anthropic2026opus} & Anthropic API & 4.18 & 6.54 & 0.0 \\
Qwen3-VL-8B, 256 frames & B200, 90\,GB MIG slice & 4.12 & 6.08 & 9.0 \\
Flash-VStream~\citep{zhang2024flashvstream}, polled & L40S & 6.95 & 10.96 & 6.0 \\
Claude Sonnet 5 agent, media tools & Anthropic API & 28.72 & 87.42 & 0.0 \\
Claude Sonnet 5 agent, shell tools & Anthropic API & 37.17 & 91.62 & 0.0 \\
\bottomrule
\end{tabular*}
\end{table}

%% file: fig_tex/tab_cost.tex
\begin{table}[tp]
\centering\small
\caption{\textbf{A fixed window keeps the cost per minute flat.} Model hours sum every logged poll of one evaluation, and the last four columns split videos by duration in minutes.}
\label{tab:cost}
\setlength{\tabcolsep}{5pt}
\begin{tabular*}{\textwidth}{@{\extracolsep{\fill}}lrrrrr@{}}
\toprule
 & & \multicolumn{4}{c}{\textbf{Model seconds per video minute}} \\
\cmidrule(lr){3-6}
\textbf{Configuration} & \textbf{Model hours}$\downarrow$ & under 1 & 1--3 & 3--5 & over 5 \\
\midrule
Qwen3-VL-4B~\citep{bai2025qwen3vl}, polled & 12.9 & 18 & 18 & 19 & 19 \\
JoyAI-VL~\citep{joyaivl2026}, native & 14.9 & 16 & 16 & 25 & 27 \\
Qwen3-VL-8B~\citep{bai2025qwen3vl}, polled & 17.1 & 23 & 24 & 25 & 25 \\
LLaVA-OV2~\citep{llavaov2}, polled & 20.0 & 26 & 27 & 29 & 33 \\
Qwen3-VL-8B, reply history & 36.9 & 29 & 36 & 50 & 116 \\
MiniCPM-o~\citep{minicpmo2026}, polled & 44.3 & 59 & 61 & 64 & 68 \\
Qwen3-VL-8B, 64 frames & 44.6 & 52 & 60 & 65 & 73 \\
Qwen2.5-Omni~\citep{xu2025qwen25omni}, polled & 65.6 & 92 & 91 & 95 & 99 \\
Gemma-3n~\citep{google2025gemma3n}, polled & 69.5 & 88 & 92 & 100 & 120 \\
Flash-VStream~\citep{zhang2024flashvstream}, polled & 103.1 & -- & -- & -- & -- \\
Qwen3-VL-8B, 256 frames & 130.2 & 66 & 142 & 215 & 254 \\
\bottomrule
\end{tabular*}
\end{table}

%% file: fig_tex/tab19_backbone.tex
\begin{table}[tp]
\centering\small
\caption{\textbf{Offline answers and online scores of the six polled backbones.} Offline accuracy uses the 688 keyed items in grounding and counts unanswered items as wrong.}
\label{tab:backbone}
\setlength{\tabcolsep}{5pt}
\begin{tabular*}{\textwidth}{@{\extracolsep{\fill}}lrrrrrr@{}}
\toprule
& \multicolumn{3}{c}{\textbf{Offline, keyed items}} & \multicolumn{3}{c}{\textbf{Online, polling adapter}} \\
\cmidrule(lr){2-4}\cmidrule(lr){5-7}
\textbf{Backbone} & \textbf{Accuracy}$\uparrow$ & \textbf{Answered, \%} & \makecell{\textbf{Accuracy}\\\textbf{if answered}$\uparrow$} & \textbf{Overall}$\uparrow$ & \textbf{Timing}$\uparrow$ & \textbf{Silence}$\uparrow$ \\
\midrule
Qwen3-VL-4B~\citep{bai2025qwen3vl} & 35.8 & 80.8 & 44.2 & 26.2 & 34.0 & 28.4 \\
Qwen3-VL-8B~\citep{bai2025qwen3vl} & 35.4 & 90.7 & 39.0 & 34.8 & 77.8 & 10.9 \\
MiniCPM-o-4.5~\citep{minicpmo2026} & 33.2 & 87.8 & 37.8 & 29.0 & 44.7 & 31.6 \\
LLaVA-OV2-8B~\citep{llavaov2} & 32.7 & 99.3 & 32.9 & 34.5 & 81.0 & 16.9 \\
Qwen2.5-Omni-7B~\citep{xu2025qwen25omni} & 26.5 & 53.6 & 49.3 & 18.7 & 34.4 & 20.7 \\
Gemma-3n-E4B~\citep{google2025gemma3n} & 12.0 & 56.2 & 21.3 & 27.8 & 81.9 & 11.9 \\
\bottomrule
\end{tabular*}
\end{table}

%% file: fig_tex/tab_keyed_agreement.tex
\begin{table}[tp]
\centering\small
\caption{\textbf{Offline and online answers disagree item by item.} Each open backbone answers the 688 keyed items under polling and in offline grounding. Percentages count items, unanswered offline items count as wrong, and $\kappa$ is Cohen's kappa \citep{cohen1960kappa}.}
\label{tab:keyed_agreement}
\setlength{\tabcolsep}{5pt}
\begin{tabular*}{\textwidth}{@{\extracolsep{\fill}}lrrrrrrr@{}}
\toprule
\textbf{Backbone} & \textbf{Online} & \textbf{Offline} & \textbf{Both} & \textbf{Online only} & \textbf{Offline only} & \textbf{Agree} & $\boldsymbol{\kappa}$ \\
\midrule
Qwen3-VL-8B~\citep{bai2025qwen3vl} & 65.0 & 35.6 & 22.4 & 42.6 & 13.2 & 44.2 & $-0.03$ \\
Qwen3-VL-4B~\citep{bai2025qwen3vl} & 43.8 & 35.8 & 18.8 & 25.0 & 17.0 & 58.0 & $0.13$ \\
LLaVA-OV2~\citep{llavaov2} & 60.0 & 33.0 & 22.4 & 37.6 & 10.6 & 51.7 & $0.10$ \\
Gemma-3n~\citep{google2025gemma3n} & 53.3 & 12.1 & 7.3 & 46.1 & 4.8 & 49.1 & $0.03$ \\
MiniCPM-o~\citep{minicpmo2026} & 35.6 & 33.6 & 21.1 & 14.5 & 12.5 & 73.0 & $0.40$ \\
Qwen2.5-Omni~\citep{xu2025qwen25omni} & 35.6 & 26.5 & 7.4 & 28.2 & 19.0 & 52.8 & $-0.09$ \\
\bottomrule
\end{tabular*}
\end{table}

%% file: fig_tex/fig11_audit.tex
\begin{figure}[tp]
\centering
\includegraphics[width=\linewidth]{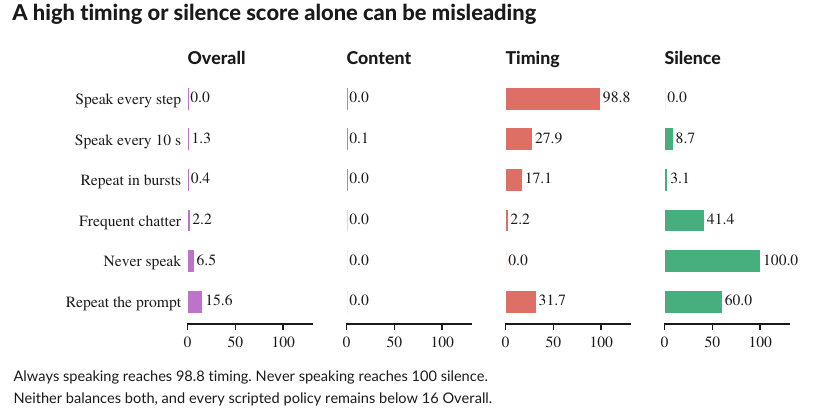}
\caption{\textbf{No scripted policy reaches a competitive Overall score.} Always speaking raises timing scores, and never speaking raises silence scores.}
\label{fig:audit}
\end{figure}

%% file: fig_tex/tab_system_scores.tex
\begin{table}[tp]
\centering\small
\caption{\textbf{System scores and 95\% bootstrap intervals.} Scores range from 0 to 100, and higher is better. Bold marks the leader of each family. Intervals resample items \citep{efron1979bootstrap}. The agentic families are named by their tools, Qwen-MM-Plugins \citep{qwenmmplugins2026} or Claude Code tools \citep{anthropic2026claudecode}. $^\dagger$Evaluated with the released code and contributed by the model authors: Senqiao Yang (Mage-VL) and Xiangyu Zeng (OneStreamer).}
\label{tab:system_scores}
\setlength{\tabcolsep}{5pt}
\renewcommand{\arraystretch}{0.865} %
\begin{tabular*}{\textwidth}{@{\extracolsep{\fill}}lrlrrr@{}}
\toprule
\textbf{System} & \textbf{Overall}$\uparrow$ & \textbf{95\% CI} & \textbf{Content}$\uparrow$ & \textbf{Timing}$\uparrow$ & \textbf{Silence}$\uparrow$ \\
\midrule
\multicolumn{6}{l}{\textit{References}}\\
Qwen3-VL-8B~\citep{bai2025qwen3vl}, blind & 15.7 & [13.9, 17.8] & 9.0 & 32.6 & 44.2 \\
Human reference & 53.6 & [49.4, 57.9] & 51.3 & 63.5 & 74.7 \\
\midrule
\multicolumn{6}{l}{\textit{Native streaming}}\\
VideoLLM-online~\citep{chen2024videollmonline} & 14.5 & [13.1, 15.9] & 3.7 & 33.6 & 47.9 \\
LiveCC~\citep{chen2025livecc} & 14.5 & -- & 25.4 & 21.0 & 7.2 \\
MMDuet2~\citep{wang2025mmduet2} & 17.0 & [15.1, 18.9] & 19.0 & 40.4 & 23.0 \\
MOSS-Video~\citep{moss2026} & 19.9 & -- & 35.1 & 24.0 & 15.5 \\
AsynKV~\citep{spotbench2026} & 27.6 & [25.1, 30.1] & 25.6 & 19.6 & \best{78.0} \\
VideoChat3~\citep{videochat3} & 28.5 & -- & 46.8 & 14.3 & 33.1 \\
OneStreamer-4B~\citep{zeng2026onestreamer}$^\dagger$ & \best{36.7} & [34.6, 38.7] & \best{52.4} & \best{57.2} & 37.9 \\
\midrule
\multicolumn{6}{l}{\textit{Turn-based polling}}\\
Flash-VStream~\citep{zhang2024flashvstream} & 14.3 & [12.8, 15.9] & 21.5 & 50.2 & 8.2 \\
Qwen2.5-Omni~\citep{xu2025qwen25omni} & 18.7 & [17.1, 20.3] & 28.3 & 34.4 & 20.7 \\
LLaVA-OV~\citep{li2025llavaonevision} & 25.6 & -- & 37.3 & 37.9 & 19.4 \\
Qwen3-VL-4B~\citep{bai2025qwen3vl} & 26.2 & [24.2, 28.2] & 35.1 & 34.0 & 28.4 \\
Gemma-3n~\citep{google2025gemma3n} & 27.8 & [26.0, 29.6] & 43.3 & 81.9 & 11.9 \\
MiniCPM-o~\citep{minicpmo2026} & 29.0 & [26.8, 31.2] & 33.1 & 44.7 & \best{31.6} \\
Mage-VL-4B~\citep{yang2026mage}$^\dagger$ & 33.7 & [31.7, 35.6] & 46.9 & 65.5 & 21.0 \\
LLaVA-OV2~\citep{llavaov2} & 34.5 & [32.6, 36.4] & 49.8 & 81.0 & 16.9 \\
Qwen3-VL-8B & 34.8 & [32.9, 36.7] & 54.2 & 77.8 & 10.9 \\
Gemini 2.5 Flash~\citep{google2026gemini} & 35.5 & [29.4, 41.4] & 55.6 & \best{93.1} & 10.8 \\
Claude Sonnet 5~\citep{anthropic2026claude} & \best{44.6} & [38.4, 50.5] & \best{66.0} & 82.9 & 17.6 \\
\midrule
\multicolumn{6}{l}{\textit{Offline temporal grounding}}\\
Gemma-3n & 16.1 & [14.3, 17.9] & 11.3 & 14.4 & 46.8 \\
Dispider~\citep{qian2025dispider} & 24.8 & -- & 26.9 & 21.9 & 76.4 \\
Qwen2.5-Omni & 27.2 & [24.8, 29.6] & 18.9 & 25.2 & 81.4 \\
MiniCPM-o & 30.6 & [28.2, 33.2] & 26.3 & 36.7 & 53.6 \\
Qwen3-VL-4B & 33.8 & [31.1, 36.3] & 26.7 & 36.2 & 57.4 \\
LLaVA-OV2 & 33.9 & [31.4, 36.4] & 25.9 & 35.6 & 63.2 \\
Qwen3-VL-8B & 35.7 & [33.1, 38.2] & 28.0 & 37.6 & 59.3 \\
Claude Sonnet 5 & 38.6 & [36.0, 41.3] & 36.0 & 32.6 & \best{91.0} \\
GPT-5.6 Sol~\citep{openai2026gpt} agent & 58.8 & [56.2, 61.3] & 56.8 & 58.4 & 80.0 \\
Claude Sonnet 5 agent & \best{66.1} & [63.4, 68.7] & \best{62.8} & \best{70.5} & 83.1 \\
\midrule
\multicolumn{6}{l}{\textit{Native Real-time Interaction System (JoyAI-VL~\citep{joyaivl2026})}}\\
Claude Opus 5~\citep{anthropic2026opus} & 30.5 & [23.6, 37.7] & 40.3 & 40.3 & \best{33.2} \\
Qwen3-VL-8B & 32.4 & [25.6, 39.0] & 50.2 & \best{90.7} & 11.0 \\
JoyAI-VL & 33.8 & [31.7, 36.0] & 45.7 & 66.8 & 29.2 \\
Qwen3-VL-4B & 35.5 & -- & 56.9 & 57.1 & 9.6 \\
Claude Sonnet 5 & \best{43.3} & [40.6, 45.9] & \best{64.4} & 77.8 & 23.0 \\
\midrule
\multicolumn{6}{l}{\textit{Agentic Interaction System (Qwen-MM-Plugins)}}\\
Gemini 3.5 Flash~\citep{google2026gemini} & 29.4 & [21.4, 37.3] & 28.8 & 23.3 & \best{66.9} \\
Claude Sonnet 5 & 56.0 & [47.2, 64.4] & 56.8 & 66.5 & 55.9 \\
GPT-5.6 Sol & \best{59.1} & [51.0, 67.0] & \best{56.8} & \best{68.6} & 60.0 \\
\midrule
\multicolumn{6}{l}{\textit{Agentic Interaction System (Claude Code tools)}}\\
Claude Sonnet 5 & \best{61.2} & [53.7, 68.2] & \best{66.7} & \best{70.7} & \best{55.5} \\
\bottomrule
\end{tabular*}
\end{table}

%% file: fig_tex/tab_task_acc.tex
\begin{table}[tp]
\centering\small
\caption{\textbf{Content accuracy by task.} Scores run from 0 to 100, higher is better, and come from the same export as Table~\ref{tab:fullset}. The agentic families are named by their tools, Qwen-MM-Plugins \citep{qwenmmplugins2026} or Claude Code tools \citep{anthropic2026claudecode}. $^\dagger$Contributed by the model authors, see Table~\ref{tab:fullset}.}
\label{tab:task_acc}
\setlength{\tabcolsep}{5pt}
\renewcommand{\arraystretch}{0.945}
\begin{tabular*}{\textwidth}{@{\extracolsep{\fill}}lrrrrrr@{}}
\toprule
\textbf{System} & Look$\uparrow$ & Recall$\uparrow$ & Time$\uparrow$ & Alert$\uparrow$ & Track$\uparrow$ & Commentate$\uparrow$ \\
\midrule
\multicolumn{7}{l}{\textbf{References}}\\
Qwen3-VL-8B~\citep{bai2025qwen3vl}, blind & 21.7 & 22.4 & 0.0 & 6.2 & 0.0 & 1.0 \\
Human reference & 78.7 & 44.4 & 63.0 & 19.5 & 36.8 & 56.9 \\
\multicolumn{7}{l}{\textbf{Native streaming}}\\
VideoLLM-online~\citep{chen2024videollmonline} & 2.4 & 6.4 & 1.6 & 3.7 & 6.8 & 0.0 \\
LiveCC~\citep{chen2025livecc} & 68.8 & 0.0 & 34.6 & 35.8 & 1.7 & 17.0 \\
MMDuet2~\citep{wang2025mmduet2} & 55.4 & 7.7 & 15.3 & 13.6 & 1.2 & 24.1 \\
MOSS-Video~\citep{moss2026} & 56.2 & 9.7 & 26.9 & 63.3 & 3.3 & 14.2 \\
AsynKV~\citep{spotbench2026} & 66.9 & 12.8 & 31.0 & 29.9 & 0.0 & 1.9 \\
VideoChat3~\citep{videochat3} & 62.5 & 6.5 & 65.4 & 72.5 & 25.0 & 18.5 \\
OneStreamer-4B~\citep{zeng2026onestreamer}$^\dagger$ & 77.7 & 47.4 & 50.0 & 79.0 & 10.5 & 18.5 \\
\multicolumn{7}{l}{\textbf{Turn-based polling}}\\
Flash-VStream~\citep{zhang2024flashvstream} & 30.7 & 2.6 & 21.4 & 42.7 & 7.4 & 2.2 \\
Qwen2.5-Omni~\citep{xu2025qwen25omni} & 23.5 & 3.2 & 42.9 & 56.0 & 4.9 & 19.4 \\
LLaVA-OV~\citep{li2025llavaonevision} & 37.5 & 22.6 & 92.3 & 39.3 & 8.3 & 28.5 \\
Qwen3-VL-4B~\citep{bai2025qwen3vl} & 71.1 & 21.2 & 15.1 & 51.5 & 1.9 & 31.0 \\
Gemma-3n~\citep{google2025gemma3n} & 57.2 & 28.8 & 49.1 & 66.6 & 7.4 & 30.7 \\
MiniCPM-o~\citep{minicpmo2026} & 75.3 & 44.2 & 15.9 & 15.9 & 8.0 & 53.8 \\
Mage-VL-4B~\citep{yang2026mage}$^\dagger$ & 68.1 & 35.3 & 51.6 & 65.5 & 7.4 & 37.3 \\
LLaVA-OV2~\citep{llavaov2} & 64.5 & 34.6 & 57.1 & 72.3 & 10.2 & 43.0 \\
Qwen3-VL-8B & 75.3 & 24.4 & 66.7 & 78.7 & 8.6 & 57.1 \\
Gemini 2.5 Flash~\citep{google2026gemini} & 81.2 & 40.0 & 46.2 & 71.3 & 17.6 & 68.1 \\
Claude Sonnet 5~\citep{anthropic2026claude} & 87.5 & 46.7 & 84.6 & 76.0 & 18.5 & 84.1 \\
\multicolumn{7}{l}{\textbf{Offline temporal grounding}}\\
Gemma-3n & 14.5 & 14.1 & 7.1 & 15.0 & 6.8 & 4.2 \\
Dispider~\citep{qian2025dispider} & 75.0 & 61.3 & 38.5 & 18.5 & 5.0 & 1.6 \\
Qwen2.5-Omni & 44.0 & 47.4 & 5.6 & 11.7 & 0.6 & 0.0 \\
MiniCPM-o & 59.6 & 66.7 & 8.7 & 11.4 & 8.0 & 2.6 \\
Qwen3-VL-4B & 63.9 & 65.4 & 7.1 & 15.3 & 3.1 & 0.8 \\
LLaVA-OV2 & 64.5 & 60.9 & 7.1 & 11.0 & 7.4 & 3.8 \\
Qwen3-VL-8B & 63.9 & 67.9 & 6.3 & 13.1 & 10.5 & 4.7 \\
Claude Sonnet 5 & 69.3 & 65.4 & 29.4 & 35.7 & 0.0 & 5.1 \\
GPT-5.6 Sol~\citep{openai2026gpt} agent & 89.2 & 91.7 & 57.9 & 52.3 & 21.7 & 18.7 \\
Claude Sonnet 5 agent & 90.4 & 92.3 & 65.9 & 57.2 & 35.1 & 29.2 \\
\multicolumn{7}{l}{\textbf{Native Real-time Interaction System (JoyAI-VL~\citep{joyaivl2026})}}\\
Claude Opus 5~\citep{anthropic2026opus} & 81.2 & 20.0 & 23.1 & 45.6 & 17.3 & 48.6 \\
Qwen3-VL-8B & 75.0 & 73.3 & 7.7 & 66.2 & 0.0 & 70.0 \\
JoyAI-VL & 59.0 & 71.8 & 41.4 & 52.7 & 3.2 & 38.4 \\
Qwen3-VL-4B & 68.8 & 38.7 & 69.2 & 85.1 & 3.3 & 37.6 \\
Claude Sonnet 5 & 79.7 & 84.1 & 70.0 & 63.4 & 19.6 & 55.8 \\
\multicolumn{7}{l}{\textbf{Agentic Interaction System (Qwen-MM-Plugins)}}\\
Gemini 3.5 Flash~\citep{google2026gemini} & 62.5 & 46.7 & 30.8 & 20.8 & 0.0 & 12.0 \\
Claude Sonnet 5 & 87.5 & 86.7 & 50.0 & 49.4 & 8.0 & 69.2 \\
GPT-5.6 Sol & 93.8 & 86.7 & 30.8 & 59.9 & 0.3 & 69.4 \\
\multicolumn{7}{l}{\textbf{Agentic Interaction System (Claude Code tools)}}\\
Claude Sonnet 5 & 93.8 & 93.3 & 53.8 & 69.4 & 14.3 & 77.9 \\
\bottomrule
\end{tabular*}
\end{table}

%% file: fig_tex/tab_task_ta.tex
\begin{table}[tp]
\centering\small
\caption{\textbf{Timing accuracy by task.} Scores run from 0 to 100, higher is better, and come from the same export as Table~\ref{tab:fullset}. The agentic families are named by their tools, Qwen-MM-Plugins \citep{qwenmmplugins2026} or Claude Code tools \citep{anthropic2026claudecode}. $^\dagger$Contributed by the model authors, see Table~\ref{tab:fullset}.}
\label{tab:task_ta}
\setlength{\tabcolsep}{5pt}
\renewcommand{\arraystretch}{0.945}
\begin{tabular*}{\textwidth}{@{\extracolsep{\fill}}lrrrrrr@{}}
\toprule
\textbf{System} & Look$\uparrow$ & Recall$\uparrow$ & Time$\uparrow$ & Alert$\uparrow$ & Track$\uparrow$ & Commentate$\uparrow$ \\
\midrule
\multicolumn{7}{l}{\textbf{References}}\\
Qwen3-VL-8B~\citep{bai2025qwen3vl}, blind & 77.5 & 95.5 & 0.7 & 6.3 & 0.0 & 25.4 \\
Human reference & 74.0 & 57.4 & 51.0 & 53.3 & 69.0 & 70.1 \\
\multicolumn{7}{l}{\textbf{Native streaming}}\\
VideoLLM-online~\citep{chen2024videollmonline} & 94.2 & 81.9 & 8.7 & 7.7 & 5.6 & 7.1 \\
LiveCC~\citep{chen2025livecc} & 71.1 & 5.6 & 24.5 & 14.5 & 2.2 & 42.0 \\
MMDuet2~\citep{wang2025mmduet2} & 62.9 & 37.8 & 37.2 & 35.2 & 26.2 & 47.4 \\
MOSS-Video~\citep{moss2026} & 71.5 & 15.7 & 18.1 & 25.3 & 4.6 & 27.7 \\
AsynKV~\citep{spotbench2026} & 59.7 & 11.9 & 20.6 & 17.2 & 0.0 & 3.1 \\
VideoChat3~\citep{videochat3} & 56.1 & 3.0 & 11.2 & 9.2 & 11.3 & 22.6 \\
OneStreamer-4B~\citep{zeng2026onestreamer}$^\dagger$ & 78.8 & 76.9 & 39.5 & 54.8 & 46.3 & 38.1 \\
\multicolumn{7}{l}{\textbf{Turn-based polling}}\\
Flash-VStream~\citep{zhang2024flashvstream} & 66.2 & 18.0 & 53.1 & 61.3 & 49.4 & 41.6 \\
Qwen2.5-Omni~\citep{xu2025qwen25omni} & 25.3 & 9.6 & 40.0 & 51.9 & 20.3 & 54.2 \\
LLaVA-OV~\citep{li2025llavaonevision} & 55.0 & 48.4 & 57.5 & 16.7 & 6.3 & 86.5 \\
Qwen3-VL-4B~\citep{bai2025qwen3vl} & 74.6 & 23.2 & 7.4 & 41.7 & 2.7 & 46.2 \\
Gemma-3n~\citep{google2025gemma3n} & 98.1 & 79.9 & 57.6 & 80.3 & 78.1 & 97.8 \\
MiniCPM-o~\citep{minicpmo2026} & 87.2 & 86.9 & 5.6 & 12.9 & 20.5 & 81.3 \\
Mage-VL-4B~\citep{yang2026mage}$^\dagger$ & 98.9 & 72.6 & 53.6 & 66.8 & 20.2 & 82.3 \\
LLaVA-OV2~\citep{llavaov2} & 98.9 & 90.5 & 76.3 & 75.3 & 55.6 & 98.2 \\
Qwen3-VL-8B & 96.6 & 44.1 & 79.6 & 79.1 & 75.3 & 96.3 \\
Gemini 2.5 Flash~\citep{google2026gemini} & 100.0 & 93.3 & 97.4 & 89.4 & 85.0 & 98.6 \\
Claude Sonnet 5~\citep{anthropic2026claude} & 98.8 & 100.0 & 84.4 & 78.7 & 52.1 & 89.0 \\
\multicolumn{7}{l}{\textbf{Offline temporal grounding}}\\
Gemma-3n & 24.0 & 23.7 & 6.2 & 7.7 & 11.0 & 17.9 \\
Dispider~\citep{qian2025dispider} & 93.8 & 100.0 & 0.0 & 0.8 & 1.2 & 0.9 \\
Qwen2.5-Omni & 69.9 & 78.8 & 2.8 & 2.0 & 1.2 & 0.0 \\
MiniCPM-o & 95.7 & 100.0 & 8.4 & 3.4 & 11.9 & 8.8 \\
Qwen3-VL-4B & 96.3 & 99.4 & 4.5 & 3.5 & 9.5 & 11.9 \\
LLaVA-OV2 & 96.3 & 99.4 & 5.8 & 4.3 & 8.3 & 5.7 \\
Qwen3-VL-8B & 96.9 & 100.0 & 2.1 & 4.0 & 13.4 & 19.1 \\
Claude Sonnet 5 & 89.7 & 87.8 & 10.5 & 5.6 & 0.8 & 6.5 \\
GPT-5.6 Sol~\citep{openai2026gpt} agent & 95.8 & 100.0 & 48.4 & 39.2 & 33.2 & 38.9 \\
Claude Sonnet 5 agent & 97.4 & 100.0 & 59.7 & 50.0 & 50.9 & 80.7 \\
\multicolumn{7}{l}{\textbf{Native Real-time Interaction System (JoyAI-VL~\citep{joyaivl2026})}}\\
Claude Opus 5~\citep{anthropic2026opus} & 85.0 & 26.7 & 20.6 & 34.9 & 28.2 & 46.3 \\
Qwen3-VL-8B & 98.8 & 80.0 & 90.5 & 97.9 & 76.2 & 97.8 \\
JoyAI-VL & 96.5 & 97.4 & 55.8 & 38.5 & 58.7 & 73.7 \\
Qwen3-VL-4B & 68.2 & 60.0 & 53.8 & 65.0 & 7.9 & 77.4 \\
Claude Sonnet 5 & 94.4 & 91.5 & 70.0 & 64.1 & 60.4 & 84.0 \\
\multicolumn{7}{l}{\textbf{Agentic Interaction System (Qwen-MM-Plugins)}}\\
Gemini 3.5 Flash~\citep{google2026gemini} & 48.8 & 44.0 & 23.1 & 12.0 & 4.8 & 10.7 \\
Claude Sonnet 5 & 93.8 & 93.3 & 31.7 & 55.8 & 46.0 & 89.6 \\
GPT-5.6 Sol & 92.5 & 93.3 & 28.5 & 50.7 & 70.4 & 87.9 \\
\multicolumn{7}{l}{\textbf{Agentic Interaction System (Claude Code tools)}}\\
Claude Sonnet 5 & 90.0 & 100.0 & 54.9 & 56.3 & 53.1 & 83.7 \\
\bottomrule
\end{tabular*}
\end{table}

%% file: fig_tex/tab_task_sc.tex
\begin{table}[tp]
\centering\small
\caption{\textbf{Silence compliance by task.} Scores run from 0 to 100, higher is better, and come from the same export as Table~\ref{tab:fullset}. The agentic families are named by their tools, Qwen-MM-Plugins \citep{qwenmmplugins2026} or Claude Code tools \citep{anthropic2026claudecode}. $^\dagger$Contributed by the model authors, see Table~\ref{tab:fullset}.}
\label{tab:task_sc}
\setlength{\tabcolsep}{5pt}
\renewcommand{\arraystretch}{0.945}
\begin{tabular*}{\textwidth}{@{\extracolsep{\fill}}lrrrrrr@{}}
\toprule
\textbf{System} & Look$\uparrow$ & Recall$\uparrow$ & Time$\uparrow$ & Alert$\uparrow$ & Track$\uparrow$ & Commentate$\uparrow$ \\
\midrule
\multicolumn{7}{l}{\textbf{References}}\\
Qwen3-VL-8B~\citep{bai2025qwen3vl}, blind & 3.6 & 69.1 & 0.0 & 66.7 & 100.0 & 15.5 \\
Human reference & 80.0 & 61.1 & 77.8 & 77.2 & 73.0 & 69.5 \\
\multicolumn{7}{l}{\textbf{Native streaming}}\\
VideoLLM-online~\citep{chen2024videollmonline} & 88.5 & 87.8 & 3.8 & 18.1 & 52.0 & 64.8 \\
LiveCC~\citep{chen2025livecc} & 0.0 & 100.0 & 3.2 & 2.6 & 3.0 & 23.9 \\
MMDuet2~\citep{wang2025mmduet2} & 54.9 & 7.3 & 0.9 & 15.1 & 34.1 & 21.7 \\
MOSS-Video~\citep{moss2026} & 6.7 & 36.4 & 0.0 & 10.2 & 3.4 & 50.1 \\
AsynKV~\citep{spotbench2026} & 89.2 & 85.7 & 71.9 & 68.6 & 100.0 & 74.5 \\
VideoChat3~\citep{videochat3} & 87.1 & 100.0 & 10.4 & 21.2 & 28.8 & 61.0 \\
OneStreamer-4B~\citep{zeng2026onestreamer}$^\dagger$ & 49.4 & 87.9 & 15.3 & 19.1 & 27.6 & 60.4 \\
\multicolumn{7}{l}{\textbf{Turn-based polling}}\\
Flash-VStream~\citep{zhang2024flashvstream} & 3.1 & 13.9 & 0.0 & 10.7 & 15.1 & 2.7 \\
Qwen2.5-Omni~\citep{xu2025qwen25omni} & 41.5 & 66.7 & 9.0 & 9.4 & 42.6 & 9.8 \\
LLaVA-OV~\citep{li2025llavaonevision} & 18.8 & 86.7 & 7.7 & 23.7 & 8.3 & 0.0 \\
Qwen3-VL-4B~\citep{bai2025qwen3vl} & 2.7 & 63.4 & 27.3 & 29.4 & 87.1 & 22.5 \\
Gemma-3n~\citep{google2025gemma3n} & 3.0 & 67.2 & 5.6 & 4.6 & 4.5 & 0.0 \\
MiniCPM-o~\citep{minicpmo2026} & 3.8 & 68.6 & 16.0 & 51.7 & 36.5 & 3.4 \\
Mage-VL-4B~\citep{yang2026mage}$^\dagger$ & 3.0 & 65.2 & 4.9 & 16.0 & 43.9 & 4.0 \\
LLaVA-OV2~\citep{llavaov2} & 3.0 & 68.3 & 0.8 & 8.7 & 23.2 & 0.0 \\
Qwen3-VL-8B & 3.0 & 66.2 & 0.0 & 9.8 & 11.7 & 0.0 \\
Gemini 2.5 Flash~\citep{google2026gemini} & 0.0 & 71.4 & 0.0 & 0.0 & 4.5 & 0.0 \\
Claude Sonnet 5~\citep{anthropic2026claude} & 0.0 & 73.3 & 0.0 & 11.1 & 15.9 & 4.0 \\
\multicolumn{7}{l}{\textbf{Offline temporal grounding}}\\
Gemma-3n & 96.4 & 100.0 & 15.9 & 39.2 & 49.4 & 57.9 \\
Dispider~\citep{qian2025dispider} & 93.8 & 100.0 & 52.6 & 62.7 & 74.6 & 95.5 \\
Qwen2.5-Omni & 100.0 & 100.0 & 39.8 & 59.3 & 88.4 & 8.3 \\
MiniCPM-o & 96.6 & 100.0 & 6.5 & 36.4 & 32.6 & 46.3 \\
Qwen3-VL-4B & 96.5 & 100.0 & 9.3 & 35.8 & 46.6 & 35.3 \\
LLaVA-OV2 & 97.6 & 100.0 & 19.4 & 38.6 & 63.1 & 79.7 \\
Qwen3-VL-8B & 97.9 & 100.0 & 11.7 & 40.9 & 48.2 & 50.9 \\
Claude Sonnet 5 & 97.9 & 100.0 & 85.5 & 81.0 & 100.0 & 79.8 \\
GPT-5.6 Sol~\citep{openai2026gpt} agent & 96.4 & 100.0 & 65.9 & 79.3 & 72.8 & 55.5 \\
Claude Sonnet 5 agent & 97.6 & 100.0 & 76.6 & 81.5 & 83.9 & 45.1 \\
\multicolumn{7}{l}{\textbf{Native Real-time Interaction System (JoyAI-VL~\citep{joyaivl2026})}}\\
Claude Opus 5~\citep{anthropic2026opus} & 13.4 & 100.0 & 42.9 & 43.8 & 24.1 & 15.0 \\
Qwen3-VL-8B & 0.0 & 66.7 & 0.0 & 3.7 & 8.4 & 0.0 \\
JoyAI-VL & 15.7 & 73.0 & 10.3 & 29.1 & 22.4 & 21.5 \\
Qwen3-VL-4B & 0.0 & 94.7 & 0.0 & 1.2 & 0.0 & 0.8 \\
Claude Sonnet 5 & 7.3 & 66.7 & 7.1 & 37.0 & 10.4 & 6.3 \\
\multicolumn{7}{l}{\textbf{Agentic Interaction System (Qwen-MM-Plugins)}}\\
Gemini 3.5 Flash~\citep{google2026gemini} & 80.0 & 87.5 & 40.0 & 62.7 & 75.7 & 32.1 \\
Claude Sonnet 5 & 68.8 & 100.0 & 54.2 & 47.5 & 47.4 & 15.0 \\
GPT-5.6 Sol & 93.8 & 100.0 & 38.5 & 69.6 & 30.7 & 16.5 \\
\multicolumn{7}{l}{\textbf{Agentic Interaction System (Claude Code tools)}}\\
Claude Sonnet 5 & 62.5 & 86.7 & 58.3 & 50.9 & 54.5 & 8.6 \\
\bottomrule
\end{tabular*}
\end{table}

%% file: fig_tex/tab_violations.tex
\begin{table}[tp]
\centering\small
\caption{\textbf{Violations per reference event.} Premature and repeated replies are divided by each item's reference events and averaged over positive items. Zero silence is the share of positive items whose silence compliance is zero. Rows cover the 30 current configurations, and the agentic families are named by their tools, Qwen-MM-Plugins \citep{qwenmmplugins2026} or Claude Code tools \citep{anthropic2026claudecode}.}
\label{tab:violations}
\setlength{\tabcolsep}{5pt}
\begin{tabular*}{\textwidth}{@{\extracolsep{\fill}}lrrr@{}}
\toprule
\textbf{System} & \textbf{Premature per event}$\downarrow$ & \textbf{Repeated per event}$\downarrow$ & \textbf{Zero silence, \%}$\downarrow$ \\
\midrule
\multicolumn{4}{l}{\textit{References}}\\
Qwen3-VL-8B~\citep{bai2025qwen3vl}, blind & 1.91 & 2.76 & 66.2 \\
Human reference & 0.09 & 0.18 & 16.3 \\
\multicolumn{4}{l}{\textit{Native streaming}}\\
VideoLLM-online~\citep{chen2024videollmonline} & 1.70 & 3.58 & 41.1 \\
MMDuet2~\citep{wang2025mmduet2} & 7.50 & 7.04 & 76.4 \\
\multicolumn{4}{l}{\textit{Turn-based polling}}\\
Flash-VStream~\citep{zhang2024flashvstream} & 9.82 & 15.41 & 97.4 \\
Qwen2.5-Omni~\citep{xu2025qwen25omni} & 3.59 & 7.23 & 75.6 \\
Qwen3-VL-4B~\citep{bai2025qwen3vl} & 0.88 & 2.78 & 77.1 \\
Gemma-3n~\citep{google2025gemma3n} & 7.23 & 19.23 & 88.5 \\
MiniCPM-o~\citep{minicpmo2026} & 1.44 & 3.40 & 71.7 \\
LLaVA-OV2~\citep{llavaov2} & 5.84 & 16.23 & 84.8 \\
Qwen3-VL-8B & 4.45 & 14.83 & 92.0 \\
Gemini 2.5 Flash~\citep{google2026gemini} & 8.26 & 18.30 & 89.5 \\
Claude Sonnet 5~\citep{anthropic2026claude} & 3.32 & 9.26 & 84.8 \\
\multicolumn{4}{l}{\textit{Offline temporal grounding}}\\
Gemma-3n & 0.61 & 0.21 & 47.0 \\
Qwen2.5-Omni & 0.09 & 0.01 & 19.0 \\
MiniCPM-o & 0.71 & 0.15 & 36.0 \\
Qwen3-VL-4B & 0.78 & 0.15 & 36.4 \\
LLaVA-OV2 & 0.41 & 0.06 & 31.1 \\
Qwen3-VL-8B & 0.39 & 0.14 & 35.7 \\
Claude Sonnet 5 & 0.04 & 0.01 & 7.3 \\
GPT-5.6 Sol~\citep{openai2026gpt} agent & 0.14 & 0.09 & 14.1 \\
Claude Sonnet 5 agent & 0.12 & 0.13 & 12.9 \\
\multicolumn{4}{l}{\textit{Native Real-time Interaction System (JoyAI-VL~\citep{joyaivl2026})}}\\
Claude Opus 5~\citep{anthropic2026opus} & 0.43 & 1.10 & 61.4 \\
Qwen3-VL-8B & 18.82 & 23.58 & 88.9 \\
JoyAI-VL & 2.83 & 8.04 & 64.2 \\
Claude Sonnet 5 & 0.92 & 3.33 & 71.8 \\
\multicolumn{4}{l}{\textit{Agentic Interaction System (Qwen-MM-Plugins)}}\\
Gemini 3.5 Flash~\citep{google2026gemini} & 0.18 & 0.11 & 27.7 \\
Claude Sonnet 5 & 0.52 & 0.70 & 40.2 \\
GPT-5.6 Sol & 0.41 & 0.63 & 36.8 \\
\multicolumn{4}{l}{\textit{Agentic Interaction System (Claude Code tools)}}\\
Claude Sonnet 5 & 0.59 & 0.75 & 41.1 \\
\bottomrule
\end{tabular*}
\end{table}

%% file: fig_tex/fig15_harness.tex
\begin{figure}[tp]
\centering
\includegraphics[width=\linewidth]{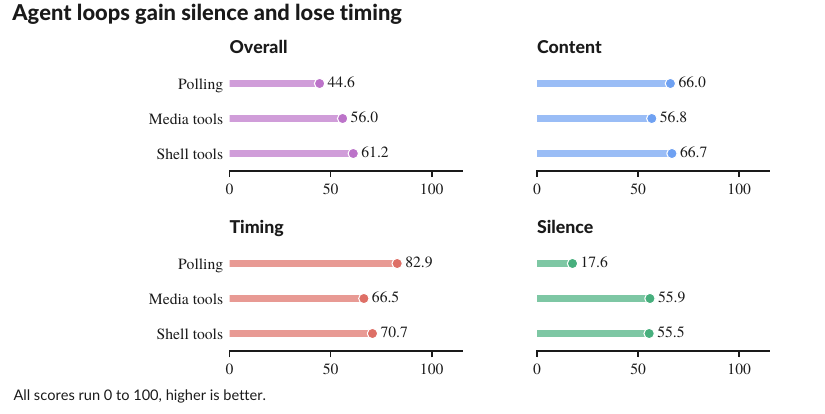}
\caption{\textbf{The response loop changes behavior.} Claude Sonnet 5 \citep{anthropic2026claude} under polling and in two agent runs, with Overall and all three component scores.}
\label{fig:harness}
\end{figure}

%% file: fig_tex/tab16_ablation.tex
\begin{table}[tp]
\centering
\caption{\textbf{Polling-adapter ablations.} Sampling, window size, response history, and prompt variants change content, timing, and silence differently.}
\label{tab:ablation}
\vspace{0pt}
\footnotesize
\setlength{\tabcolsep}{4pt}
\begin{tabular*}{\textwidth}{@{\extracolsep{\fill}}l c ccc@{}}
\toprule
\textbf{Configuration} & \textbf{Overall}$\uparrow$ & \textbf{Content}$\uparrow$ & \textbf{Timing}$\uparrow$ & \textbf{Silence}$\uparrow$ \\
\midrule
\multicolumn{5}{l}{\textit{Sampling rate, Qwen3-VL-8B \citep{bai2025qwen3vl}, fresh prompt, 16 frames}}\\
4\,fps & 33.4 & 51.8 & 75.7 & 10.5 \\
2\,fps & 34.8 & 54.2 & 77.8 & 10.9 \\
1\,fps & 34.7 & 52.7 & 76.9 & 13.1 \\
0.5\,fps & 35.4 & 53.6 & 76.6 & 14.5 \\
\midrule
\multicolumn{5}{l}{\textit{Window length, Qwen3-VL-8B~\citep{bai2025qwen3vl}, fresh prompt, 2\,fps}}\\
16 frames & 34.8 & 54.2 & 77.8 & 10.9 \\
64 frames & 35.6 & 54.3 & 78.8 & 13.8 \\
256 frames & 39.0 & 58.7 & 81.3 & 17.2 \\
\midrule
\multicolumn{5}{l}{\textit{Fresh prompt versus reply history}}\\
Qwen3-VL-8B, fresh prompt & 34.8 & 54.2 & 77.8 & 10.9 \\
Qwen3-VL-8B, reply history & 32.9 & 41.8 & 76.0 & 25.6 \\
Qwen3-VL-4B~\citep{bai2025qwen3vl}, fresh prompt & 26.2 & 35.1 & 34.0 & 28.4 \\
Qwen3-VL-4B, reply history & 34.6 & 40.6 & 57.5 & 36.0 \\
LLaVA-OV2-8B \citep{llavaov2}, fresh prompt & 34.5 & 49.8 & 81.0 & 16.9 \\
LLaVA-OV2-8B~\citep{llavaov2}, reply history & 24.9 & 31.1 & 72.2 & 21.6 \\
Gemma-3n~\citep{google2025gemma3n}, fresh prompt & 27.8 & 43.3 & 81.9 & 11.9 \\
Gemma-3n, reply history & 19.4 & 12.8 & 31.0 & 54.7 \\
MiniCPM-o~\citep{minicpmo2026}, fresh prompt & 29.0 & 33.1 & 44.7 & 31.6 \\
MiniCPM-o, reply history & 39.1 & 39.6 & 60.5 & 48.5 \\
Qwen2.5-Omni~\citep{xu2025qwen25omni}, fresh prompt & 18.7 & 28.3 & 34.4 & 20.7 \\
Qwen2.5-Omni, reply history & 22.2 & 28.8 & 59.6 & 26.5 \\
\midrule
\multicolumn{5}{l}{\textit{Prompt hints, Qwen3-VL-8B, fresh prompt, agent items}}\\
Default prompt & 33.9 & 57.7 & 75.5 & 5.9 \\
Hint set v2 & 34.6 & 57.3 & 70.0 & 8.8 \\
Hint set v3 & 34.8 & 56.9 & 70.2 & 7.2 \\
\bottomrule
\end{tabular*}
\end{table}

%% file: fig_tex/fig13_ablation.tex
\begin{figure}[tp]
\centering
\includegraphics[width=\linewidth]{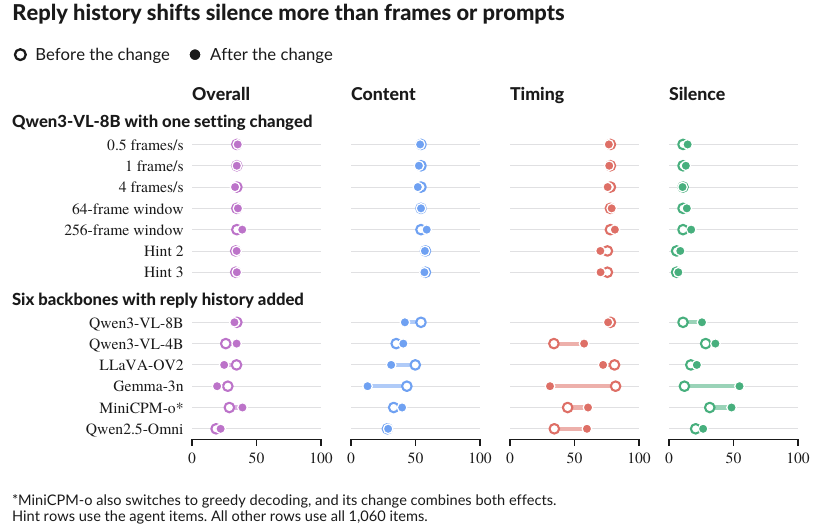}
\caption{\textbf{History shifts silence more than frames.} Upper rows change one Qwen3-VL-8B \citep{bai2025qwen3vl} setting, and lower rows add reply history to six backbones.}
\label{fig:ablation}
\end{figure}

%% file: fig_tex/fig2_response_errors.tex
\begin{figure}[tp]
\centering
\includegraphics[width=\linewidth]{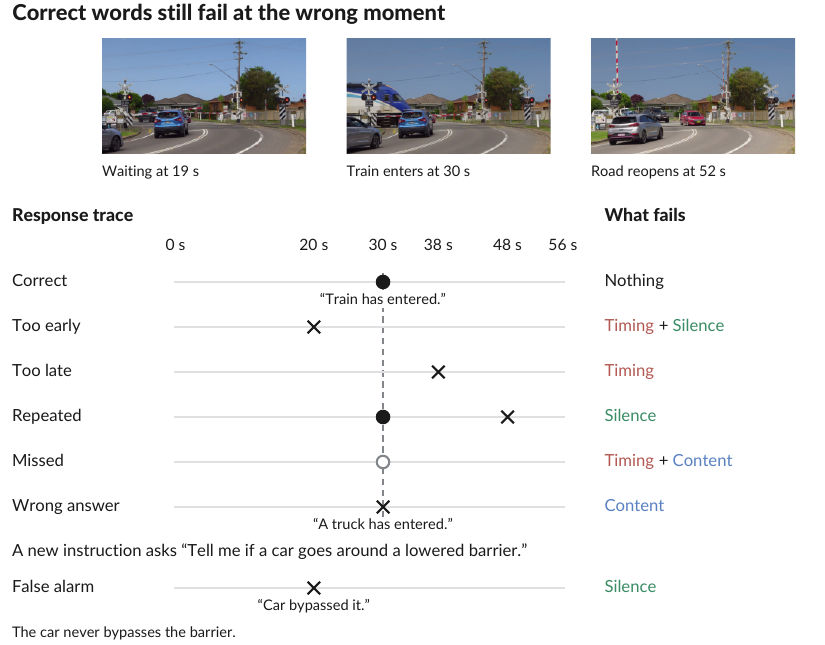}
\caption{\textbf{Six ways to fail on one safety video.} The instruction at 0\,s asks ``Tell me once when the train enters the crossing.'' Constructed traces separate content, timing, and silence errors.}
\label{fig:response_errors}
\end{figure}

%% file: fig_tex/fig18_cases.tex
\begin{figure}[p]
\centering
\includegraphics[width=\linewidth]{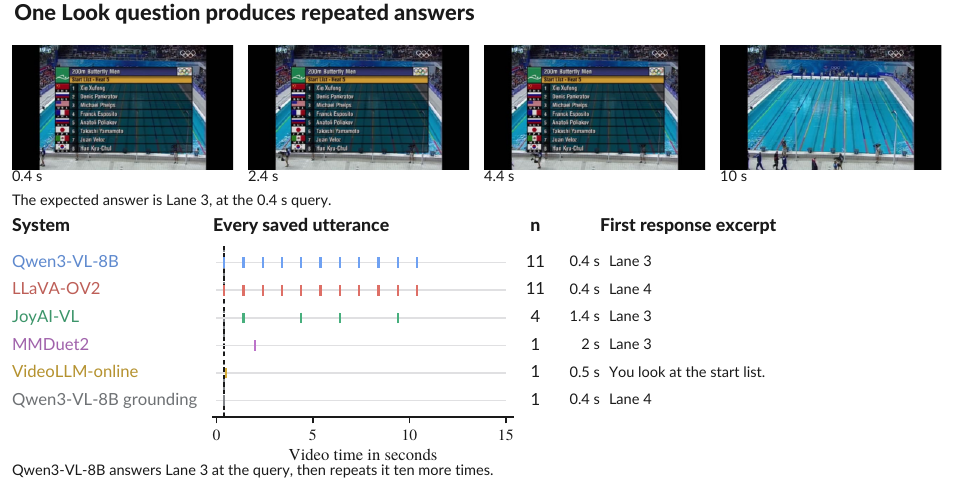}
\caption{\textbf{Look case.} The user asks ``According to the start list, which lane is Michael Phelps swimming in?'' at 0.4\,s in saved item \texttt{-6QYAYbS6eU\#0}. Source frames, reference events, and six saved response traces. Excerpts quote the first reply and truncate long text.}
\label{fig:case_IVQA}
\end{figure}
\begin{figure}[p]
\centering
\includegraphics[width=\linewidth]{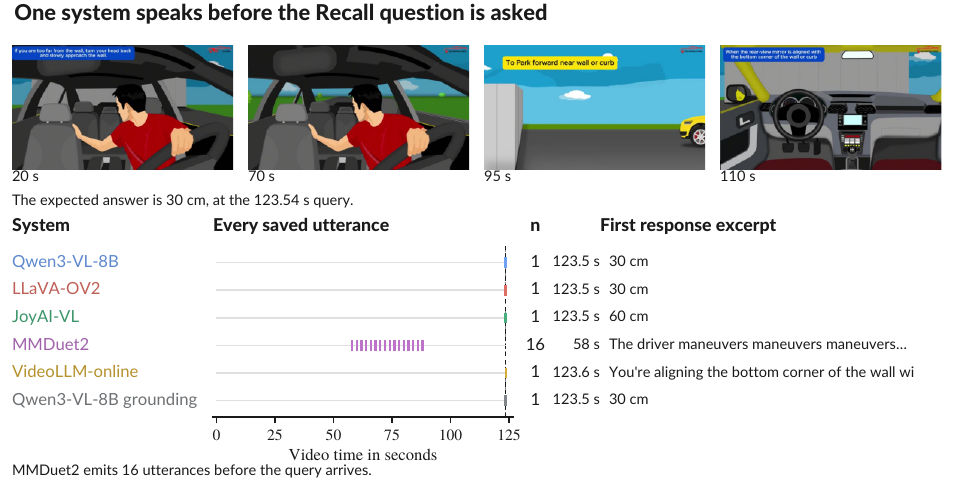}
\caption{\textbf{Recall case.} The user asks ``When you align the bottom corner of the wall with your front door handle, how far is the back of the car from the wall?'' at 123.54\,s in saved item \texttt{-Hv4pJjjZpY\#0}. Source frames, reference events, and six saved response traces. Excerpts quote the first reply and truncate long text.}
\label{fig:case_LVM}
\end{figure}
\begin{figure}[p]
\centering
\includegraphics[width=\linewidth]{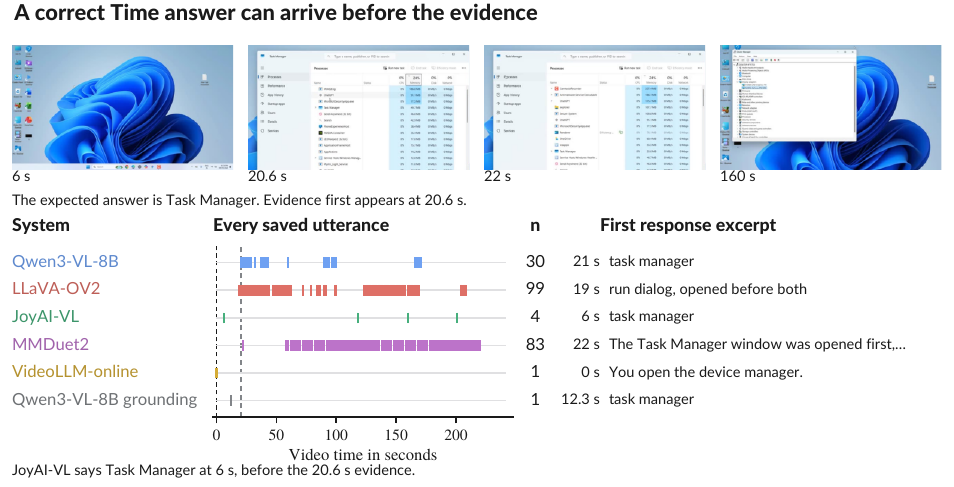}
\caption{\textbf{Time case.} The user asks ``In the video, which was opened first, Task Manager or Device Manager?'' at 0\,s in saved item \texttt{-6qvDiJJ388\#0}. Source frames, reference events, and six saved response traces. Excerpts quote the first reply and truncate long text.}
\label{fig:case_TOA}
\end{figure}
\begin{figure}[p]
\centering
\includegraphics[width=\linewidth]{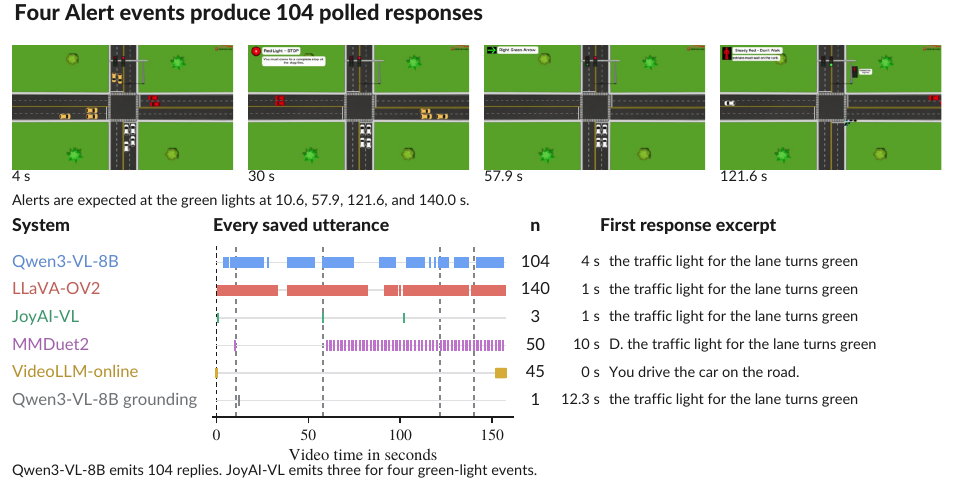}
\caption{\textbf{Alert case.} The user asks ``Let me know when the traffic light for the lane turns green.'' at 0\,s in saved item \texttt{-31At72iGnk\#0}. Source frames, reference events, and six saved response traces. Excerpts quote the first reply and truncate long text.}
\label{fig:case_PTR}
\end{figure}
\begin{figure}[p]
\centering
\includegraphics[width=\linewidth]{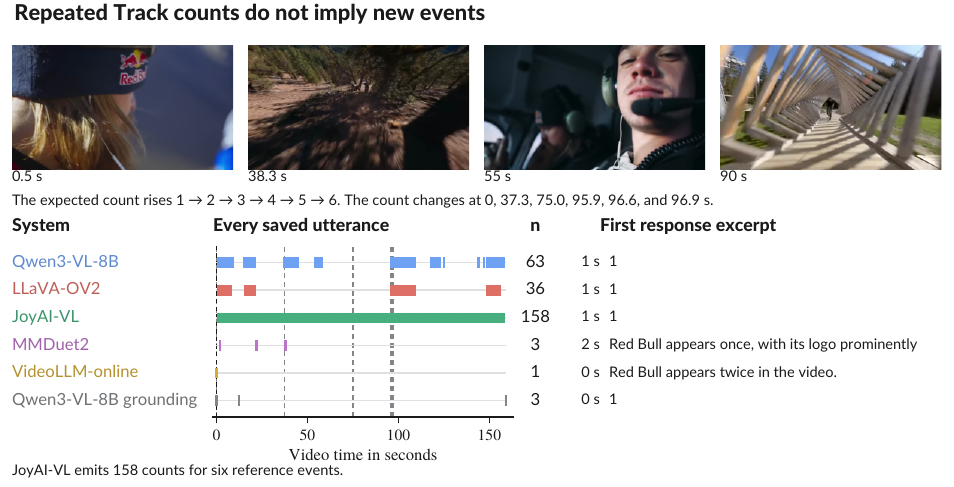}
\caption{\textbf{Track case.} The user asks ``How many times does Red Bull appear?'' at 0\,s in saved item \texttt{--VCWegJ\_oY\#0}. Source frames, reference events, and six saved response traces. Excerpts quote the first reply and truncate long text.}
\label{fig:case_CST}
\end{figure}
\begin{figure}[p]
\centering
\includegraphics[width=\linewidth]{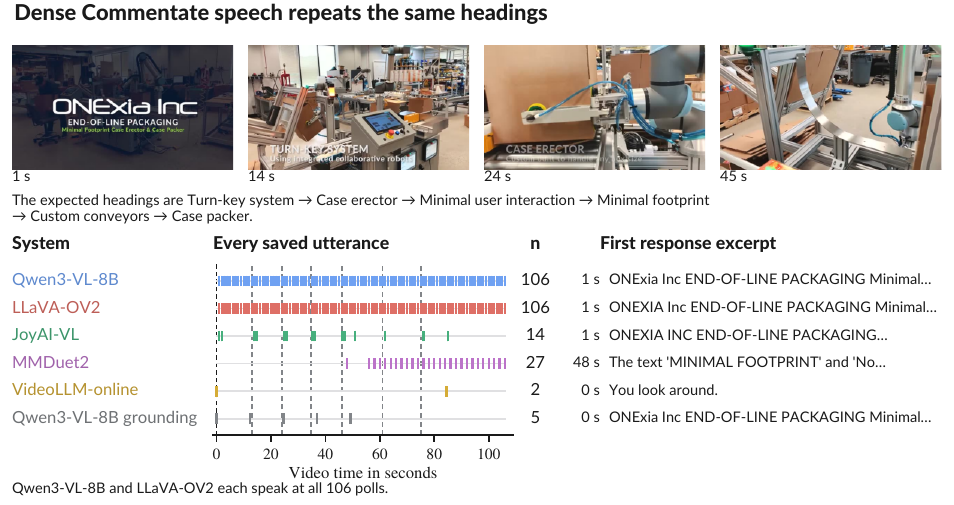}
\caption{\textbf{Commentate case.} The user asks ``Output each on-screen feature heading in real time.'' at 0\,s in saved item \texttt{-CESiuu7CYs\#0}. Source frames, reference events, and six saved response traces. Excerpts quote the first reply and truncate long text.}
\label{fig:case_LCG}
\end{figure}

%% file: fig_tex/fig19_main_case.tex
\begin{figure}[tp]
\centering
\includegraphics[width=\linewidth]{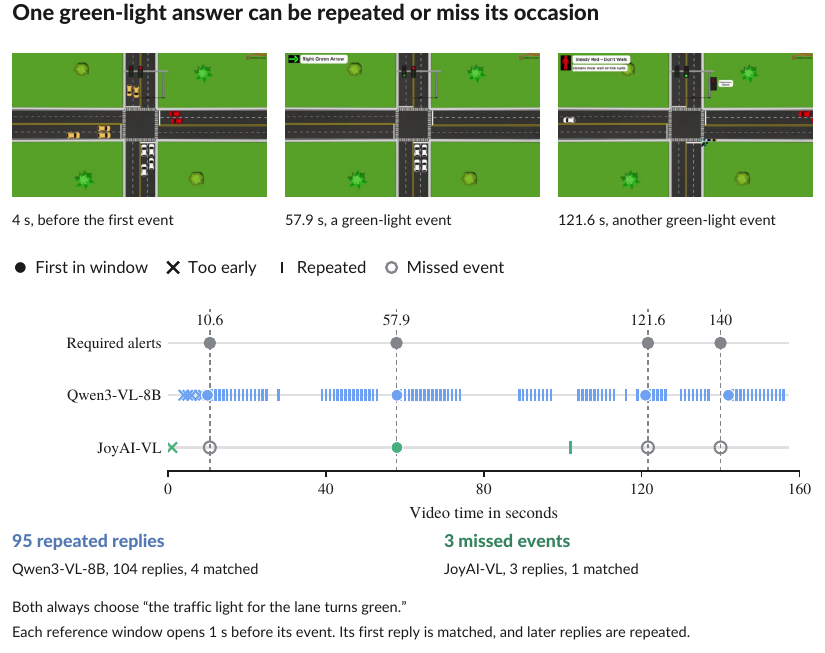}
\caption{\textbf{A correct answer can be repeated or missed.} The saved Alert instruction is ``Let me know when the traffic light for the lane turns green.'' Rows mark the required alerts and each system's replies on one saved Alert item, classified by the response-window rule.}
\label{fig:measured_case}
\end{figure}

%% file: fig_tex/tab_agents_summary.tex
\begin{table}[tp]
\centering\small
\caption{\textbf{Agentic systems raise silence.} All rows share the agent items. Agents use Qwen-MM-Plugins \citep{qwenmmplugins2026} media tools except the shell-tool run. s/poll is the median wall-clock seconds per poll, and bold marks the best value in each column.}
\label{tab:agents_summary}
\setlength{\tabcolsep}{5pt}
\begin{tabular*}{\textwidth}{@{\extracolsep{\fill}}lrrrrr@{}}
\toprule
\textbf{System} & \textbf{Overall}$\uparrow$ & \textbf{Content}$\uparrow$ & \textbf{Timing}$\uparrow$ & \textbf{Silence}$\uparrow$ & \textbf{s/poll}$\downarrow$ \\
\midrule
Claude Sonnet 5~\citep{anthropic2026claude}, polled \citep{anthropic2026claude} & 44.6 & 66.0 & 82.9 & 17.6 & 2.4 \\
Claude Sonnet 5 agent & 56.0 & 56.8 & 66.5 & 55.9 & 28.7 \\
Claude Sonnet 5 agent, shell tools & \best{61.2} & \best{66.7} & 70.7 & 55.5 & 37.2 \\
GPT-5.6 Sol~\citep{openai2026gpt} agent \citep{openai2026gpt} & 59.1 & 56.8 & 68.6 & 60.0 & -- \\
Gemini 2.5 Flash~\citep{google2026gemini}, polled & 35.5 & 55.6 & \best{93.1} & 10.8 & \best{1.9} \\
Gemini 3.5 Flash agent \citep{google2026gemini} & 29.4 & 28.8 & 23.3 & \best{66.9} & -- \\
\bottomrule
\end{tabular*}
\end{table}

%% file: fig_tex/tab_oracle_restraint.tex
\begin{table}[tp]
\centering\small
\caption{\textbf{Oracle restraint.} Reference annotations remove premature, repeated, and negative-stream replies. Retained answers and timestamps are unchanged.}
\label{tab:oracle_restraint}
\setlength{\tabcolsep}{3pt}
\begin{tabular*}{\textwidth}{@{\extracolsep{\fill}}lrrrr@{}}
\toprule
\textbf{Configuration} & \textbf{Overall}$\uparrow$ & \textbf{Content}$\uparrow$ & \textbf{Timing}$\uparrow$ & \textbf{Silence}$\uparrow$ \\
\midrule
Qwen3-VL-8B~\citep{bai2025qwen3vl}, original & 33.9 & 57.7 & 75.5 & 5.9 \\
Qwen3-VL-8B, oracle restraint & 66.2 & 56.0 & 75.5 & 100.0 \\
JoyAI-VL~\citep{joyaivl2026}, original & 38.4 & 53.4 & 69.2 & 25.9 \\
JoyAI-VL, oracle restraint & 63.9 & 53.0 & 69.2 & 100.0 \\
\bottomrule
\end{tabular*}
\end{table}

%% file: fig_tex/tab_threshold.tex
\begin{table}[tp]
\centering\small
\caption{\textbf{VideoLLM-online at three response thresholds.} The 0.725 row is the released setting of VideoLLM-online \citep{chen2024videollmonline}, scored on the same items.}
\label{tab:threshold}
\setlength{\tabcolsep}{5pt}
\begin{tabular*}{\textwidth}{@{\extracolsep{\fill}}lrrrr@{}}
\toprule
\textbf{Threshold} & \textbf{Overall}$\uparrow$ & \textbf{Content}$\uparrow$ & \textbf{Timing}$\uparrow$ & \textbf{Silence}$\uparrow$ \\
\midrule
0.5 & 17.2 & 6.3 & 30.2 & 56.4 \\
0.725 & 16.2 & 7.3 & 34.1 & 47.0 \\
0.9 & 7.4 & 1.0 & 51.7 & 18.4 \\
\bottomrule
\end{tabular*}
\end{table}

%% file: fig_tex/fig_blind.tex
\begin{figure}[!htb]
\centering
\includegraphics[width=\linewidth]{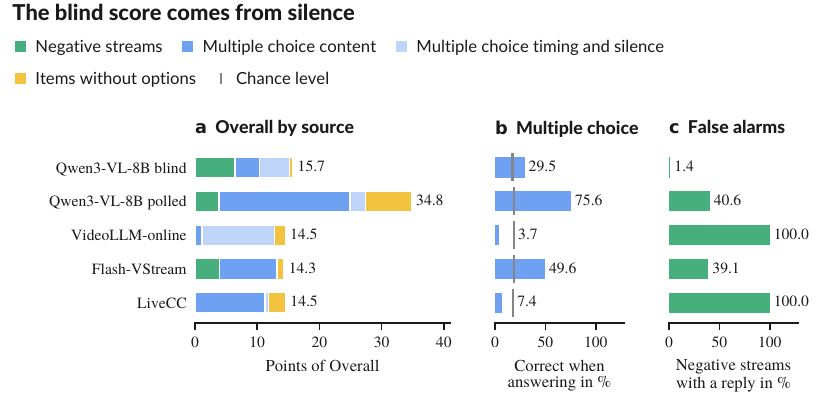}
\caption{\textbf{The blind score comes from silence.} Panel \textbf{a} splits Overall by item group, \textbf{b} gives accuracy on answered multiple-choice items, and \textbf{c} the share of negative streams with a reply.}
\label{fig:blind}
\end{figure}

%% file: fig_tex/fig_reply_history.tex
\begin{figure}[!htb]
\centering
\includegraphics[width=\linewidth]{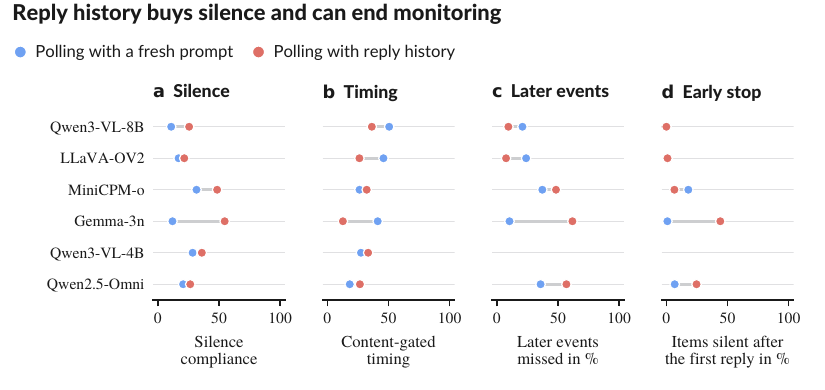}
\caption{\textbf{Reply history buys silence and can end monitoring.} Panels \textbf{a} and \textbf{b} use all items, and \textbf{c} and \textbf{d} use counting instructions, where Qwen3-VL-4B~\citep{bai2025qwen3vl} has too few first answers to show.}
\label{fig:reply_history}
\end{figure}

%% file: fig_tex/tab_errors_duration.tex
\begin{table}[tp]
\centering\small
\caption{\textbf{Longer monitoring costs detection and leaves content unchanged.} Each row is the change per doubling of monitored time in a regression with system, task, and domain fixed effects.}
\label{tab:errors_duration}
\setlength{\tabcolsep}{5pt}
\begin{tabular*}{\textwidth}{@{\extracolsep{\fill}}lrl@{}}
\toprule
\textbf{Outcome} & \textbf{Change per doubling} & \textbf{95\% interval} \\
\midrule
Missed events per 100 events, standing instructions & +2.4 & [+1.1, +3.7] \\
Wrong content per 100 events, standing instructions & 0.0 & [$-$1.1, +1.2] \\
Negative streams with a reply, \% & +8.5 & [+2.4, +14.6] \\
Accuracy when asked, per doubling of history & $-$2.2 & [$-$3.8, $-$0.6] \\
\bottomrule
\end{tabular*}
\end{table}

%% file: fig_tex/tab_errors_domain.tex
\begin{table}[tp]
\centering\small
\caption{\textbf{Domains differ mainly in content.} Each cell is the median over the 19 systems run on all 1,060 interactions. Missed events and wrong content are shares of the reference events.}
\label{tab:errors_domain}
\setlength{\tabcolsep}{5pt}
\begin{tabular*}{\textwidth}{@{\extracolsep{\fill}}lrrrr@{}}
\toprule
\textbf{Domain} & \textbf{Interactions} & \makecell{\textbf{Content}\\\textbf{accuracy}$\uparrow$} & \makecell{\textbf{Missed}\\\textbf{events \%}$\downarrow$} & \makecell{\textbf{Wrong}\\\textbf{content \%}$\downarrow$} \\
\midrule
Cooking and kitchen & 127 & 16.4 & 73.4 & 15.9 \\
Crafts, creation and skills & 46 & 23.1 & 74.4 & 19.8 \\
Medical, first-aid and care & 44 & 28.9 & 52.2 & 19.6 \\
Home and life skills & 115 & 29.2 & 66.2 & 16.7 \\
Games, AR and entertainment & 85 & 29.3 & 81.4 & 13.2 \\
Professional, office and accessibility & 104 & 31.6 & 68.8 & 18.2 \\
Sports and exercise coaching & 254 & 33.4 & 77.9 & 13.9 \\
Driving, traffic and mobility & 76 & 38.5 & 69.9 & 12.3 \\
Education and children & 185 & 39.3 & 59.2 & 16.8 \\
Industrial, repair and operations & 24 & 39.3 & 57.4 & 29.8 \\
\bottomrule
\end{tabular*}
\end{table}

%% file: fig_tex/fig_failure_cases.tex
\begin{figure}[p]
\centering
\includegraphics[width=\linewidth,height=\dimexpr\textheight-48pt\relax,keepaspectratio]{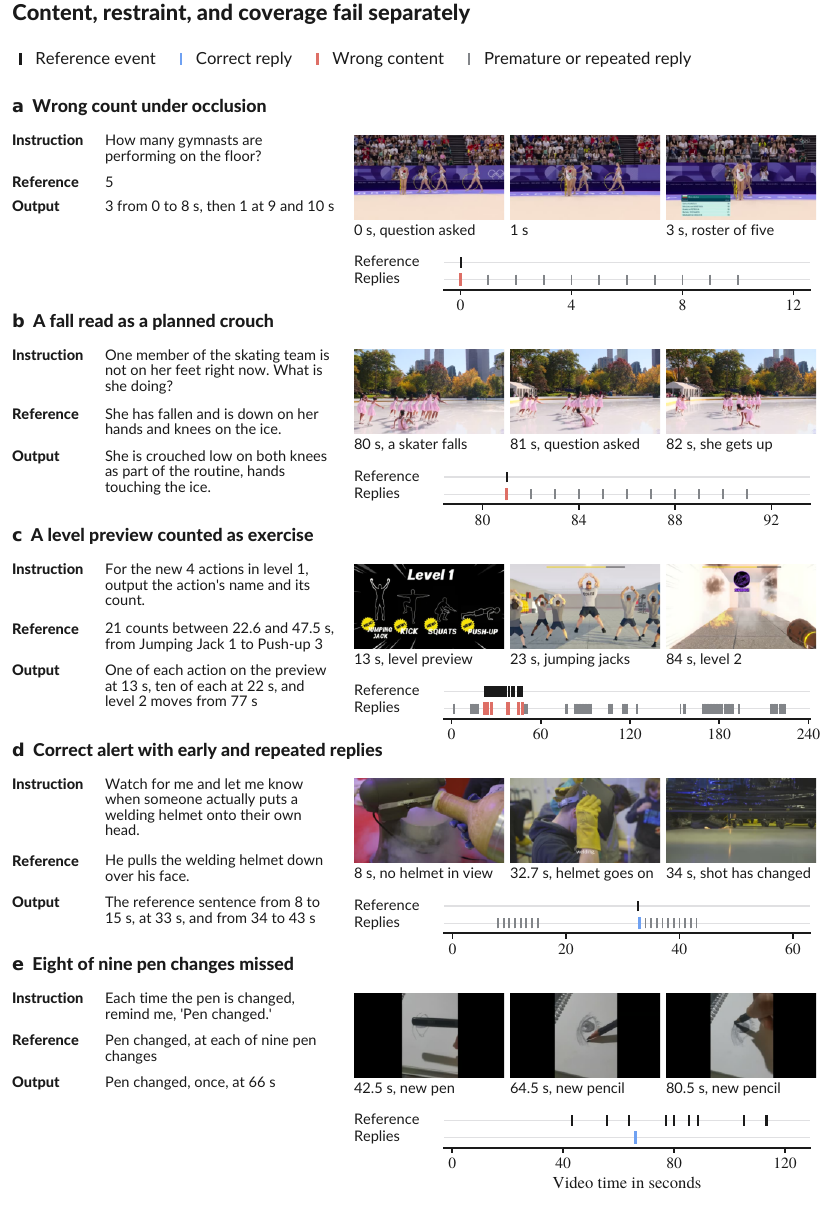}
\caption{\textbf{Five verified failures of polled Qwen3-VL-8B~\citep{bai2025qwen3vl}.} Each panel is typical of one error type and marks wrong content separately from premature and repeated replies.}
\label{fig:failure_cases}
\end{figure}

%% file: fig_tex/fig_scene_target.tex
\begin{figure}[!htb]
\centering
\includegraphics[width=\linewidth]{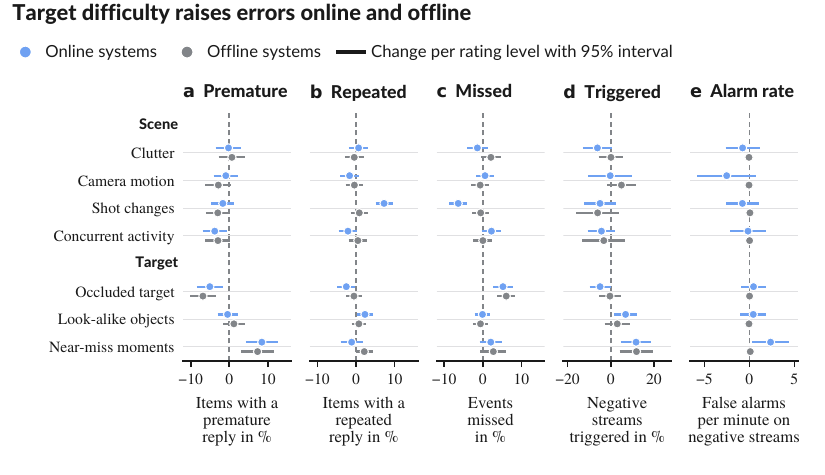}
\caption{\textbf{Target difficulty raises errors online and offline.} Each marker is the change per rating level with its 95\% interval, for the 10 online and the 9 offline full-set configurations.}
\label{fig:scene_target}
\end{figure}